\documentclass[runningheads]{llncs}
\usepackage{accv}
\usepackage{accvabbrv}

\usepackage{graphicx}
\usepackage{booktabs}
\usepackage[utf8]{inputenc}
\usepackage[T1]{fontenc}
\usepackage{url}
\usepackage{amsfonts}
\usepackage{nicefrac}
\usepackage{microtype}
\usepackage{xcolor}

\usepackage{amsmath}
\usepackage{amssymb}
\usepackage{bm}
\usepackage[disable]{todonotes}
\usepackage{algorithm}
\usepackage{algorithmic}
\usepackage{float}
\usepackage{cite}
\usepackage{multirow}
\usepackage{mwe}
\usepackage{caption}
\usepackage{subcaption}

\usepackage[accsupp]{axessibility}  
\usepackage{hyperref}
\usepackage{orcidlink}

\DeclareMathOperator{\Tr}{Tr}
\DeclareMathOperator*{\argmax}{arg\,max}
\newcommand{\mat}[1]{#1}
\newcommand{\rowlabel}[1]{\raisebox{0.55cm}{\parbox[c][0.5cm][c]{0.8cm}{\centering #1}}}

\begin{document}

\title{Progressive Permutation Alignment with Temperature-Controlled Spectral Sharpening} 
\titlerunning{Progressive Permutation Alignment with Spectral Sharpening}
\author{Yuta Horioka\inst{}\thanks{Equal contribution.}\orcidlink{0009-0000-6139-9578} \and
Naoya Chiba\inst{}$^{\star}$\orcidlink{0000-0003-3332-4426} \and
Yuki Uranishi\inst{}\orcidlink{0000-0001-6027-0334}}

\authorrunning{Y.~Horioka et al.}
\institute{The University of Osaka}

\maketitle

\begin{abstract}
In implicit neural representations (INRs), 3D signals are encoded directly in network weights, making stable weight-space operations essential. However, permutation symmetry causes independently trained yet functionally equivalent models to appear distant in weight space, often inducing large loss barriers under linear interpolation. Existing neuron alignment methods either commit early to discrete matches and become trapped in inconsistent local optima or rely on computationally expensive gradient optimization over continuous relaxations, both of which are unreliable in challenging cases such as NeRF.
We adopt progressive alignment, a layer-wise strategy that follows inter-layer dependencies from input to output. At each layer, a correlation matrix is estimated and aligned with a discrete anchor obtained by the Hungarian algorithm, after which temperature-controlled spectral annealing gradually sharpens the correlation structure toward a permutation matrix. This update is deterministic and lightweight, avoiding the inner-loop gradient optimization required by continuous relaxation methods. A final refinement stage further optimizes the network layer-by-layer via weight matching, starting from the aligned solution.
Experiments on MLPs, CNNs, and Neural Radiance Fields (NeRFs) show that the proposed method reliably recovers neuron correspondences and achieves stable alignment across architectures. In practical experiments aligning independently trained NeRF models, our method reduces the loss barriers under weight interpolation compared with existing approaches.
\end{abstract}

\section{Introduction}
\label{sec:intro}

\begin{figure}[tb]
\centering
\includegraphics[width=0.9\linewidth]{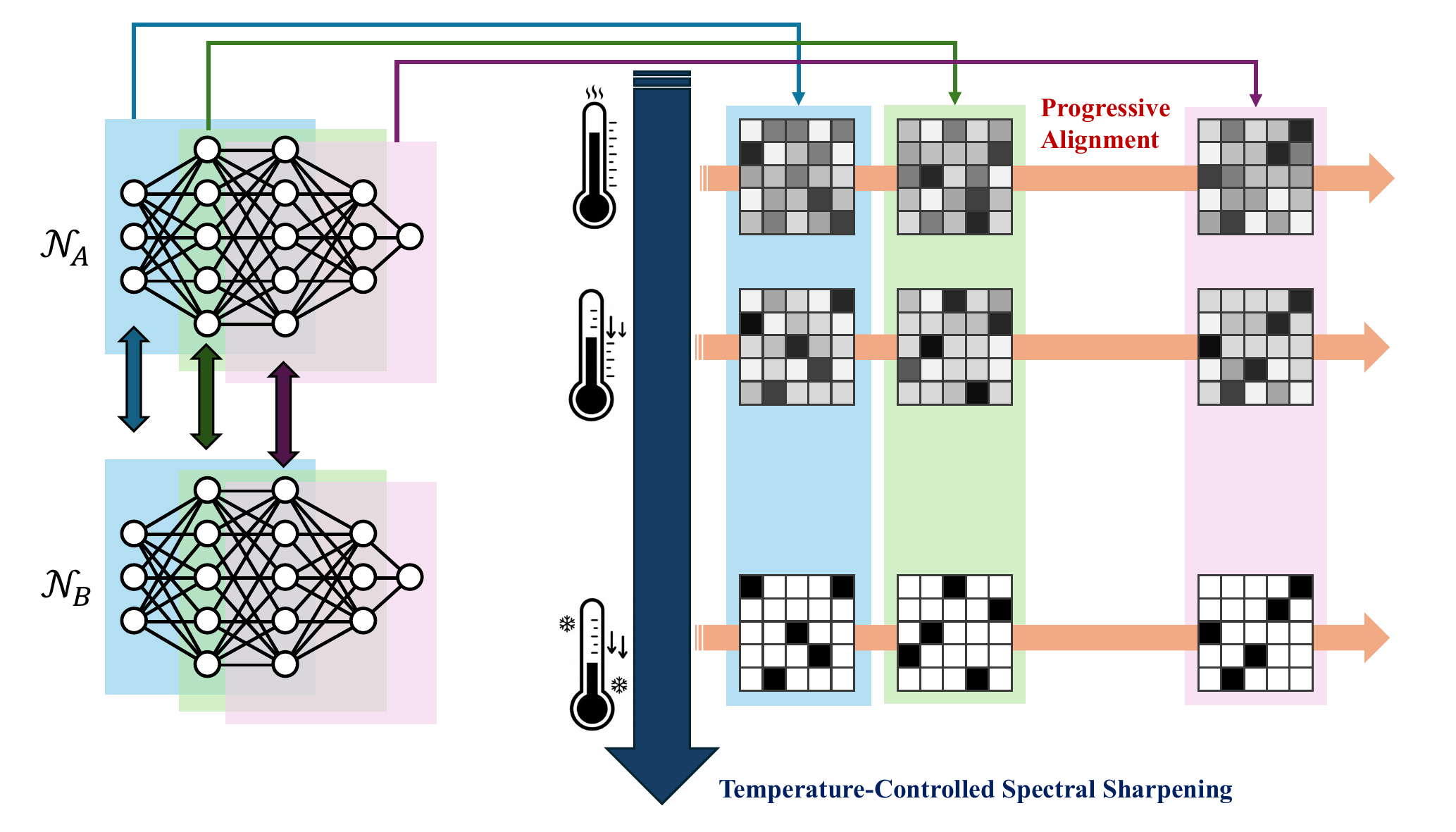} 
\caption{Overview of Progressive Alignment with Temperature-Controlled Spectral Sharpening.
Two independently trained neural fields, $\mathcal{N}_A$ and $\mathcal{N}_B$ exhibit permutation symmetry across the hidden layers. Our method aligns them in a progressive, layer-wise manner following the inter-layer dependency from input to output. At each layer, a correlation matrix is computed and gradually sharpened via temperature-controlled spectral projection.}
\label{fig:proposed_overview}
\end{figure}

Implicit Neural Representations (INRs) encode continuous signals directly in network weights. NeRF represents 3D scenes as neural weights, allowing models to function as data representations~\cite{deluigi2023inr2vec,zamaramirez2024nf2vec,ballerini2026weight,amaduzzi2025weights2space}. Along with advances such as Model Soups and task arithmetic~\cite{wortsman2022modelsoups,ilharco2023taskarithmetic}, direct manipulation in weight space, such as interpolation and merging, has become an important research topic~\cite{navon2024equivariant,schurholt2024sane,gupta2026deepweightflow}.

However, neural networks exhibit permutation symmetry. In Multi-Layer Perceptrons (MLPs), the units in each hidden layer can be arbitrarily permuted without changing the network's function, provided the weights of adjacent layers are correspondingly transformed. Therefore, numerous sets of weights exist that represent the same function, and simple weight averaging or linear interpolation often leads to performance degradation, resulting in a high loss barrier along the interpolation path~\cite{entezari2022role}.

Prior work has shown the existence of Linear Mode Connectivity (LMC), which maps independently trained models into the same loss basin via appropriate permutations, enabling interpolation without loss barriers~\cite{entezari2022role}. LMC has since been extended to more complex architectures~\cite{theus2025glmc}, multi-model settings~\cite{ito2025stemm}, model merging~\cite{xu2025prodistill,nasery2025pleas}, and weight-space generative modeling~\cite{gupta2026deepweightflow}.

Previous studies have attempted alignment in the weight space by estimating the unit correspondences layer by layer. Git Re-Basin proposed weight matching, which solves a linear assignment problem for each layer, and showed that linear interpolation is enabled by moving models into the same basin~\cite{ainsworth2023gitrebasin}. In contrast, OT4P, focusing on the fact that permutation matrices are a subset of orthogonal matrices, proposed a gradient-based framework by parameterizing orthogonal matrices using Lie algebras~\cite{guo2024ot4p}.

While effective, existing approaches face limitations for challenging targets, such as NeRF. Weight matching commits to discrete permutations early, which may propagate errors across layers. OT4P performs continuous global optimization but requires costly iterative gradient updates and can struggle in complex settings such as NeRF.

The permutation of each layer depends on its neighboring layers, revealing a clear inter-layer dependency structure. Several prior works adopted sequential layer-wise alignment strategies to stabilize optimization~\cite{wang2020federated,stoica2024zipit,xu2025prodistill,nasery2025pleas}, suggesting that alignment should proceed from input to output.

We propose progressive alignment, which advances alignment sequentially from input to output while respecting inter-layer dependencies, combined with temperature-controlled spectral annealing. At each layer, a correlation matrix is aligned to a Hungarian anchor and progressively sharpened toward a permutation matrix through spectral annealing. This bridges continuous relaxation and discrete matching without iterative optimization. 

We demonstrate that the proposed method enables reliable alignment even for Neural Radiance Fields (NeRFs)~\cite{mildenhall2020nerf}, where weights encode 3D scenes in practice. This facilitates smoother interpolation and improves the usability of NeRFs as weight-space representations~\cite{navon2024equivariant,gupta2026deepweightflow,deluigi2023inr2vec,zamaramirez2024nf2vec,ballerini2026weight}.
Synthetic permutation recovery experiments can be solved by most existing methods. However, in practical scenarios, such as independently trained NeRF models, the correlation structure becomes highly ambiguous and noisy. In such cases, early discrete decisions made by weight matching often propagate errors across layers, whereas the proposed spectral annealing method gradually sharpens alignment while preserving global consistency.

The main contributions of this study are as follows:
\begin{itemize}
\item We formulate Permutation Alignment as a sequential problem with inter-layer dependencies and propose progressive alignment, which advances the alignment step-by-step from input to output.
\item We introduce a temperature-controlled spectral projection that combines Hungarian anchors with spectral annealing, avoiding premature fixation to local optima.
\item We introduce a two-stage strategy that performs sequential layer-wise weight matching using the proposed alignment as a robust initialization, enabling highly stable optimization.
\item We apply the proposed method to NeRFs, a challenging alignment setting, and demonstrate reliable alignment and reduced loss barriers for independently trained models.
\end{itemize}

\section{Related Work}

The permutation symmetries of neural networks have been studied in model merging, mode connectivity, and weight-space learning. However, their role in INRs remains unclear. Here, we review the permutation alignment methods and INR weight representations.

\subsection{Permutation Symmetries and Mode Connectivity of Neural Networks}

Neural networks are invariant to the permutations of hidden units, resulting in exponentially many equivalent parameterizations~\cite{brea2019saddles}. This symmetry underlies the geometry of loss landscape and mode connectivity~\cite{garipov2018loss,draxler2018landscape,freeman2017topology,frankle2020linear}. Entezari et al.~\cite{entezari2022role} conjectured that appropriate permutation alignment places independently trained models within a single connected basin, enabling the LMC. This idea has since been extended to large-scale architectures and multi-model settings~\cite{theus2025glmc,ito2025stemm}, as well as training-time alignment strategies~\cite{li2025tnapfn}.

\subsection{Permutation Alignment of Neural Networks}

Representative studies that explicitly address permutation alignment include methods that estimate neuron correspondences using Optimal Transport~\cite{singh2020ot} and Git Re-Basin~\cite{ainsworth2023gitrebasin}. 
In particular, Git Re-Basin proposed weight matching, which estimates optimal permutations by constructing a correlation matrix for each layer and solving a Linear Assignment Problem. 
Because permutations are fixed discretely using the Hungarian algorithm at each update, incorrect correspondences may propagate across layers, and the results can depend on the update order and initialization.
Additionally, the layer update order is typically determined randomly, which can lead to variance in the results depending on the initial conditions and random seed.

To avoid the difficulties of discrete permutation optimization, several studies relaxed permutation matrices to continuous spaces. 
A common approach optimizes the Birkhoff polytope, which is the convex hull of permutation matrices, using doubly stochastic relaxations and Sinkhorn regularization~\cite{mena2018learning,birdal2019probabilistic}. 
In the context of a model re-basin, Sinkhorn operators combined with implicit differentiation enable gradient-based end-to-end optimization~\cite{pena2023rebasin}. 
Beyond convex relaxations, OT4P~\cite{guo2024ot4p} extends the search space to the orthogonal group by parameterizing orthogonal matrices via Lie algebras and performs gradient-based optimization on the orthogonal manifold.

Orthogonal to permutation optimization itself, it has been observed that even after alignment, linear interpolation may suffer from Variance Collapse of activation. Methods such as REPAIR~\cite{jordan2023repair} address this issue by recalibrating the activation statistics after merging.

Progressive layer-wise alignment, which advances matching from input to output, has gained attention for federated learning and model merging. 
FedMA~\cite{wang2020federated} sequentially matches and integrates neurons starting from the early layers. 
ZipIt!~\cite{stoica2024zipit} performs step-by-step merging while propagating the transformation matrices forward through the network. 
Similarly, ProDistill~\cite{xu2025prodistill} and PLeaS~\cite{nasery2025pleas} adopted progressive layer-wise distillation and alignment to improve the scalability and performance of large models. 
These works collectively indicate that respecting directional inter-layer dependencies can stabilize alignment and merging.

Spectral decomposition has also been used to resolve permutation ambiguities, for example, in transformer head matching via permutation-invariant spectra~\cite{rinaldi2025transfusion}. The eigenvalue and singular value spectra can provide symmetry-aware signals for alignment.

\subsection{Weight Representations of INRs}

INRs, such as NeRF~\cite{mildenhall2020nerf} and DeepSDF~\cite{park2019deepsdf} encode signals directly in network weights, connecting to prior work that treats weights as data~\cite{ha2017hypernetworks,schurholt2021zoo}. Methods such as inr2vec and nf2vec~\cite{deluigi2023inr2vec,zamaramirez2024nf2vec}, Graph Meta-Networks~\cite{ballerini2026weight}, and Spatial LLaNA~\cite{amaduzzi2025weights2space} learn compact representations of INR weights for downstream tasks. However, these approaches are sensitive to permutation symmetries, and misalignment often leads to high loss barriers or representation collapse~\cite{navon2024equivariant}. Symmetry-aware architectures, including graph-based processing~\cite{lim2024graph}, DEEP-ALIGN~\cite{navon2024equivariant}, and Universal Neural Functionals~\cite{zhou2024unf}, attempt to address this issue at the architectural level. Nevertheless, the stable post-hoc alignment of independently trained INRs remains challenging.

To directly address symmetry, alternative approaches design permutation-equivariant architectures. Graph-based processing of networks~\cite{lim2024graph}, inference-based permutation prediction using Deep Weight Space Networks (DEEP-ALIGN)~\cite{navon2024equivariant}, and Universal Neural Functionals (UNFs)~\cite{zhou2024unf} aim to incorporate symmetry handling into model design itself. Nevertheless, aligning independently optimized INRs in a stable and scalable manner remains challenging.

\section{Proposed Method}
\label{sec:proposed_method}

In this paper, we propose an alignment method for the weight space of neural networks. Our approach introduces a temperature parameter $\tau$ into the correspondence matrix to enable continuous relaxation and performs alignment progressively following the order of layers (Fig.~\ref{fig:proposed_overview}). 

While adopting a progressive layer-wise framework, the proposed method incorporates a temperature-controlled spectral projection from the unconstrained space of the correlation matrices to the permutation matrices. Specifically, we applied temperature scaling to the singular values of the correlation matrix. For numerical robustness, we solely rely on SVD for this spectral projection process.

This strategy enables the method to avoid the local optima that conventional approaches are prone to while exploring globally consistent alignments more efficiently than stochastic gradient-based optimization methods.

\subsection{Problem Formulation}

First, we formulate the alignment problem of the neural networks.  
Given two neural networks $\mathcal{N}_A$ and $\mathcal{N}_B$, let $\mat{W}_l^{(A)}$ and $\mat{W}_l^{(B)}$ denote the weight matrices at layers $l \in \{1, \dots, L\}$, respectively.  
Our objective is to estimate permutation matrices $\mat{P}_l$ that reorder the hidden neurons of $\mathcal{N}_B$ such that they become functionally aligned with $\mathcal{N}_A$, thereby minimizing the distance between the two models in the weight space.

The global objective function $\mathcal{L}$ is defined as the sum of the squared differences between the aligned weights at each layer:
\begin{equation}
\min_{\{\mat{P}_l\}_{l=1}^{L-1}}
\mathcal{L}
=
\sum_{l=1}^{L}
\left\|
\mat{W}_l^{(A)}
-
\mat{P}_l
\mat{W}_l^{(B)}
\mat{P}_{l-1}^\top
\right\|_F^2,
\label{eq:global_objective}
\end{equation}
where $\|\cdot\|_F$ denotes the Frobenius norm. 
For simplicity of exposition, the bias terms are omitted, but the formulation can be straightforwardly extended to incorporate them.

Each $\mat{P}_l$ is a permutation matrix, and $\mat{P}_0$ and $\mat{P}_L$, corresponding to the input and output layers, respectively, are fixed to the identity matrix $\mat{I}$.  
Because the permutation at each layer is structurally coupled with that of the adjacent layers, this optimization problem becomes a discrete combinatorial problem with exponential complexity. In general, this is NP-hard~\cite{ainsworth2023gitrebasin,guo2024ot4p}.

\subsection{Progressive Alignment Based on Correlation Maximization}

Optimizing all permutation matrices $\{\mat{P}_l\}_{l=1}^{L-1}$ jointly is computationally intractable owing to the combinatorial explosion inherent in discrete permutation searches.  
Although approaches such as OT4P~\cite{guo2024ot4p} relax parameters to continuous values and apply gradient-based optimization, they may fail to produce satisfactory results in challenging alignment scenarios, particularly for models such as NeRF.
To balance computational efficiency and optimization stability, we adopt a progressive optimization strategy that decomposes the problem into layer-wise subproblems and solves them iteratively.

When updating the permutation matrix $\mat{P}_l$ at layer $l$, fixing the adjacent permutation matrices $\mat{P}_{l-1}$ and $\mat{P}_{l+1}$ reduces the number of terms in the objective function (Eq.~\ref{eq:global_objective}) that depend on $\mat{P}_l$, yielding the following correlation maximization problem:
\begin{equation}
\max_{\mat{P}_l} \Tr(\mat{P}_l^\top \mat{C}_l),
\label{eq:sub_problem}
\end{equation}
where the structure of the correlation matrix $\mat{C}_l$ reflects the neuron-wise alignment strength. Under fixed neighboring permutations, $\mat{C}_l$ is computed from the current weights as follows:
\begin{equation}
\mat{C}_l
=
\mat{W}_{l}^{(A)} (\mat{P}_{l-1} \mat{W}_{l}^{(B)})^\top
+
(\mat{W}_{l+1}^{(A)})^\top \mat{P}_{l+1} \mat{W}_{l+1}^{(B)}.
\end{equation}
The derivation is provided in the Supplementary Material.
Intuitively, $\mat{C}_l$ represents the local consistency at layer $l$ based on its connections to the preceding and succeeding layers.  
Its spectral structure reflects the distribution of the alignment strength within layer $l$. 
Solving this optimization problem corresponds to reordering the correspondences at layer $l$ to simultaneously maximize consistency with both adjacent layers.

In contrast, Git Re-Basin~\cite{ainsworth2023gitrebasin} solves this subproblem in a random layer order by addressing a linear assignment problem and immediately updates to a discrete permutation matrix,
\begin{equation}
\mat{P}_l = \argmax_{\mat{P}\in\mathcal{P}_n} \Tr(\mat{P}^\top \mat{C}_l).
\end{equation}
However, this constitutes a premature and irreversible decision. As a result, incorrect correspondences determined in the early stages may propagate to subsequent layers, potentially causing the entire network to converge to a poor local optimum.  
(A concrete numerical example on a small-scale case is provided in the Supplementary Material.)

\subsection{Temperature-Controlled Spectral Annealing}

\begin{figure}[tb]
\centering
\includegraphics[width=0.9\linewidth]{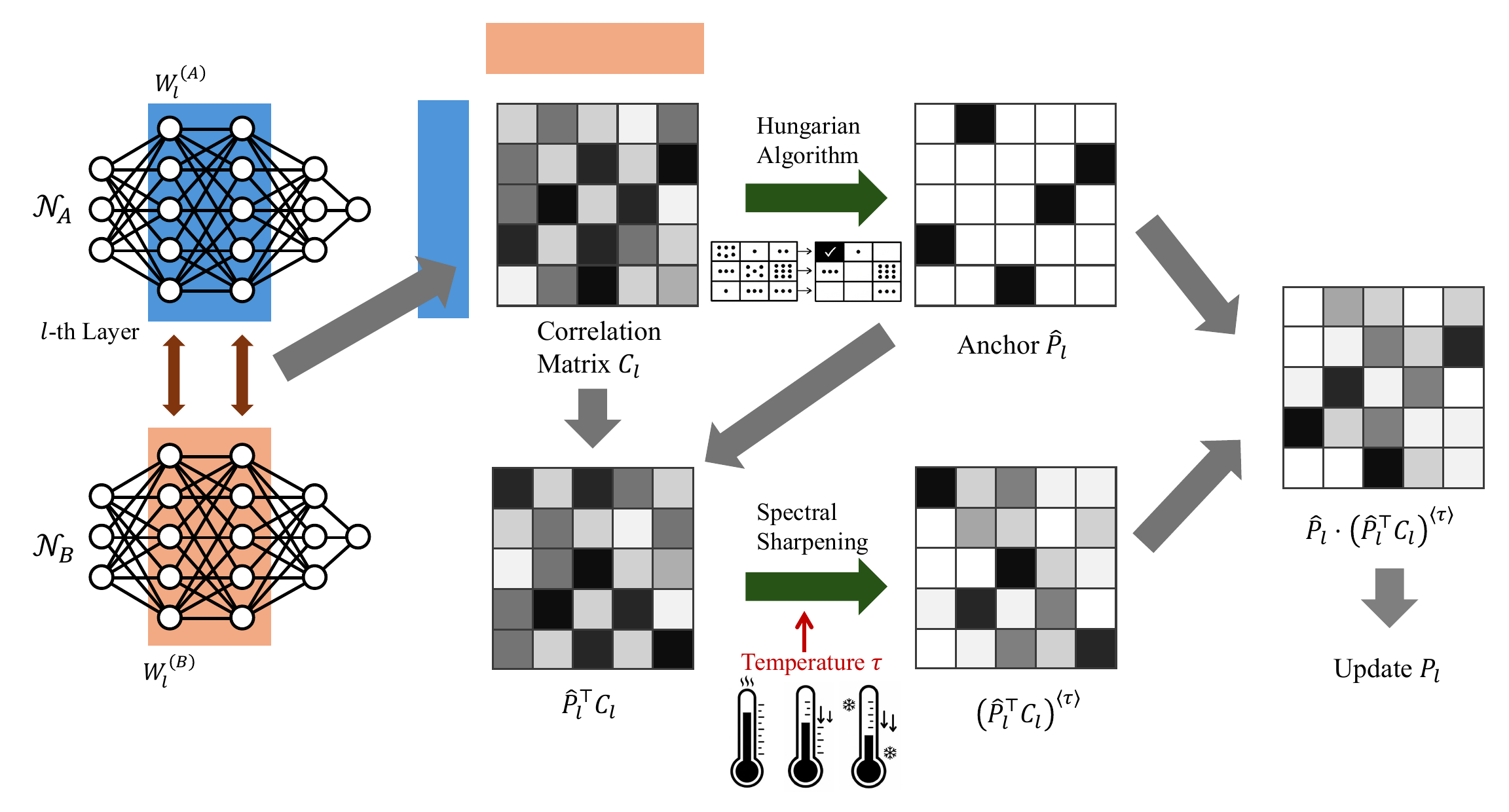} 
\caption{Temperature-Controlled Spectral Sharpening for Layer-wise Permutation Update.
At layer $l$, given weights $W_l^{(A)}$ and $W_l^{(B)}$, a correlation matrix $\mat{C}_l$ that captures the neuron-wise similarity under fixed neighboring permutations is computed.
The Hungarian algorithm first produces an anchor permutation $\hat{\mat{P}}_l$ that provides a discrete reference alignment.
Temperature-controlled spectral sharpening is applied to the aligned correlation matrix $\hat{\mat{P}}_l^\top \mat{C}_l$, where spectral values are exponentiated by a temperature parameter $\tau$.}
\vspace{-3mm}
\label{fig:update_p}
\end{figure}

Inspired by the OT4P approach~\cite{guo2024ot4p}, we address the aforementioned local optimum issue by proposing an update rule based on temperature-controlled spectral annealing. Starting with the correlation matrix $\mat{C}_l$, the proposed method gradually drives the solution toward a permutation matrix.
An overview of the update procedure is shown in Fig.~\ref{fig:update_p}, and the complete algorithm is presented in Alg.~\ref{alg:proposed}.  
The key feature of our method is that instead of relying on gradient-based optimization, it directly controls the spectral structure of the correlation matrix. By adjusting the temperature parameter, the solution is progressively sharpened toward a discrete permutation matrix in a deterministic manner.

Our method starts with the correlation matrix $\mat{C}_l \in \mathbb{R}^{n \times n}$ at each layer.  
First, we solve the linear assignment problem
\begin{equation}
\hat{\mat{P}}_l
=
\argmax_{\mat{P} \in \mathcal{P}_n}
\Tr(\mat{P}^\top \mat{C}_l),
\end{equation}
using the Hungarian algorithm to obtain an anchor permutation matrix $\hat{\mat{P}}_l$ that is most consistent with the correlation structure.
Next, we align the correlation matrix with the anchor permutation,
\begin{equation}
\mat{M}_l
=
\hat{\mat{P}}_l^\top \mat{C}_l,
\end{equation}
and apply temperature-controlled spectral annealing to $\mat{M}_l$.
Let the singular value decomposition (SVD) of the matrix $\mat{M} \in \mathbb{R}^{n \times n}$ be given by
$\mat{M} = \mat{U} \boldsymbol{\Sigma} \mat{V}^\top$, where $\boldsymbol{\Sigma} = \mathrm{diag}(\sigma_1, \dots, \sigma_n)$ is a diagonal matrix of singular values.

In this study, we define a spectral operator that applies a temperature parameter $\tau \in [0,1]$ to the singular values as
$
\mat{M}^{\langle \tau \rangle}
=
\mat{U}
\boldsymbol{\Sigma}^{\tau}
\mat{V}^\top$ where $\boldsymbol{\Sigma}^{\tau}
=
\mathrm{diag}(\sigma_1^{\tau}, \dots, \sigma_n^{\tau})$.

This operation preserves the orthogonal bases while applying the spectral function $\sigma \mapsto \sigma^\tau$ to the singular values. As $\tau$ decreases, the singular spectrum is flattened, effectively projecting the unconstrained correlation matrix toward the orthogonal group. By strictly employing SVD, our approach ensures numerical stability and avoids the complex values that can arise when computing the eigendecomposition of asymmetric matrices.

To ensure a smooth transition from the continuous spectral relaxation to a discrete permutation, we introduce a linear interpolation mechanism. 
Using the obtained anchor $\hat{\mat{P}}_l$ and the original correlation matrix $\mat{C}_l$, we update the permutation matrix by linearly interpolating between the spectrally annealed correlation matrix and the identity matrix according to the current temperature parameter $\tau_k \in \mathcal{T}$, as follows:
\begin{equation}
\mat{P}_l^{(\text{new})}
=
\hat{\mat{P}}_l
\left[
\tau_k
\left(
\hat{\mat{P}}_l^\top \mat{C}_l
\right)^{\langle \tau_k \rangle}
+
(1 - \tau_k) \mat{I}
\right].
\label{eq:projection_update}
\end{equation}
Here, $\mathcal{T} = \{\tau_1, \dots, \tau_K\}$ denotes the temperature schedule that decreases from $1.0$ to $0.0$, and $\mat{I}$ represents the identity matrix.
This linear interpolation explicitly anchors the continuous relaxation. As $\tau_k$ approaches $0$, the matrix inside the bracket converges to the identity matrix $\mat{I}$, forcing the overall update $\mat{P}_l^{(\text{new})}$ to smoothly and deterministically land on the discrete anchor permutation $\hat{\mat{P}}_l$. This mechanism prevents the exploration from remaining overly broad in the continuous unconstrained space.

By iteratively performing progressive alignment along the layer order while gradually decreasing $\tau$, the correlation matrix $\mat{C}_l$ becomes increasingly concentrated around the anchor permutation $\hat{\mat{P}}_l$, tending toward a discrete correspondence.
Because our method performs only a single update per temperature value, it does not require iterative gradient-based optimization as in OT4P, enabling fast and deterministic estimation of correspondences.
During this annealing process, the continuous structure of the correlation matrix is preserved in the early stages, allowing the exploration of global consistency. As $\tau$ decreases, the high-confidence modes are progressively emphasized, ultimately leading to convergence to discrete neuron correspondences.

\label{supp:alg}
\begin{algorithm}[t]
\caption{Progressive Alignment with Temperature-Controlled Spectral Sharpening}
\label{alg:proposed}
\scriptsize
\begin{algorithmic}[1]
\REQUIRE Models $\mathcal{N}_A, \mathcal{N}_B$, Temperature schedule $\mathcal{T}$, Max iterations $M$
\STATE Initialize $\mat{P}_l = \mat{I}$ for all layers $l$
\STATE Generate schedule $\mathcal{T} = \{\tau_1, \dots, \tau_K\}$ from $1.0$ down to $0.0$
\FOR{$\tau \in \mathcal{T}$}
    \FOR{Layer iteration $m = 1$ to $M$} 
        \FOR{layer $l = 1$ to $L-1$}
            \STATE Compute correlation matrix $\mat{C}_l$ using current $\mat{P}_{l-1}, \mat{P}_{l+1}$
            \STATE Obtain discrete anchor: $\hat{\mat{P}}_l \leftarrow \operatorname{Hungarian}(\mat{C}_l)$
            \STATE Align correlation matrix: $\mat{M}_l \leftarrow \hat{\mat{P}}_l^\top \mat{C}_l$
            \STATE Compute SVD: $\mat{U}, \boldsymbol{\Sigma}, \mat{V}^\top \leftarrow \operatorname{SVD}(\mat{M}_l)$
            \STATE Apply spectral flattening: $\mat{M}_l^{\langle \tau \rangle} \leftarrow \mat{U}\boldsymbol{\Sigma}^{\tau}\mat{V}^\top$
            \STATE Update with anchored continuous relaxation: $\mat{P}_l \leftarrow \hat{\mat{P}}_l \left[ \tau \mat{M}_l^{\langle \tau \rangle} + (1 - \tau) \mat{I} \right]$
        \ENDFOR
    \ENDFOR
\ENDFOR
\RETURN Aligned model $\mathcal{N}_B'$ permuted by $\{\mat{P}_l\}$
\end{algorithmic}
\end{algorithm}

\section{Experiment}
\label{sec:experiments}

In this section, we present two experimental results using traditional evaluation of neural network alignment and NeRF alignment to evaluate the alignment capability and effectiveness of the proposed method. All experiments were conducted on a computer with an NVIDIA GeForce RTX 5090 GPU.

\subsection{Comparison Methods}
\label{sec:comparison_methods}
To evaluate the performance of the proposed method, we compared it with the following approaches.
\begin{enumerate}
\item \textbf{Naive Interpolation:} Linear interpolation is performed directly on the weights without any permutation alignment.
\item \textbf{Weight Matching~\cite{ainsworth2023gitrebasin}:}  
The neuron alignment problem is formulated as a linear assignment problem and solved using greedy weight matching.
\item \textbf{Progressive Weight Matching:}  
In contrast to the original weight matching~\cite{ainsworth2023gitrebasin}, which selects layers in random order for optimization, this variant optimizes the permutations sequentially from the input layer toward the output layer.
\item \textbf{OT4P~\cite{guo2024ot4p}:}  
This method relaxes the discreteness of permutation matrices and performs optimization in a continuous space using stochastic gradient descent (SGD).
\item \textbf{Progressive OT4P~\cite{guo2024ot4p}:}  
A modified version of standard OT4P normally optimizes all layers jointly, where gradient-based optimization is performed sequentially from the input side of the network to the output.
\item \textbf{Ours w/o weight matching refinement:}  
An ablated version of our method that relies solely on correlation-based relaxation and temperature-controlled progressive projection, omitting the final refinement stage.
\item \textbf{Ours:} 
Our complete two-stage approach. It utilizes the continuously aligned solution obtained from the spectral projection as a robust initialization, followed by iterative layer-wise weight matching to achieve highly stable optimization.
\end{enumerate}

For weight matching, random correspondence estimation was performed 10 times using shuffled layer orders covering all layers. 
For progressive variants, 100 sequential sweeps were conducted. 
For the proposed method, the temperature schedule was set to $\mathcal{T}=\{1.0, 0.5, 0.0\}$ and a single sweep was performed for each temperature value, followed by a final sequential layer-wise refinement.
For OT4P-based optimization, we used AdamW~\cite{loshchilov2019adamw} as the optimizer and terminated the optimization if no improvement was observed for 10 consecutive steps. 
The learning rate was set to $0.05$ for OT4P.

\subsection{Evaluation of Alignment Accuracy on MLPs and CNNs}

\begin{table}[t]
    \centering
    \caption{$L_1$ distance (converted by $\log(1+x)$) and Precision (\%) of alignment algorithms across standard architectures (MLP, CNN)}
    \label{tab:alignment_results_cnn}
    \resizebox{\linewidth}{!}{%
    \begin{tabular}{lcccccc}
        \toprule
        \multirow{2}{*}{\textbf{Algorithm}} & \multicolumn{2}{c}{\textbf{MLP5}} & \multicolumn{2}{c}{\textbf{VGG11}} & \multicolumn{2}{c}{\textbf{ResNet18}} \\
        \cmidrule(lr){2-3} \cmidrule(lr){4-5} \cmidrule(lr){6-7}
        & $L_1$ Dist. & Precision & $L_1$ Dist. & Precision & $L_1$ Dist. & Precision \\
        \midrule
        Weight Matching                 & $0.000$ {\tiny $\pm 0.000$} & $100.0$ {\tiny $\pm 0.000$} & $0.000$ {\tiny $\pm 0.000$} & $100.0$ {\tiny $\pm 0.000$} & $0.000$ {\tiny $\pm 0.000$} & $100.0$ {\tiny $\pm 0.000$} \\
        Progressive Weight Matching     & $0.000$ {\tiny $\pm 0.000$} & $100.0$ {\tiny $\pm 0.000$} & $0.000$ {\tiny $\pm 0.000$} & $100.0$ {\tiny $\pm 0.000$} & $0.000$ {\tiny $\pm 0.000$} & $100.0$ {\tiny $\pm 0.000$} \\
        OT4P ($\tau=0.7$)               & $0.000$ {\tiny $\pm 0.000$} & $100.0$ {\tiny $\pm 0.000$} & $6.223$  {\tiny $\pm 0.271$} & $99.74$ {\tiny $\pm 0.033$} & $6.376$ {\tiny $\pm 0.298$} & $99.29$ {\tiny $\pm 0.090$} \\
        OT4P ($\tau=0.5$)               & $0.000$ {\tiny $\pm 0.000$} & $100.0$ {\tiny $\pm 0.000$} & $7.075$ {\tiny $\pm 0.201$} & $99.64$ {\tiny $\pm 0.036$} & $5.768$ {\tiny $\pm 0.273$} &
        $99.45$ {\tiny $\pm 0.065$} \\
        OT4P ($\tau=0.3$)               & $0.000$ {\tiny $\pm 0.000$} & $100.0$ {\tiny $\pm 0.000$} & $6.746$ {\tiny $\pm 0.225$}   & $99.74$ {\tiny $\pm 0.029$} & $5.820$ {\tiny $\pm 0.611$} & $99.42$ {\tiny $\pm 0.135$} \\
        Progressive OT4P($\tau=0.7$)    & $8.448$ {\tiny $\pm 0.096$} & $66.27$ {\tiny $\pm 4.331$} & $8.690$ {\tiny $\pm 0.284$} & $99.15$ {\tiny $\pm 0.160$} & $11.58$ {\tiny $\pm 0.006$} & $60.37$ {\tiny $\pm 1.552$} \\
        Progressive OT4P($\tau=0.5$)    & $8.705$ {\tiny $\pm 0.031$} & $51.76$ {\tiny $\pm 2.286$} & $9.722$ {\tiny $\pm 0.928$} & $97.58$ {\tiny $\pm 1.999$} & $11.59$ {\tiny $\pm 0.003$} & $58.06$ {\tiny $\pm 1.131$} \\
        Progressive OT4P($\tau=0.3$)    & $8.897$ {\tiny $\pm 0.006 $} & $34.2$ {\tiny $\pm 0.608$} & $12.28$ {\tiny $\pm 0.001$} & $74.88$ {\tiny $\pm 0.045$} & $11.61$ {\tiny $\pm 0.005$} & $54.53$ {\tiny $\pm 0.489$} \\
        Proposed (Ours) w/o Ref.               & $0.000$ {\tiny $\pm 0.000$} & $100.0$ {\tiny $\pm 0.000$} & $0.000$ {\tiny $\pm 0.000$} & $100.0$ {\tiny $\pm 0.000$} & $0.000$ {\tiny $\pm 0.000$} & $100.0$ {\tiny $\pm 0.000$} \\
        Proposed (Ours)  & $0.000$ {\tiny $\pm 0.000$} & $100.0$ {\tiny $\pm 0.000$} & $0.000$ {\tiny $\pm 0.000$} & $100.0$ {\tiny $\pm 0.000$} & $0.000$ {\tiny $\pm 0.000$} & $100.0$ {\tiny $\pm 0.000$}\\
        \bottomrule
    \end{tabular}
}
\end{table}

Following prior work~\cite{guo2024ot4p}, we evaluated whether the proposed method can accurately recover the correct neuron correspondences. As evaluation targets, we used a simple 5-layer MLP (with randomly initialized weights) and standard VGG11 and ResNet18 architectures (with the official ImageNet-pretrained weights provided by torchvision). 
We construct Model B by randomly permuting the intermediate weights of a reference Model A (Ground Truth), preserving the network's input-output function. Each method is then applied to align Model B back to Model A.
To account for permutation variability, experiments were conducted five times using different random seeds, and we report the mean and standard deviation.

\noindent \textbf{Evaluation Metrics} 
To quantitatively evaluate the alignment performance, we followed prior work~\cite{guo2024ot4p} and adopted the following metrics:
\begin{itemize}
    \item \textbf{\(L_1\) Distance:} The \(L_1\) distance between models in the weight space of neural networks. A smaller value indicates that the original weight configuration has been recovered more accurately.
    \item \textbf{Precision:} In the single-model shuffle recovery experiment, the proportion of elements in the estimated permutation matrices that exactly match the Ground Truth. A higher value indicates a more accurate identification of the original correct permutations.
\end{itemize}

Table~\ref{tab:alignment_results_cnn} shows the recovery accuracy and post-alignment L1 distance. On the small MLP5, OT4P recovers the permutations exactly at every temperature. On both CNNs it still recovers more than 99\% of the units, but it does not reach exact recovery, and a residual misalignment remains in the L1 distance. Its cost also grows with the layer width: on VGG11, whose fully connected layers have 4096 units, each run takes about 40 minutes because of gradient optimization over large orthogonal manifolds. Progressive OT4P is considerably less reliable. It optimizes each layer only once while the subsequent layers are still unaligned, so errors made in earlier layers are never corrected, and its precision varies widely with the network and $\tau$ (34--66\% on MLP5, 75--99\% on VGG11, and 55--60\% on ResNet18). Even at its best temperature, it does not achieve exact recovery on any of the three networks. This sensitivity makes it difficult to choose an appropriate temperature in advance.

In contrast, both weight matching and our method perfectly recovered the shuffled weights (Precision 100\%, \(L_1\) distance $\approx 0$) regardless of network depth or layer dimensionality, completing the alignment efficiently without iterative gradient updates (detailed runtime comparisons are provided in the Supplementary Material).

\subsection{Alignment and LMC Evaluation on NeRF}

While previous studies on neural network alignment have primarily focused on MLP and CNN, this study extends permutation alignment to Neural Radiance Fields (NeRF), a 3D representation model, for the first time and evaluates its effectiveness in this setting.

\subsubsection{Evaluation on Random Permutation Recovery in NeRF}

We first evaluated the ability to identify correct correspondences in NeRF using the same experimental protocol as in the previous section.  
We used the NeRF Synthetic dataset (eight scenes)~\cite{mildenhall2020nerf} and performed optimization with the NeRF implementation provided by \texttt{nerfstudio}.  
A pretrained reference model (Model A) and a randomly permuted version (Model B) are prepared.  
As in the MLP and CNN alignment experiments, Model B was aligned back to Model A using each method, and the accuracy of permutation recovery was evaluated.

Table~\ref{tab:alignment_results_nerf} summarizes the results of the single-model permutation recovery experiment using NeRF.  
OT4P was sensitive to the temperature: it recovered about 99\% of the units at $ \tau=0.5$ and $0.7$ but fell into poor local optima at $	\tau=0.3$ (Precision 14.55\%), and Progressive OT4P failed at every temperature (Precision $\leq$6.3\%).
In contrast, both the proposed method and weight matching successfully restored the original weight configuration with 100\% precision (\(L_1\) distance $\approx 0$), demonstrating reliable alignment capability, even for 3D representation models such as NeRF.

\subsubsection{Alignment of NeRF Models Trained from Different Initializations}

\begin{table}[t]
\centering
    \caption{$L_1$ distance (converted by $\log(1+x)$) and Precision (\%) on NeRF}
    \label{tab:alignment_results_nerf}
    \begin{tabular}{lcc}
        \toprule
        \textbf{Algorithm} & $L_1$ Dist. & Precision \\
        \midrule
        Weight Matching                 & 0.000   & 100.0   \\
        Progressive Weight Matching     & 0.000   & 100.0     \\
        OT4P ($\tau=0.3$)               & 12.76   & 14.55     \\
        OT4P ($\tau=0.5$)               & 6.818   & 99.16     \\
        OT4P ($\tau=0.7$)               & 6.720   & 99.21     \\
        Progressive OT4P ($\tau=0.3$)   & 12.92   & 1.455  \\
        Progressive OT4P ($\tau=0.5$)   & 12.91   & 3.657  \\
        Progressive OT4P ($\tau=0.7$)   & 12.90   & 6.259  \\
        Proposed (Ours) w/o Ref.                & 0.000   & 100.0  \\
        Proposed (Ours)   & 0.000   & 100.0     \\
        \bottomrule
    \end{tabular}
\end{table}

\begin{table}[t]
\centering
    \caption{Quantitative comparison of interpolated aligned models at $\lambda=0.5$}
    \label{tab:quantitative_results}
    \begin{tabular}{lcc}
    \toprule
    \textbf{Method} & \textbf{RMSE} $\downarrow$ & \textbf{PSNR} $\uparrow$ \\
    \midrule
    Naive                          & 0.491 & 6.205 \\
    Weight Matching                & 0.294 & 11.07 \\
    Progressive Weight Matching    & 0.306 & 10.64 \\
    OT4P ($\tau=0.3$)               & 0.489 & 6.239 \\
    OT4P ($\tau=0.5$)               & 0.495 & 6.118 \\
    OT4P ($\tau=0.7$)               & 0.494 & 6.160 \\
    Progressive OT4P ($\tau=0.3$)   & 0.493 & 6.170 \\
    Progressive OT4P ($\tau=0.5$)   & 0.493 & 6.167 \\
    Progressive OT4P ($\tau=0.7$)   & 0.492 & 6.182 \\
    Proposed (Ours) w/o Ref.               & 0.199 & 14.22 \\
    Proposed (Ours)  & 0.199 & 14.40 \\
    \bottomrule
\end{tabular}
\end{table}

\begin{figure}[t]
\centering
\resizebox{0.99\columnwidth}{!}{%
\setlength{\tabcolsep}{2pt}
\begin{tabular}{lccccc|c}
\toprule
& $\lambda=0.0$ & $\lambda=0.25$ & $\lambda=0.50$ & $\lambda=0.75$ & $\lambda=1.0$ & Ground Truth \\
\midrule
\rowlabel{(a)}
& \includegraphics[width=0.12\textwidth]{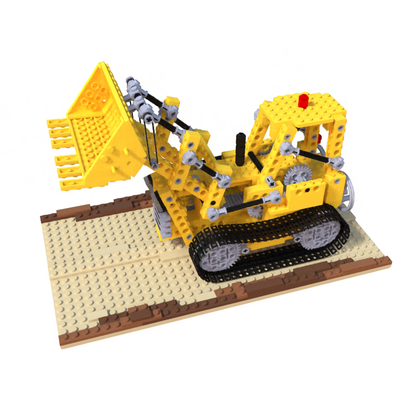}
& \includegraphics[width=0.12\textwidth]{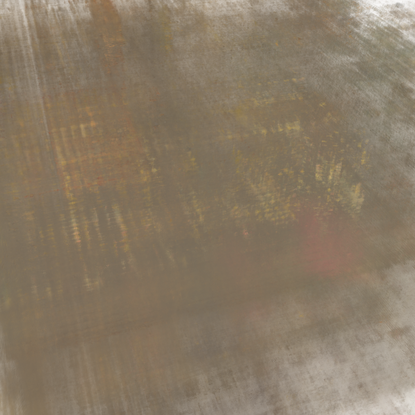}
& \includegraphics[width=0.12\textwidth]{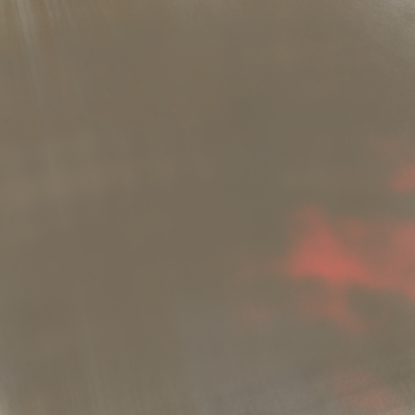}
& \includegraphics[width=0.12\textwidth]{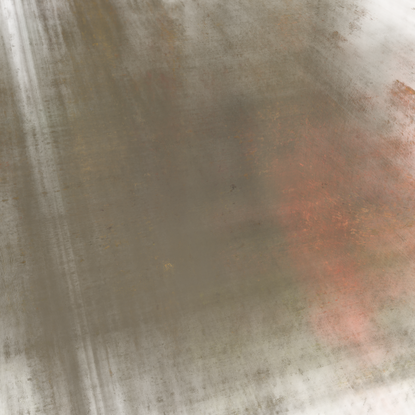}
& \includegraphics[width=0.12\textwidth]{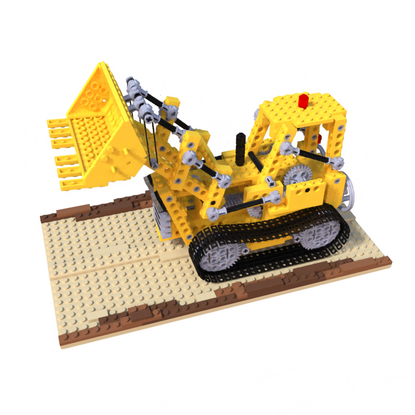}
& \includegraphics[width=0.12\textwidth]{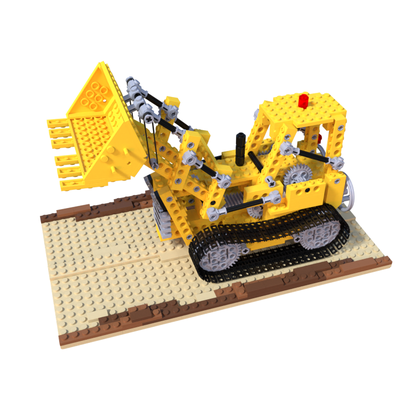} \\
\rowlabel{(b)}
& \includegraphics[width=0.12\textwidth]{figures/rendered_images/naive/lambda_0.0.png}
& \includegraphics[width=0.12\textwidth]{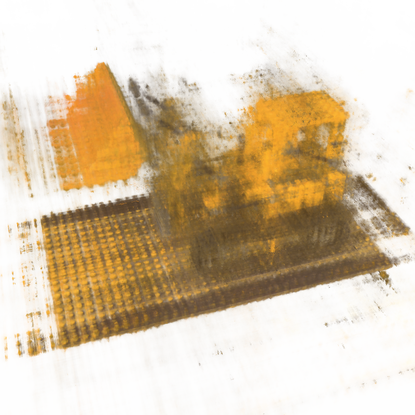}
& \includegraphics[width=0.12\textwidth]{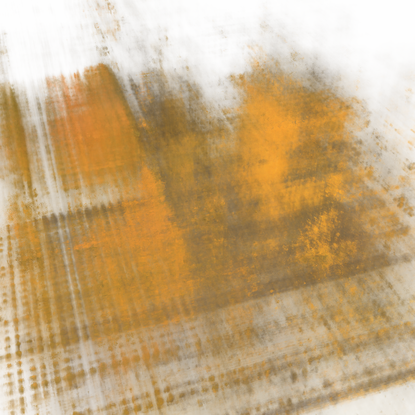}
& \includegraphics[width=0.12\textwidth]{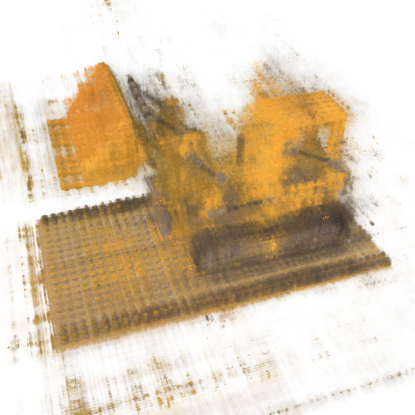}
& \includegraphics[width=0.12\textwidth]{figures/rendered_images/naive/lambda_1.0.png}
& \includegraphics[width=0.12\textwidth]{figures/rendered_images/GT.png} \\
\rowlabel{(c)}
& \includegraphics[width=0.12\textwidth]{figures/rendered_images/naive/lambda_0.0.png}
& \includegraphics[width=0.12\textwidth]{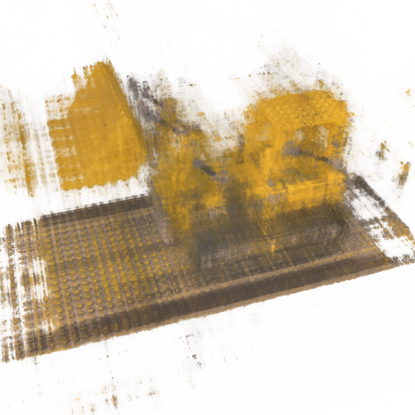}
& \includegraphics[width=0.12\textwidth]{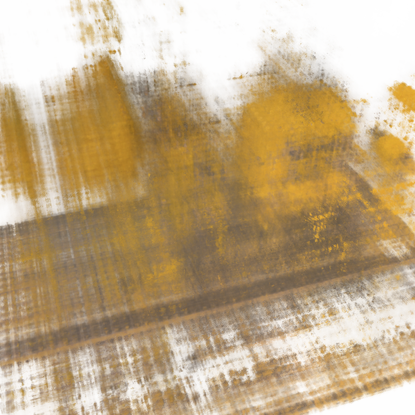}
& \includegraphics[width=0.12\textwidth]{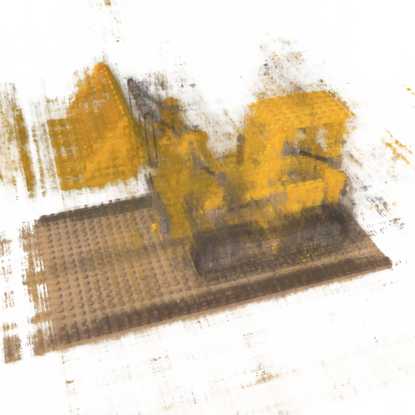}
& \includegraphics[width=0.12\textwidth]{figures/rendered_images/naive/lambda_1.0.png}
& \includegraphics[width=0.12\textwidth]{figures/rendered_images/GT.png} \\
\rowlabel{(d)}
& \includegraphics[width=0.12\textwidth]{figures/rendered_images/naive/lambda_0.0.png}
& \includegraphics[width=0.12\textwidth]{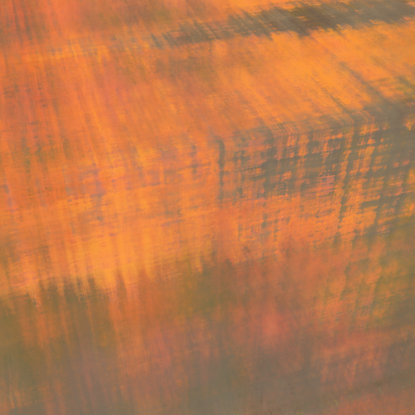}
& \includegraphics[width=0.12\textwidth]{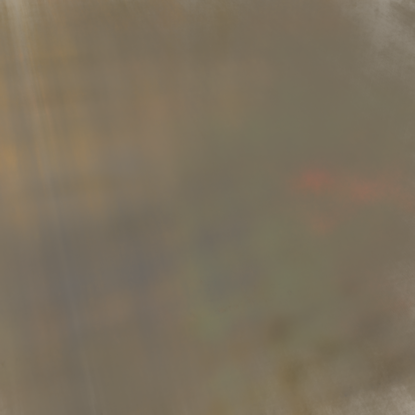}
& \includegraphics[width=0.12\textwidth]{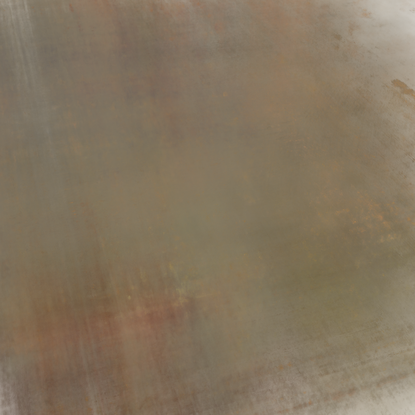}
& \includegraphics[width=0.12\textwidth]{figures/rendered_images/naive/lambda_1.0.png}
& \includegraphics[width=0.12\textwidth]{figures/rendered_images/GT.png} \\
\rowlabel{(e)}
& \includegraphics[width=0.12\textwidth]{figures/rendered_images/naive/lambda_0.0.png}
& \includegraphics[width=0.12\textwidth]{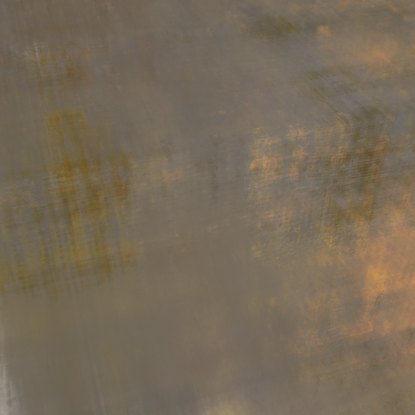}
& \includegraphics[width=0.12\textwidth]{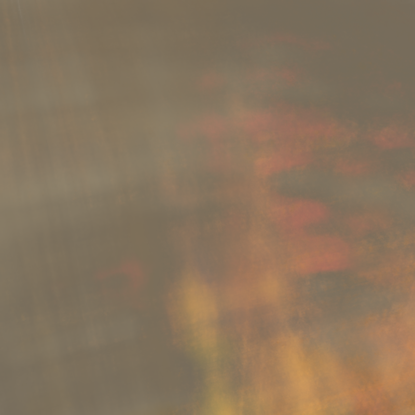}
& \includegraphics[width=0.12\textwidth]{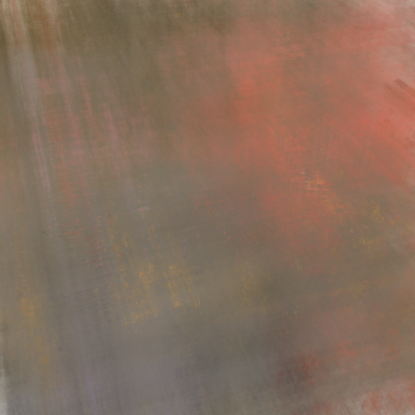}
& \includegraphics[width=0.12\textwidth]{figures/rendered_images/naive/lambda_1.0.png}
& \includegraphics[width=0.12\textwidth]{figures/rendered_images/GT.png} \\
\rowlabel{(f)}
& \includegraphics[width=0.12\textwidth]{figures/rendered_images/naive/lambda_0.0.png}
& \includegraphics[width=0.12\textwidth]{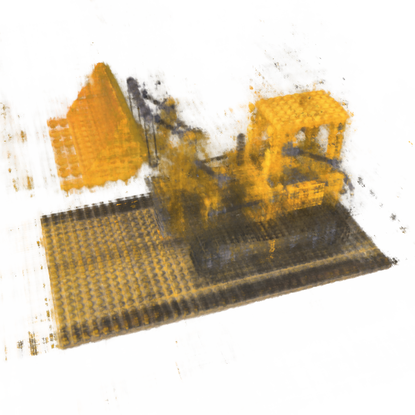}
& \includegraphics[width=0.12\textwidth]{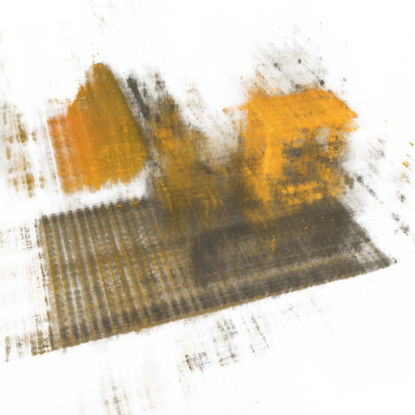}
& \includegraphics[width=0.12\textwidth]{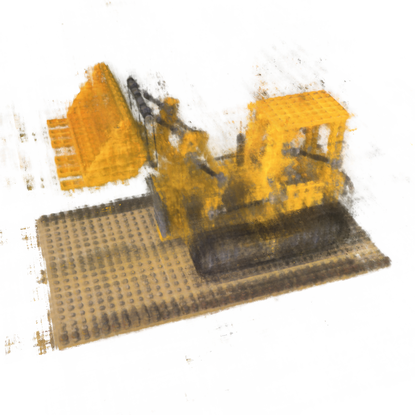}
& \includegraphics[width=0.12\textwidth]{figures/rendered_images/naive/lambda_1.0.png}
& \includegraphics[width=0.12\textwidth]{figures/rendered_images/GT.png} \\
\rowlabel{(g)}
& \includegraphics[width=0.12\textwidth]{figures/rendered_images/naive/lambda_0.0.png}
& \includegraphics[width=0.12\textwidth]{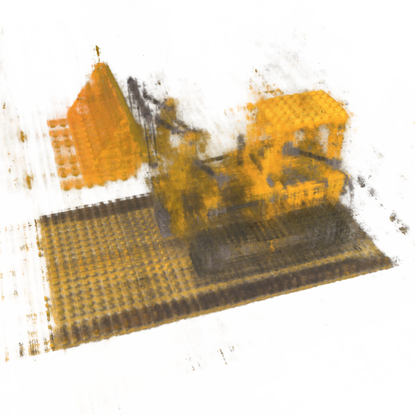}
& \includegraphics[width=0.12\textwidth]{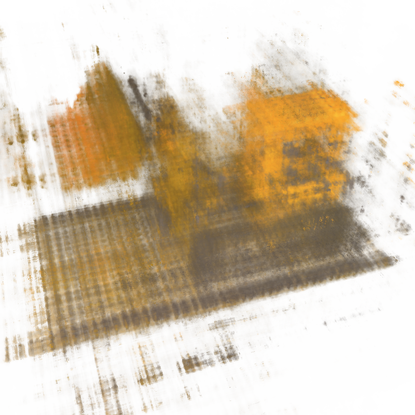}
& \includegraphics[width=0.12\textwidth]{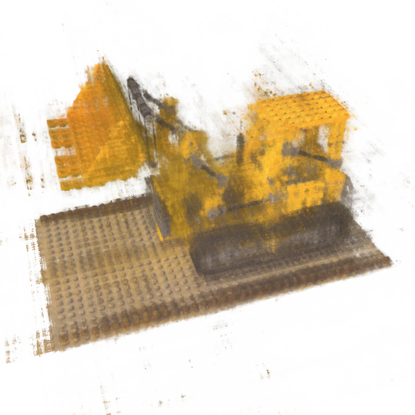}
& \includegraphics[width=0.12\textwidth]{figures/rendered_images/naive/lambda_1.0.png}
& \includegraphics[width=0.12\textwidth]{figures/rendered_images/GT.png} \\
\bottomrule
\end{tabular}
}
\caption{Example of interpolation results for different $\lambda$ values. (a) Naive, (b) Weight Matching, (c) Progressive Weight Matching, (d) OT4P, (e) Progressive OT4P, (f) Proposed (w/o weight matching refinement), and (g) Proposed.}
\label{fig:lego_extended}
\end{figure}

As a realistic and challenging alignment task for NeRF, we evaluated whether the proposed method can mitigate the loss barrier between models independently trained from different random initializations.  
In this setting, ground-truth permutation is unavailable. Therefore, the evaluation was conducted by comparing rendered images from linearly interpolated models with the corresponding ground-truth images.
We generated interpolated models in the parameter space as follows:
\begin{equation}
\mat{W}_{\lambda} = (1-\lambda)\mat{W}_A + \lambda \mat{W}_{B'},
\end{equation}
where $\mat{W}_{B'}$ denotes the weights of Model B after alignment.  
The interpolation coefficient $\lambda$ was varied from $0.0$ to $1.0$ in increments of $0.1$, and the rendering quality along the interpolation path was evaluated.
In this experiment, we considered a pair of random seeds.  
For each scene, 10 images were randomly sampled from the test views for evaluation.  
The reported results correspond to the mean and variance computed over 10 test views for each of the eight scenes.
In this experiment, we used $\tau=0.5$ for OT4P and Progressive OT4P.

\noindent \textbf{Evaluation Metrics:}
\begin{itemize}
\item \textbf{RMSE (Root Mean Square Error):} The root mean square error between pixel values of the ground-truth image and the rendered image. A smaller value indicates a lower rendering error.
\item \textbf{PSNR (Peak Signal-to-Noise Ratio):} A standard metric for evaluating rendering quality. A higher value indicates better image quality of the interpolated model.
\end{itemize}

\begin{figure}[t]
\centering
\begin{minipage}{0.42\columnwidth}
    \centering
    \includegraphics[width=\linewidth]{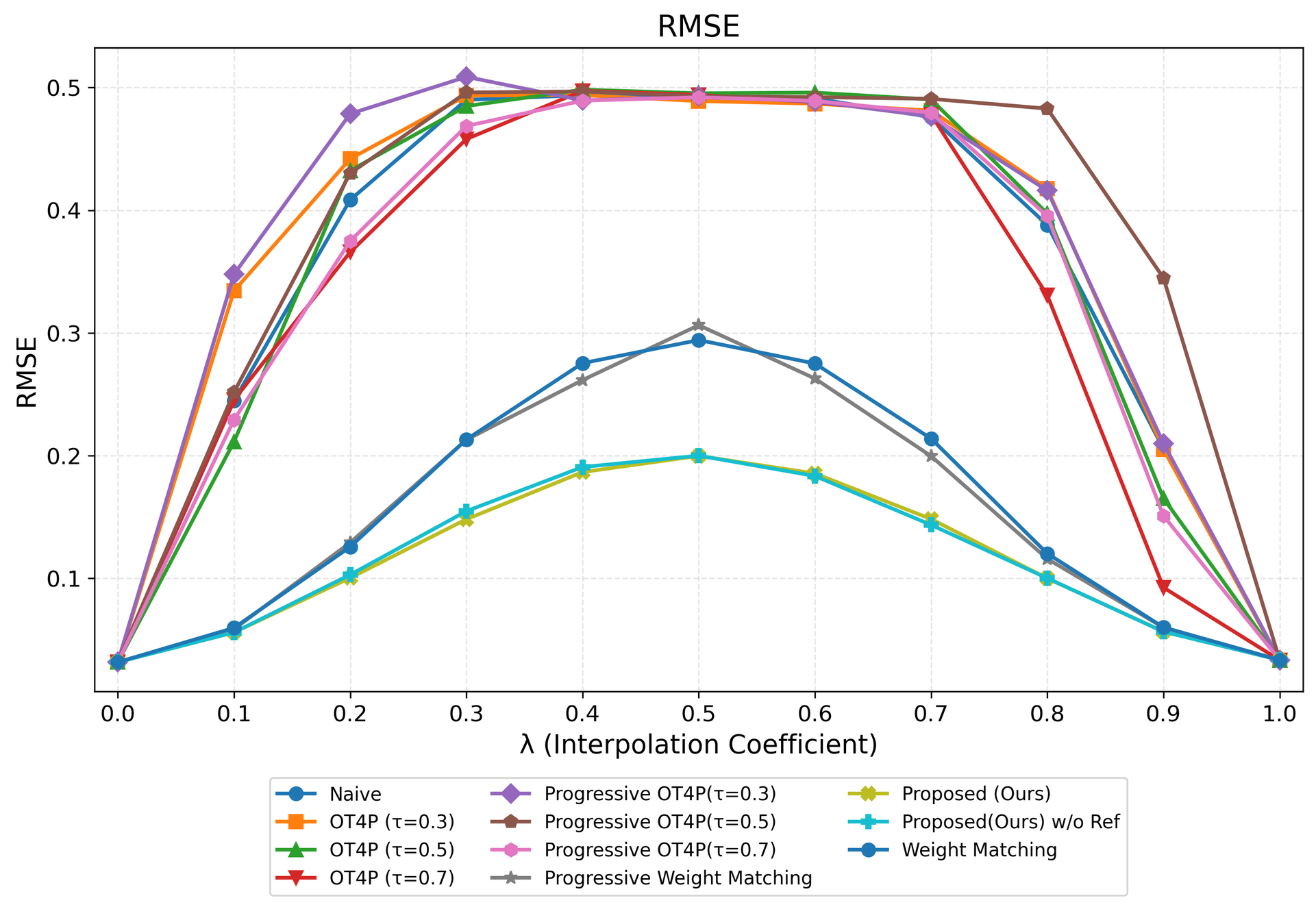}
    \caption*{(a) RMSE $\downarrow$}
\end{minipage}
\hfill
\begin{minipage}{0.42\columnwidth}
    \centering
    \includegraphics[width=\linewidth]{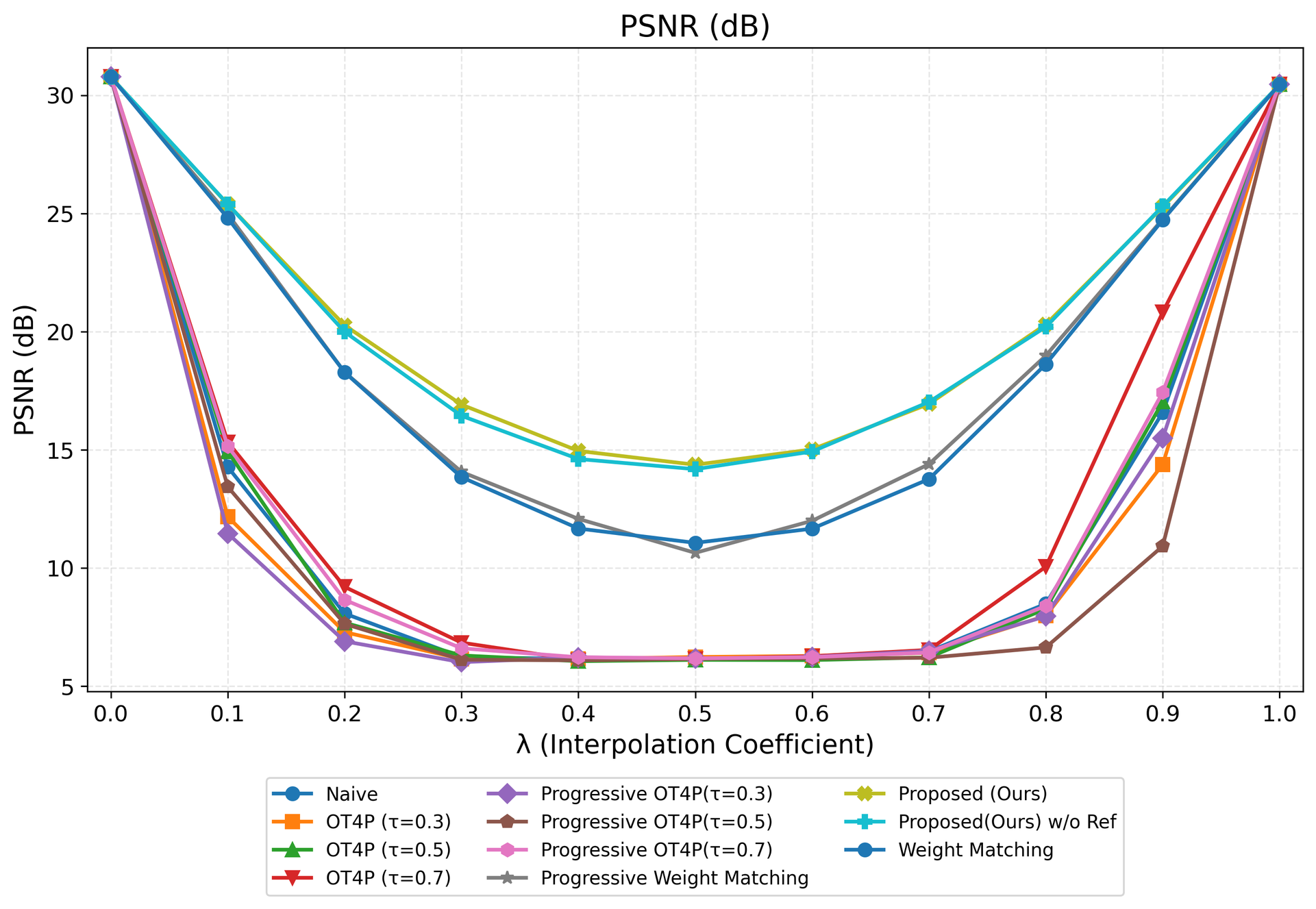}
    \caption*{(b) PSNR $\uparrow$}
\end{minipage}
\caption{Quantitative comparison across different metrics.}
\label{fig:lmc_result}
\end{figure}

Figure~\ref{fig:lmc_result} shows the PSNR and RMSE along the interpolated weight rendering results.
The computational time is presented in the Supplementary Material.
Naive Interpolation exhibits a sharp quality drop around $\lambda=0.5$, indicating that the two models lie in distinct basins.  
Existing alignment methods partially reduce this barrier, but still degrade the midpoint.
In contrast, the proposed method substantially suppressed the loss barrier.  
Although slight blurring remains at $\lambda=0.5$, the overall scene structure and object contours are well preserved, demonstrating superior interpolation quality and suggesting that the aligned models reside in a flatter shared basin.
Both the base method and the variant with sequential refinement outperform existing approaches by a large margin, while their average accuracy is comparable (14.22 vs. 14.40 dB PSNR at $\lambda=0.5$). The improvement therefore stems mainly from the temperature-controlled spectral sharpening (Supplementary Material, Tab. 2); the refinement stage mainly reduces the sensitivity to the temperature schedule (Supplementary Material, Tab. 3). Detailed evaluation metrics for each scene (RMSE, PSNR, and SSIM) and runtime comparisons are provided in the Supplementary Material.

\section{Conclusion}
\label{sec:conclusion}

In this paper, we addressed permutation symmetry in Neural Fields and INRs by proposing progressive alignment with temperature-controlled spectral sharpening.  
Each layer-wise update is reduced to trace maximization over a correlation matrix. Starting from this continuous representation, we combined a Hungarian anchor with temperature-$\tau$-controlled spectral annealing to converge progressively to a discrete permutation.  
This strategy mitigates premature hard decisions in weight matching while avoiding the costly iterative optimization required by gradient-based relaxations, such as OT4P, yielding a deterministic and efficient alignment procedure.  
A subsequent sequential layer-wise refinement stage further improves joint consistency across layers.
Experiments on MLPs and CNNs have demonstrated the accurate recovery of known permutations.  
Extending permutation alignment to NeRF, we showed reliable permutation recovery and reduced loss barriers between independently trained models, enabling smoother interpolation in the weight space. Our evaluation of independently trained models is limited to MLP-based NeRFs on NeRF Synthetic; other INRs and real scenes remain future work.

\section*{Acknowledgements}
This work was supported by JST PRESTO Grant Number JPMJPR25T6.

\clearpage

\bibliographystyle{splncs04}
\bibliography{ref}

\clearpage
\appendix

\setcounter{figure}{0}
\setcounter{table}{0}
\renewcommand{\thefigure}{S\arabic{figure}}
\renewcommand{\thetable}{S\arabic{table}}

\begin{center}
    {\Large\bfseries Supplemental Materials of Progressive Permutation Alignment with Temperature-Controlled Spectral Sharpening \par}
\end{center}

\section{Additional Baseline Comparisons: Activation Matching and STE}
\label{sec:additional_baselines}

In this section, we provide an extended quantitative comparison of our proposed method against two additional baselines: Activation Matching (AM) and the Straight-Through Estimator (STE). 

Table~\ref{tab:additional_baselines} presents the interpolation results at the midpoint ($\lambda=0.5$). Our method consistently outperforms the weight-based STE across all metrics (RMSE, PSNR, and SSIM). 

While Activation Matching (AM) achieves slightly better alignment metrics, it is crucial to note that \textbf{AM operates under a fundamentally different assumption}. AM is a \textit{data-dependent} approach. In the context of NeRFs, it requires sampling specific camera poses and rays to compute network activations. This requirement not only incurs a massive computational overhead but also severely limits its applicability in practical scenarios where the original training data distribution is unavailable or inaccessible.

In contrast, both STE and our proposed approach are purely \textit{weight-based (activation-free)}, aiming to align models solely using their parameter spaces without relying on any external input data. Comparing data-dependent methods (like AM) directly with weight-only methods is inherently an unequal comparison. We include the AM results here for the sake of completeness, explicitly separating the categories to ensure a fair contextualization of our weight-based method's performance.

\begin{table}[h]
\centering
\caption{Quantitative comparison against Activation Matching (AM) and STE at the interpolation midpoint ($\lambda=0.5$). Note that AM is a data-dependent method, whereas STE and our proposed method are strictly weight-based (activation-free).}
\label{tab:additional_baselines}
\resizebox{0.7\linewidth}{!}{
\begin{tabular}{@{}llccc@{}}
\toprule
\textbf{Method} & \textbf{Category} & \textbf{RMSE $\downarrow$} & \textbf{PSNR $\uparrow$} & \textbf{SSIM $\uparrow$} \\ \midrule
AM & Data-dependent & 0.161 & 16.22 & 0.775 \\ \midrule
STE & Weight-based & 0.268 & 11.84 & 0.661 \\
Ours & Weight-based & \textbf{0.199} & \textbf{14.40} & \textbf{0.710} \\ \bottomrule
\end{tabular}
}
\end{table}

\section{Ablation Study}
\label{supp:ablation}

In this section, we conduct ablation studies to evaluate the contribution of each component in our proposed method and to analyze the sensitivity of the performance to the temperature scheduling of $\tau$. All experiments in this section were performed using the NeRF Synthetic dataset.

\subsection{Importance of Temperature-Controlled Spectral Projection}
\label{supp:ablation_svd}

To verify the effectiveness of our spectral sharpening and projection mechanism, we compared the variants of our proposed method. For reference, we also include the baseline methods (Naive, Weight Matching, and Progressive Weight Matching) detailed in Section 4.1 of the main manuscript:
\begin{itemize}
    \item \textbf{Ours (w/o SVD, w/o Ref.)}: Progressive alignment that bypasses the SVD and spectral flattening steps (Lines 9--10 in Algorithm \ref{alg:proposed}). Instead of the sharpened matrix $\mat{M}_l^{\langle \tau \rangle}$, it directly uses the raw matrix $\mat{M}_l$ for the continuous relaxation (Line 11: $\mat{P}_l \leftarrow \hat{\mat{P}}_l [ \tau \mat{M}_l + (1 - \tau) \mat{I} ]$), and omits the final refinement stage.
    \item \textbf{Ours (w/o SVD)}: The variant without spectral flattening (as described above), but followed by the sequential layer-wise refinement.
    \item \textbf{Ours (w/o Ref.)}: The method executing Algorithm \ref{alg:proposed} completely (i.e., with spectral flattening), but omitting the final refinement stage.
    \item \textbf{Ours}: Our complete two-stage method combining the spectrally-sharpened continuous alignment (Algorithm \ref{alg:proposed}) with the final sequential layer-wise refinement.
\end{itemize}

\begin{table}[ht]
    \centering
    \caption{Quantitative ablation results on the importance of Spectral Projection and Sequential Layer-wise Refinement. Average metrics across the NeRF Synthetic dataset are reported at the midpoint of interpolation ($\lambda=0.5$).}
    \label{tab:ablation_svd}
    \begin{tabular}{lccc}
        \toprule
        \textbf{Method} & \textbf{RMSE} $\downarrow$ & \textbf{PSNR} $\uparrow$ & \textbf{SSIM} $\uparrow$ \\
        \midrule
        Naive                       & 0.491 & 6.205 & 0.627 \\
        Weight Matching             & 0.294 & 11.07 & 0.653 \\
        Progressive Weight Matching & 0.306 & 10.64 & 0.637 \\
        Ours (w/o SVD, w/o Ref.)    & 0.264 & 12.39 & 0.674 \\
        Ours (w/o SVD)              & 0.248 & 12.70 & 0.669 \\
        Ours (w/o Ref.)             & 0.199 & 14.22 & 0.697 \\
        Ours                        & 0.199 & 14.40 & 0.710 \\
        \bottomrule
    \end{tabular}
\end{table}

\begin{figure}[ht]
    \centering
    \includegraphics[width=0.95\linewidth]{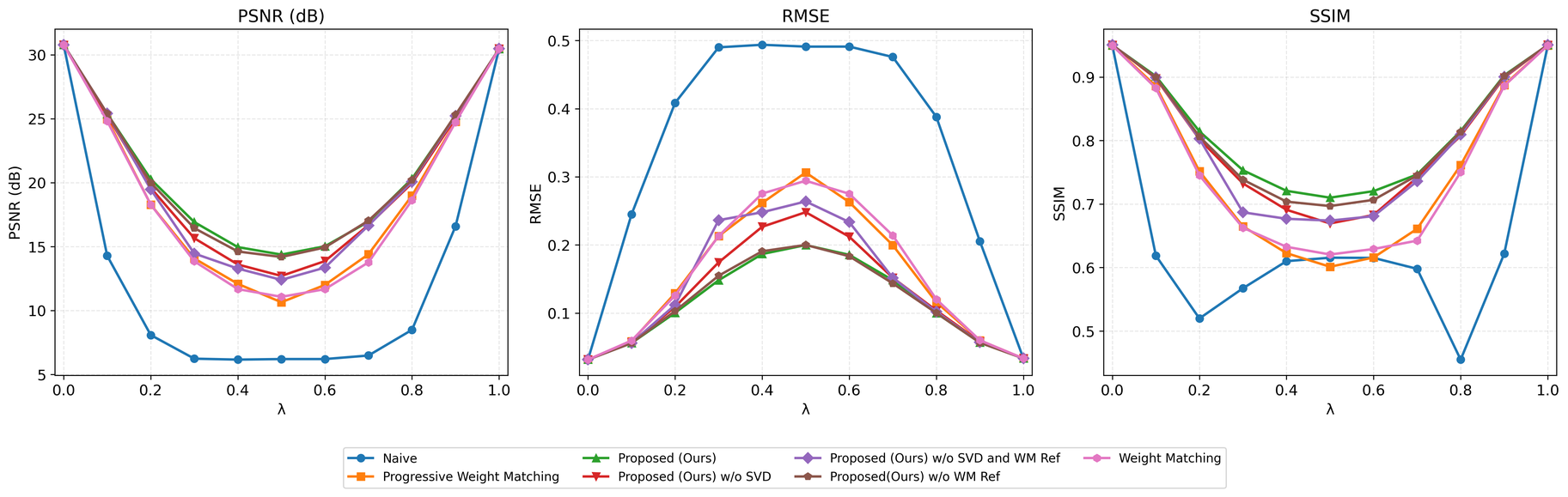}
    \caption{Quantitative evaluation of interpolation trajectories for different ablation variants across $\lambda \in [0, 1]$. The integration of spectral projection significantly elevates the overall performance curve, while the addition of sequential layer-wise refinement further stabilizes the alignment, achieving the highest quality at the midpoint ($\lambda=0.5$).}
    \label{fig:ablation_svd_visuals}
\end{figure}

Table~\ref{tab:ablation_svd} and Figure~\ref{fig:ablation_svd_visuals} summarize the ablation results. The consistent performance improvement observed from ``Ours (w/o SVD, w/o Ref.)'' to the complete ``Ours'' underscores the complementary nature of our algorithmic components. 
Specifically, the substantial gain achieved by introducing spectral projection (comparing ``Ours (w/o SVD)'' vs. ``Ours'') demonstrates its critical role: simply interpolating with the raw correlation matrix $\mat{M}_l$ is insufficient to navigate the complex loss landscape. The temperature-controlled spectral flattening (generating $\mat{M}_l^{\langle \tau \rangle}$) is essential to actively suppress noise and emphasize dominant permutation structures during the progressive updates. Furthermore, the final sequential layer-wise refinement consistently improves PSNR, indicating that the spectrally-sharpened continuous solution serves as a robust initialization for discrete local optimization.

\subsection{Effect of Temperature Scheduling for $\tau$}
\label{supp:ablation_tau}

Our method relies on a temperature parameter $\tau$ to control the sharpening of the spectrum. We investigated how the granularity of the $\tau$ schedule affects the rendering quality of interpolated models, and how this continuous initialization interacts with the final sequential layer-wise refinement. We compared three different schedules:
\begin{enumerate}
    \item \textbf{Two-step}: $\tau \in \{1.0, 0.0\}$.
    \item \textbf{Three-step}: $\tau \in \{1.0, 0.5, 0.0\}$.
    \item \textbf{Five-step}: $\tau \in \{1.0, 0.75, 0.5, 0.25, 0.0\}$.
\end{enumerate}

\begin{table}[ht]
    \centering
    \caption{Quantitative ablation results on the temperature scheduling of $\tau$. We evaluate different schedule granularities both with and without the final refinement. Average metrics across the NeRF Synthetic dataset are reported at $\lambda=0.5$.}
    \label{tab:ablation_tau}
    \begin{tabular}{lccc}
        \toprule
        \textbf{Schedule} & \textbf{RMSE} $\downarrow$ & \textbf{PSNR} $\uparrow$ & \textbf{SSIM} $\uparrow$ \\
        \midrule
        Naive                               & 0.491 & 6.205 & 0.616 \\
        Weight Matching                     & 0.294 & 11.07 & 0.621 \\
        Progressive Weight Matching         & 0.306 & 10.64 & 0.601 \\
        \midrule
        Two-step ($\tau \in \{1.0, 0.0\}$) w/o Ref.             & 0.241 & 12.89 & 0.698 \\
        Two-step ($\tau \in \{1.0, 0.0\}$)                      & 0.256 & 12.85 & 0.697 \\
        Three-step ($\tau \in \{1.0, 0.5, 0.0\}$) w/o Ref.      & 0.199 & 14.22 & 0.697 \\
        Three-step ($\tau \in \{1.0, 0.5, 0.0\}$)               & 0.199 & 14.40 & 0.710 \\
        Five-step ($\tau \in \{1.0, 0.75, 0.5, 0.25, 0.0\}$) w/o Ref. & 0.256 & 12.35 & 0.695 \\
        Five-step ($\tau \in \{1.0, 0.75, 0.5, 0.25, 0.0\}$)    & 0.228 & 13.45 & 0.698 \\
        \bottomrule
    \end{tabular}
\end{table}

\begin{figure}[ht]
    \centering
    \includegraphics[width=0.95\linewidth]{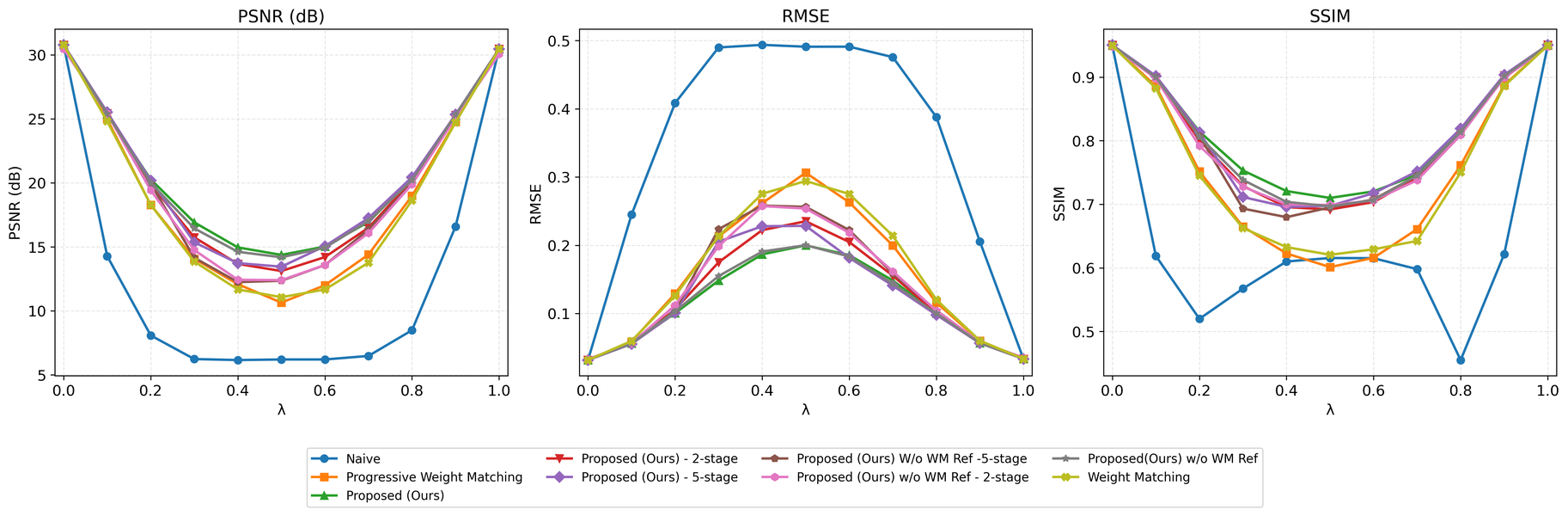}
    \caption{Impact of temperature scheduling on interpolation quality across different $\lambda$ values. The three-step schedule strikes the optimal balance between continuous exploration and discrete exploitation, outperforming both the abrupt binary hard-switch (two-step) and the overly gradual approach (five-step).}
    \label{fig:ablation_tau_plot}
\end{figure}

The results, presented in Table~\ref{tab:ablation_tau} and Figure~\ref{fig:ablation_tau_plot}, reveal a non-monotonic relationship between the schedule granularity and the final alignment quality. While all proposed temperature schedules significantly outperform the standard Weight Matching baselines, the three-step schedule emerges as the clear optimal choice. 

Specifically, the two-step binary schedule effectively reduces the search space but occasionally leads to premature fixation on sub-optimal permutations due to the abrupt hard-switch. Conversely, while one might expect a highly gradual schedule (five-step) to yield the best results, it performs similarly to the two-step approach. We hypothesize that excessive intermediate relaxation steps cause the optimization to drift, accumulating alignment errors in the continuous state before the permutations can properly solidify. The three-step schedule provides the optimal ``sweet spot'' of relaxation: it offers sufficient continuous exploration at $\tau=0.5$ to escape local optima without introducing alignment drift, thereby achieving the highest and most stable performance across all evaluated metrics, both with and without the final sequential refinement.

\section{Additional Quantitative and Qualitative Results}
\label{supp:additional_results}

In this section, we provide further quantitative and qualitative evaluations to complement the findings presented in the main manuscript. 
We first evaluate the computational efficiency of the proposed method by reporting the execution time across different network architectures. 
This is followed by an extensive qualitative analysis, where we present the full-size versions of the comparisons moved from the main paper, rendering results from novel test viewpoints, and comprehensive results for all remaining scenes in the datasets.

\subsection{Computational Efficiency}
\label{supp:runtime}

Tables \ref{tab:computational_time_cnn} and \ref{tab:computational_time_nerf} summarize the mean and standard deviation of the time required for permutation alignment across various architectures. 
Our method provides a significant efficiency advantage, as it relies on deterministic progressive alignment via spectral projection, entirely avoiding the costly inner-loop gradient optimization required by continuous relaxation methods like OT4P~\cite{guo2024ot4p}. 

Particularly for high-dimensional layers, such as the fully connected layers in VGG11, our complete method achieves alignment in under a minute (48 seconds), whereas the gradient-based methods (OT4P) require about 40 minutes per run, roughly 50 times longer. Furthermore, the results demonstrate that the addition of the final sequential layer-wise refinement (comparing ``Proposed w/o Ref.'' and ``Proposed'') introduces only a small computational overhead (e.g., adding approximately  3 seconds for NeRF). This highlights that our approach successfully achieves the robust alignment quality characteristic of continuous relaxation methods, while maintaining the extreme computational efficiency of purely discrete solvers.

\begin{table}[h!]
    \centering
    \caption{Alignment computation time (mean and standard deviation, [sec])}
    \label{tab:computational_time_cnn}
    \resizebox{0.8\linewidth}{!}{%
    \begin{tabular}{lccc}
        \toprule
        \textbf{Algorithm} 
        & MLP5
        & VGG11
        & ResNet18 \\
        \midrule
        Weight Matching                 
            & 0.538 $\pm$ 0.080 
            & 169.5 $\pm$ 14.06 
            & 7.937 $\pm$ 0.501 \\
        Progressive Weight Matching     
            & 0.052 $\pm$ 0.047 
            & 10.43 $\pm$ 0.064 
            & 0.752 $\pm$ 0.091 \\
        OT4P ($\tau=0.7$)                            
            & 15.54 $\pm$ 1.574 
            & 2307 $\pm$ 12.17 
            & 54.99 $\pm$ 5.747 \\
        OT4P ($\tau=0.5$)                            
            & 16.36 $\pm$ 0.358 
            & 2605 $\pm$ 591.2 
            & 53.06 $\pm$ 0.097 \\
        OT4P ($\tau=0.3$)                            
            & 17.24 $\pm$ 0.850 
            & 2380 $\pm$ 19.86
            & 55.64 $\pm$ 5.661 \\
        Progressive OT4P ($\tau=0.7$)                
            & 10.24 $\pm$ 0.626 
            & 2059 $\pm$ 219.8  
            & 46.66 $\pm$ 4.084 \\
        Progressive OT4P ($\tau=0.5$)                
            & 7.330 $\pm$ 0.911 
            & 2137 $\pm$ 109.8
            & 46.72 $\pm$ 6.345 \\
        Progressive OT4P ($\tau=0.3$)                
            & 14.46 $\pm$ 0.497 
            & 2216 $\pm$ 11.25 
            & 60.33 $\pm$ 3.450 \\
        Proposed (Ours) w/o Ref.                
            & 0.753 $\pm$ 0.055
            & 38.95 $\pm$ 14.43 
            & 1.698 $\pm$ 0.035 \\
        Proposed (Ours)  
            & 0.758 $\pm$ 0.054 
            & 40.22 $\pm$ 14.49
            & 1.842 $\pm$ 0.022 \\
        \bottomrule
    \end{tabular}
    }
\end{table}

\begin{table}[h!]
    \centering
    \caption{Computation time for NeRF alignment (mean and variance, [sec])}
    \label{tab:computational_time_nerf}
    \resizebox{0.85\linewidth}{!}{%
    \begin{tabular}{lcc}
        \toprule
        \textbf{Algorithm} 
        & NeRF (Shuffled Weights) 
        & NeRF (Different Initializations) \\
        \midrule
        Weight Matching         
            & 15.60 $\pm$ 0.871 
            & 153.7 $\pm$ 25.09 \\
        Progressive Weight Matching     
            & 1.266 $\pm$ 0.097 
            & 14.55 $\pm$ 2.430 \\
        OT4P ($\tau=0.3$)                           
            & 48.40 $\pm$ 23.75 
            & 38.72 $\pm$ 11.11 \\   
        OT4P ($\tau=0.5$)                           
            & 66.50 $\pm$ 5.688 
            & 17.05 $\pm$ 4.585 \\
        OT4P ($\tau=0.7$)                           
            & 60.63 $\pm$ 7.621 
            & 66.06 $\pm$ 13.83 \\
        Progressive OT4P ($\tau=0.3$)               
            & 93.62 $\pm$ 3.201 
            & 92.21 $\pm$ 2.056 \\
        Progressive OT4P ($\tau=0.5$)               
            & 93.68 $\pm$ 1.052 
            & 81.29 $\pm$ 2.985 \\
        Progressive OT4P ($\tau=0.7$)               
            & 91.66 $\pm$ 1.154 
            & 91.51 $\pm$ 2.454 \\
        Proposed (Ours) w/o Ref.                
            & 3.063 $\pm$ 0.009
            & 3.220 $\pm$ 0.005 \\
        Proposed (Ours)
            & 3.165 $\pm$ 0.149
            & 6.001 $\pm$ 0.075 \\
        \bottomrule
    \end{tabular}
    }
\end{table}

\subsection{Extended Qualitative Comparisons}
\label{supp:qualitative}

We provide additional quantitative and qualitative results on model interpolation for the NeRF Synthetic dataset.
While the main paper presents representative examples, this supplementary material reports results for all eight scenes, including SSIM, RMSE, and PSNR curves, together with qualitative interpolation results.

\begin{figure}[h!]
    \centering
    \includegraphics[width=\linewidth]{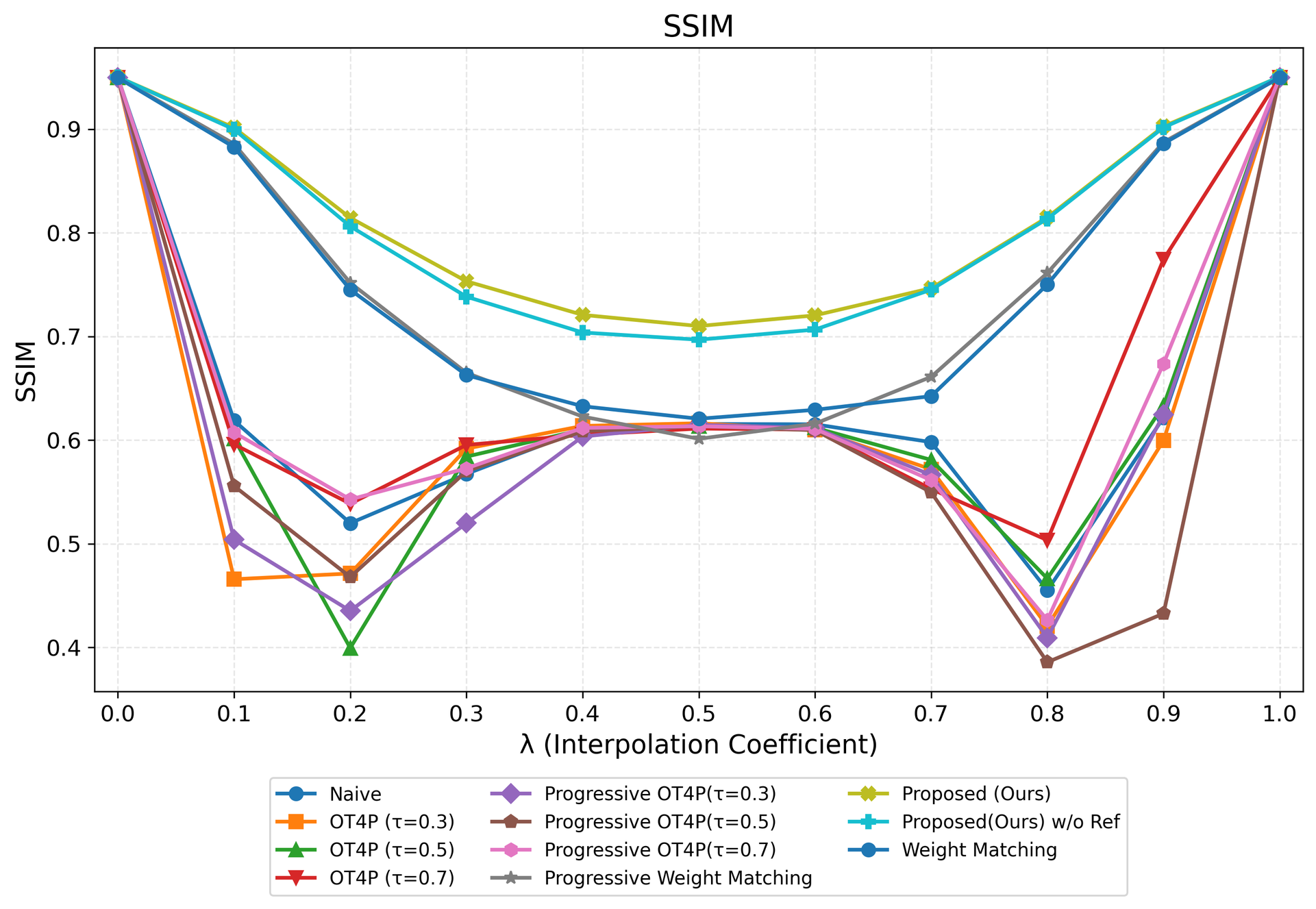}
    \caption{Average SSIM over the eight scenes of the NeRF Synthetic dataset. The horizontal axis denotes the interpolation coefficient $\lambda$, and the vertical axis denotes the mean SSIM.}
    \label{fig:supp_aggregate_ssim}
\end{figure}

Figure~\ref{fig:supp_aggregate_ssim} shows the average SSIM across all scenes.
The proposed method preserves higher structural similarity along the interpolation path than the baselines, particularly near the middle of the interpolation path, where the interpolation quality typically degrades.

\begin{figure}[t]
    \centering
    \begin{subfigure}{0.48\linewidth}
        \includegraphics[width=\linewidth]{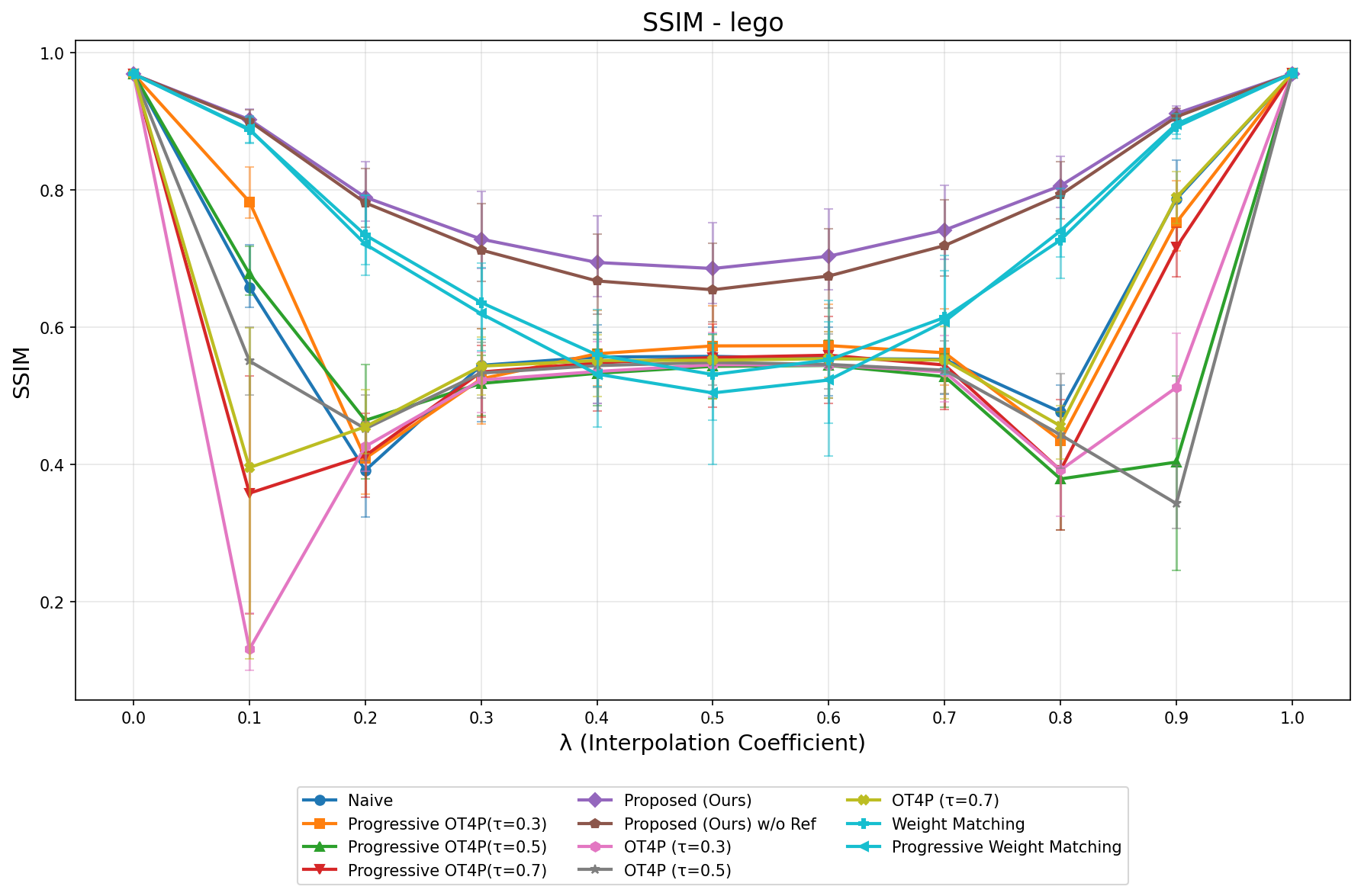}
        \caption{Lego}
    \end{subfigure}
    \hfill
    \begin{subfigure}{0.48\linewidth}
        \includegraphics[width=\linewidth]{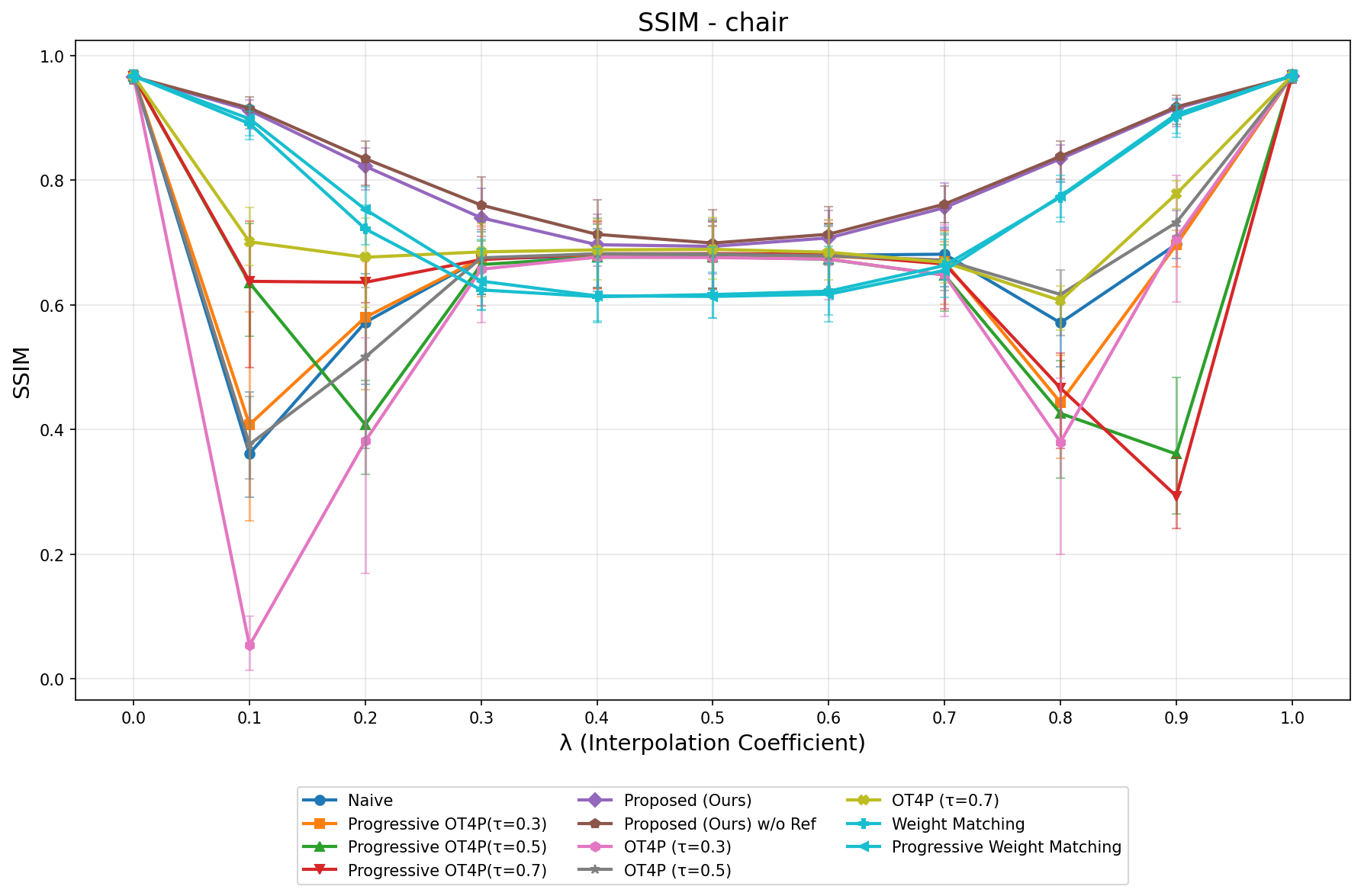}
        \caption{Chair}
    \end{subfigure}
    \vspace{3mm}
    \begin{subfigure}{0.48\linewidth}
        \includegraphics[width=\linewidth]{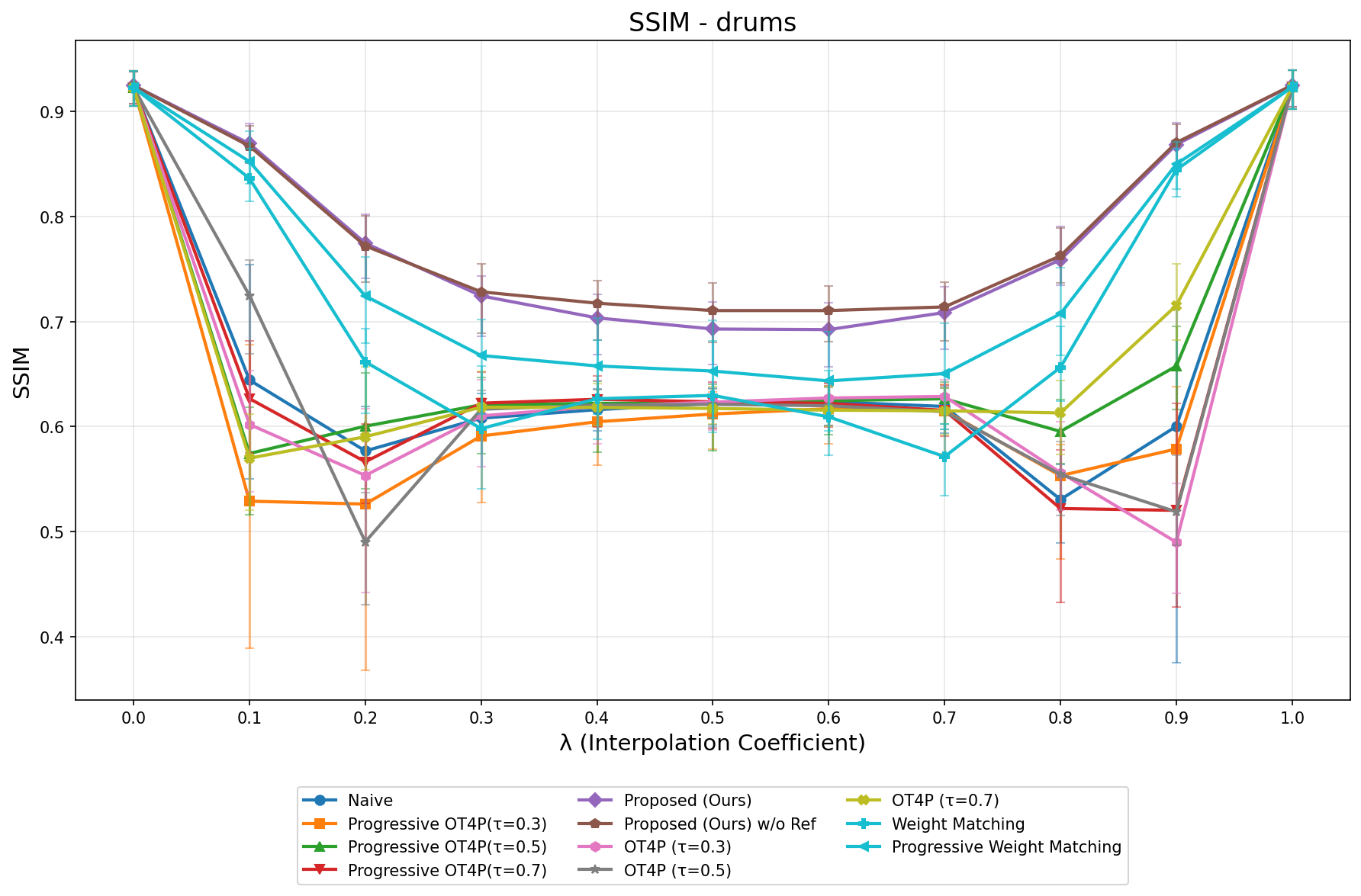}
        \caption{Drums}
    \end{subfigure}
    \hfill
    \begin{subfigure}{0.48\linewidth}
        \includegraphics[width=\linewidth]{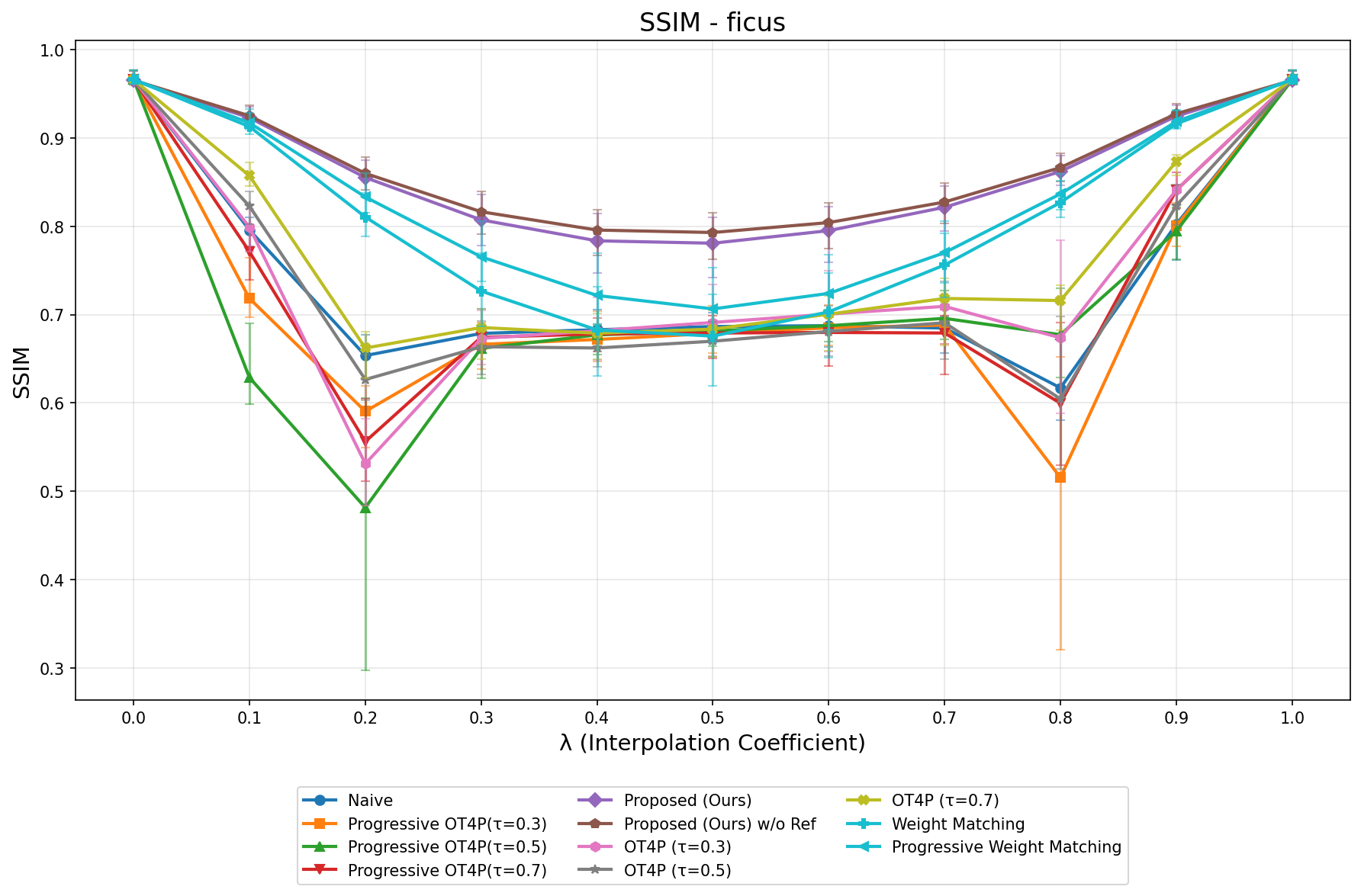}
        \caption{Ficus}
    \end{subfigure}
    \vspace{3mm}
    \begin{subfigure}{0.48\linewidth}
        \includegraphics[width=\linewidth]{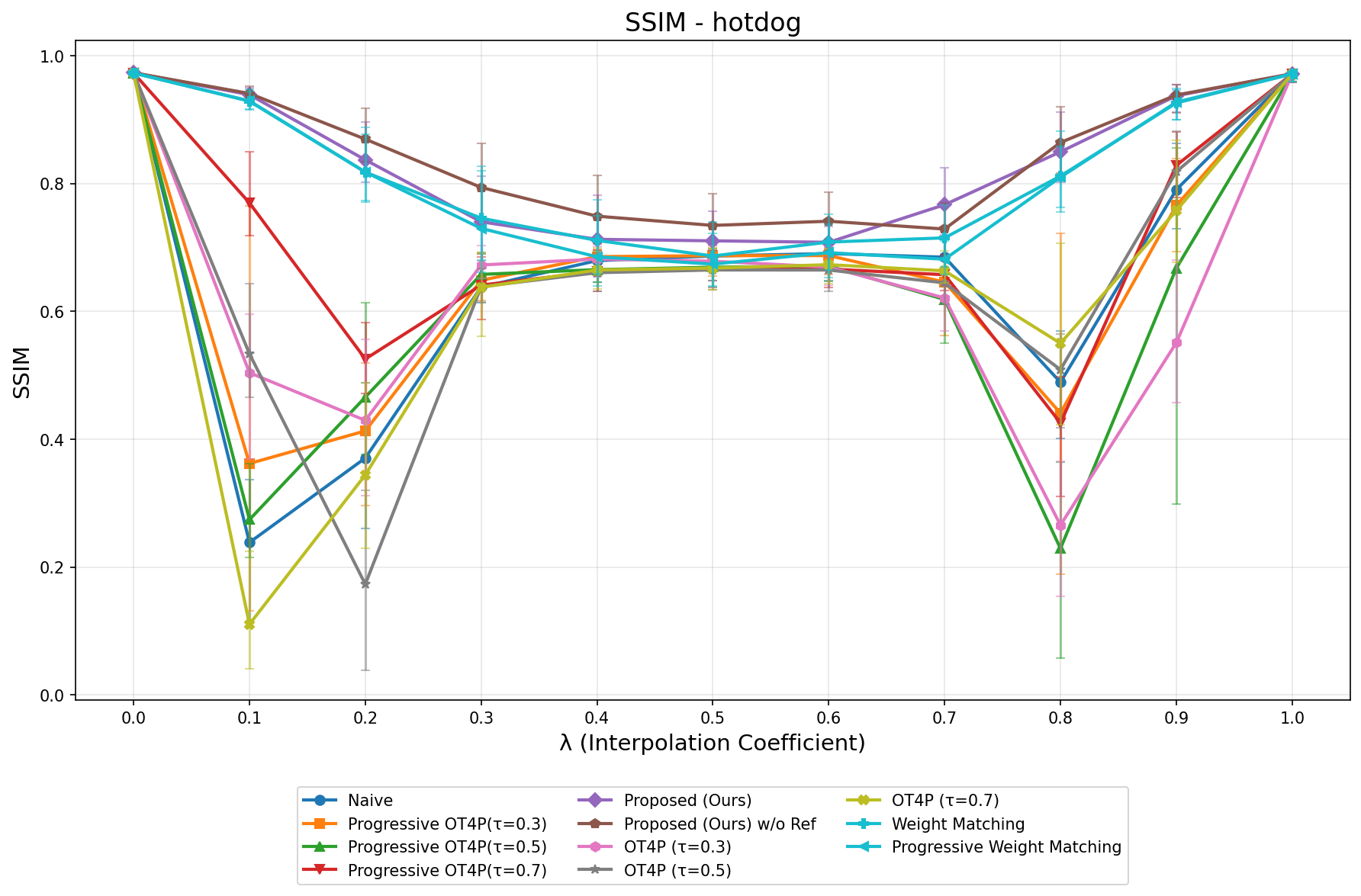}
        \caption{Hotdog}
    \end{subfigure}
    \hfill
    \begin{subfigure}{0.48\linewidth}
        \includegraphics[width=\linewidth]{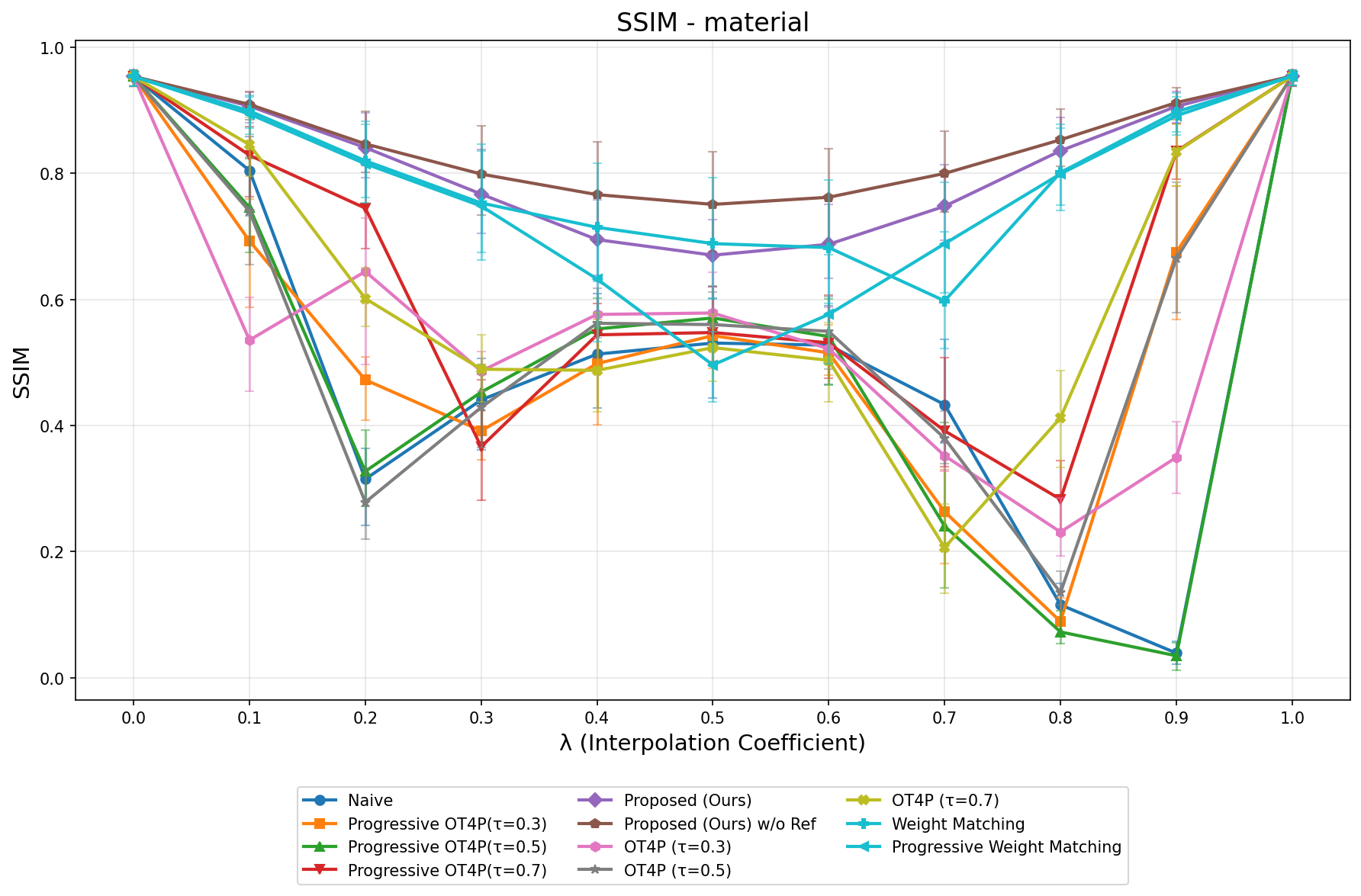}
        \caption{Materials}
    \end{subfigure}
    \vspace{3mm}
    \begin{subfigure}{0.48\linewidth}
        \includegraphics[width=\linewidth]{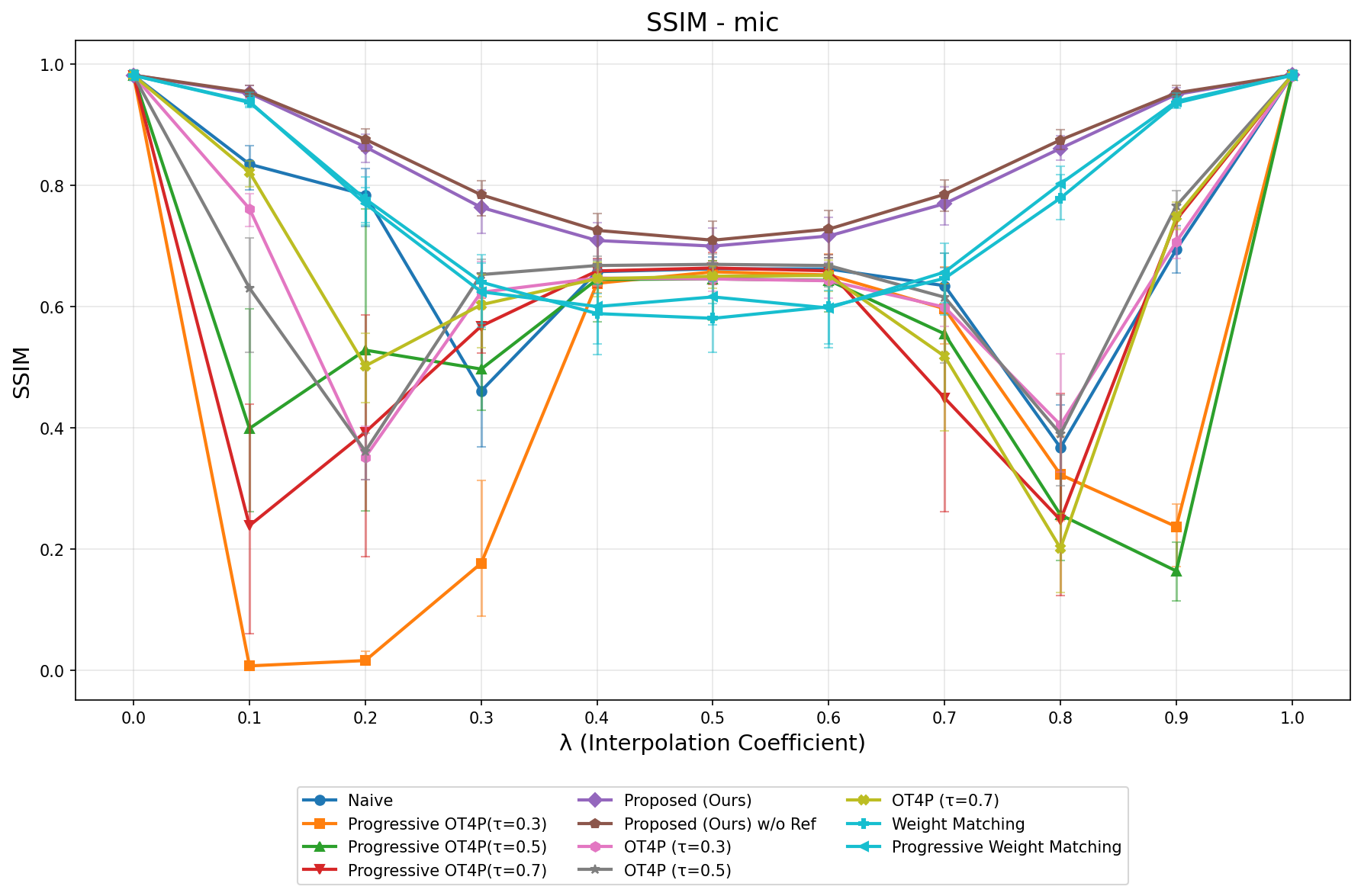}
        \caption{Mic}
    \end{subfigure}
    \hfill
    \begin{subfigure}{0.48\linewidth}
        \includegraphics[width=\linewidth]{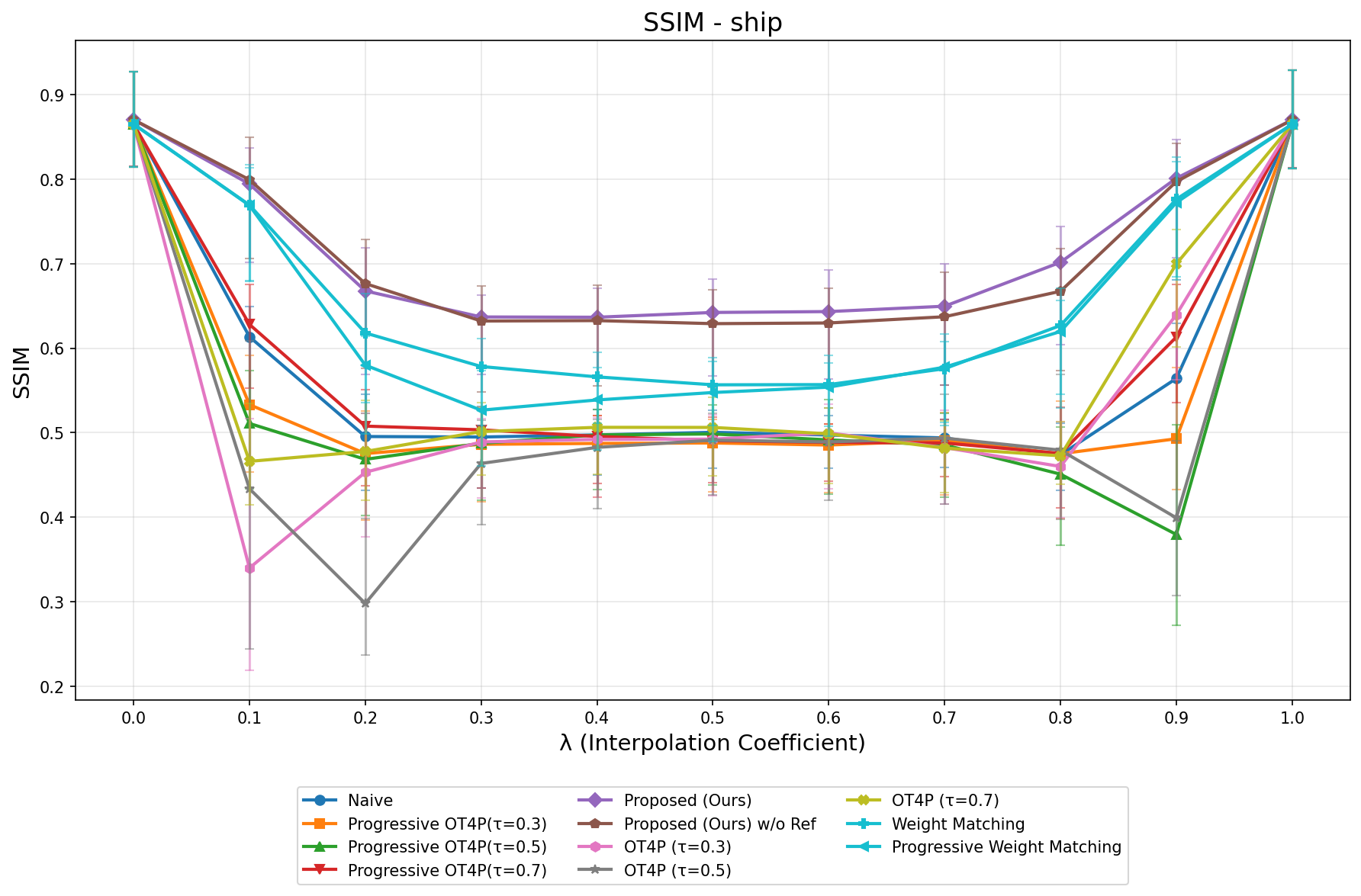}
        \caption{Ship}
    \end{subfigure}
    \caption{SSIM for each scene in the NeRF Synthetic dataset. The proposed method shows a smaller drop in SSIM in the intermediate region of interpolation.}
    \label{fig:supp_individual_ssim}
\end{figure}

Figure~\ref{fig:supp_individual_ssim} shows the SSIM curves for each scene.
A consistent trend is observed across all eight scenes, indicating that the proposed method reduces the performance drop around the middle of the interpolation path.

\begin{figure}[t]
    \centering
    \begin{subfigure}{0.24\linewidth}
        \includegraphics[width=\linewidth]{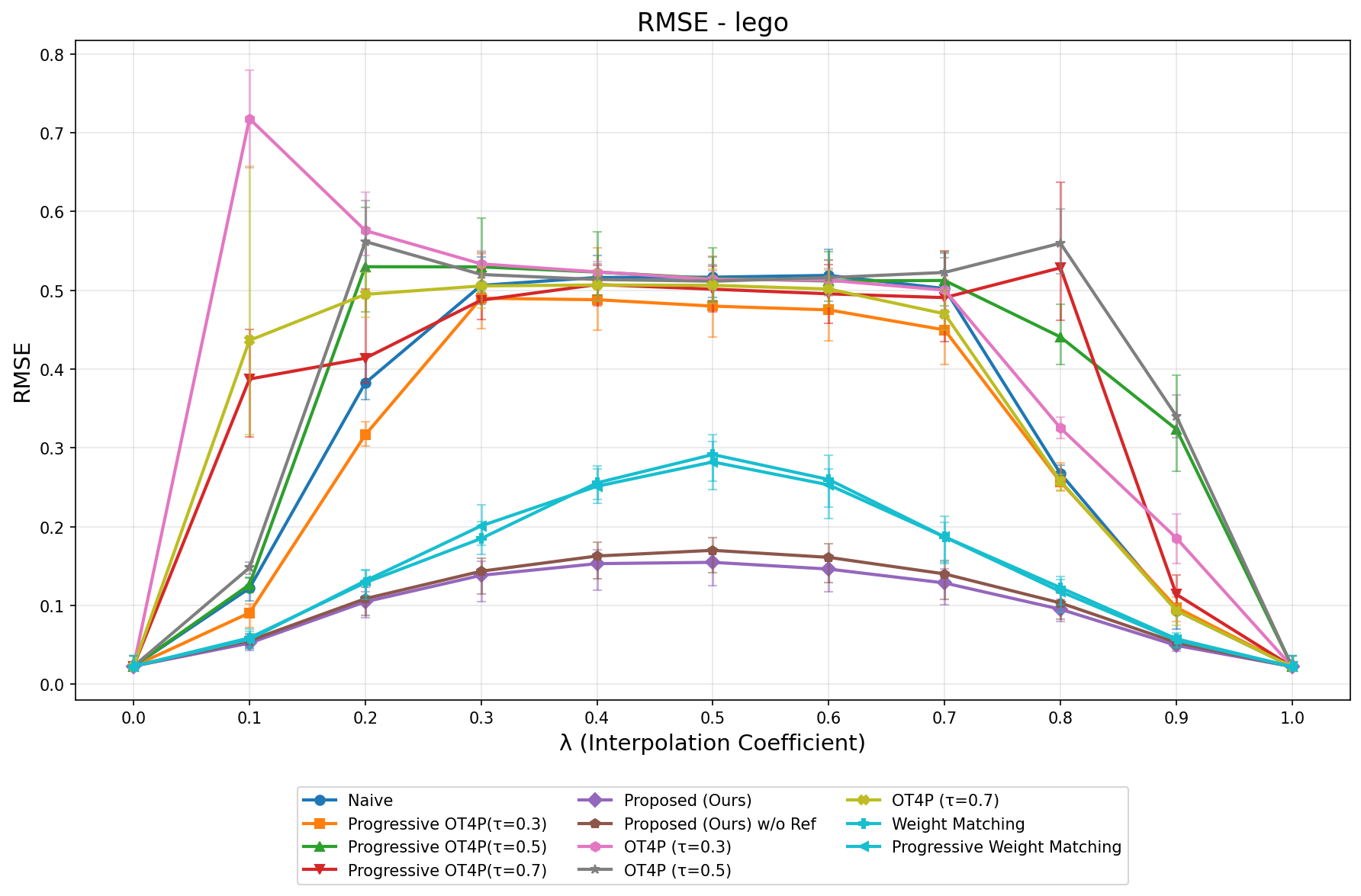}
        \caption{Lego}
    \end{subfigure}
    \hfill
    \begin{subfigure}{0.24\linewidth}
        \includegraphics[width=\linewidth]{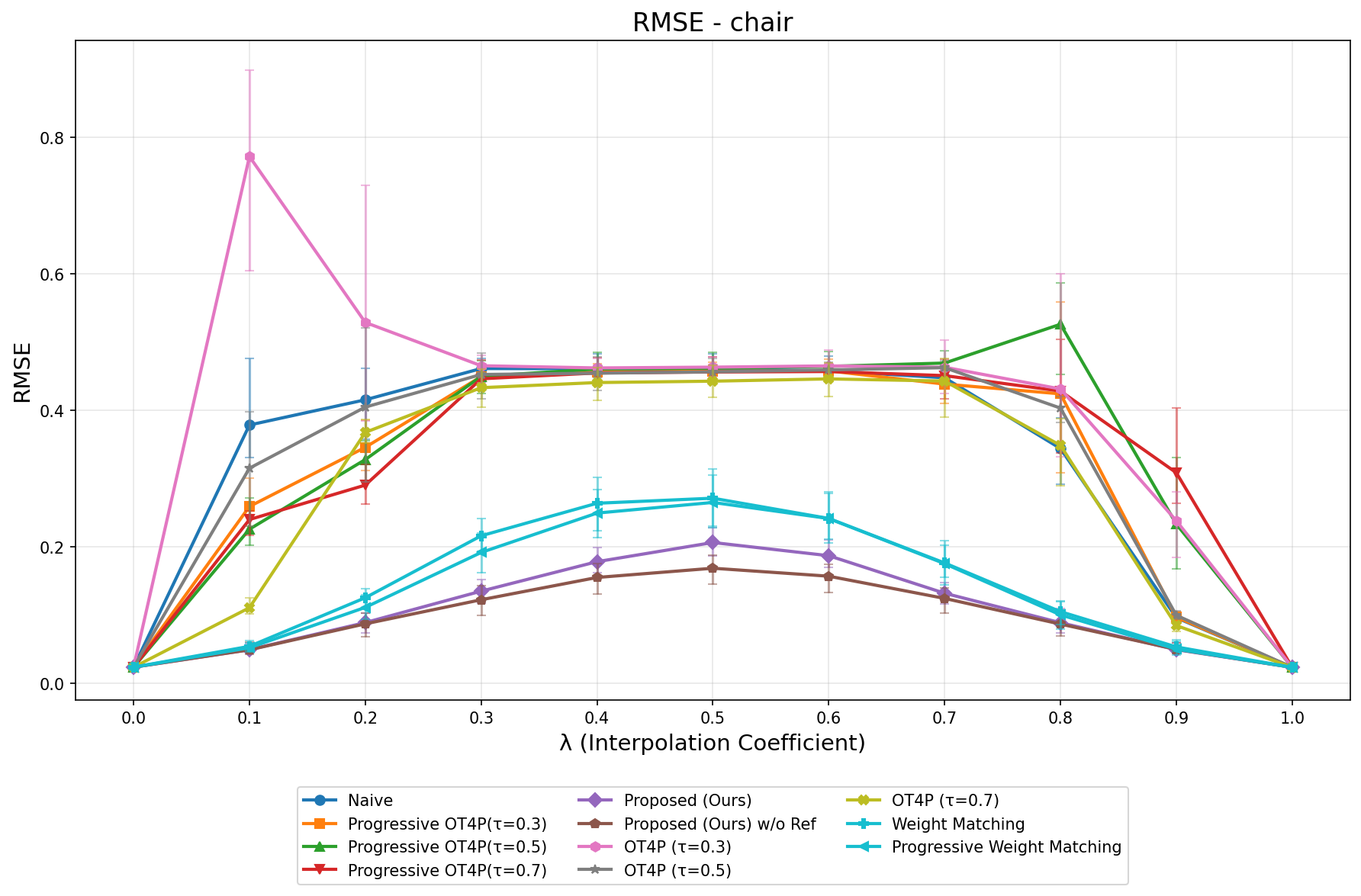}
        \caption{Chair}
    \end{subfigure}
    \hfill
    \begin{subfigure}{0.24\linewidth}
        \includegraphics[width=\linewidth]{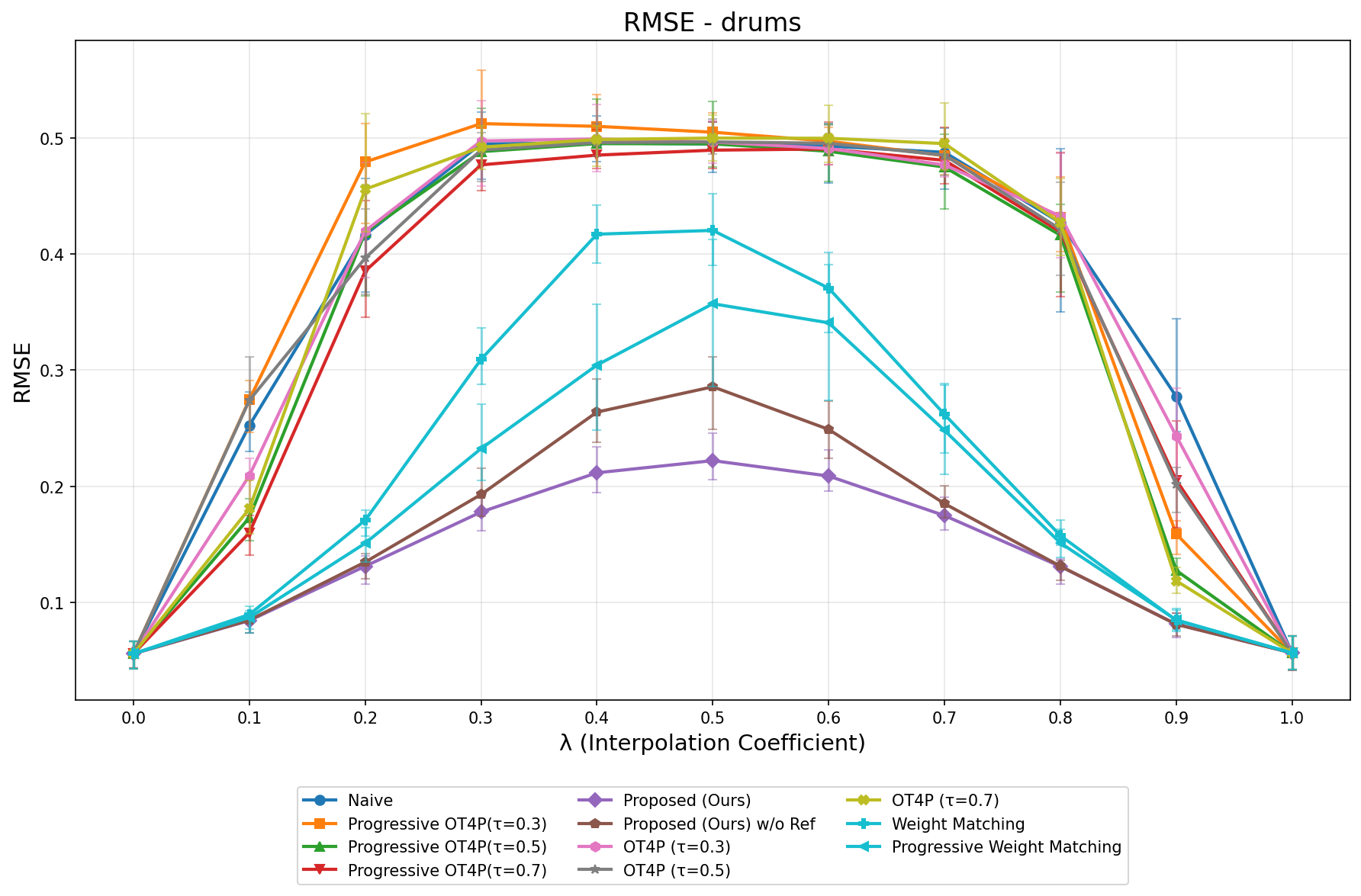}
        \caption{Drums}
    \end{subfigure}
    \hfill
    \begin{subfigure}{0.24\linewidth}
        \includegraphics[width=\linewidth]{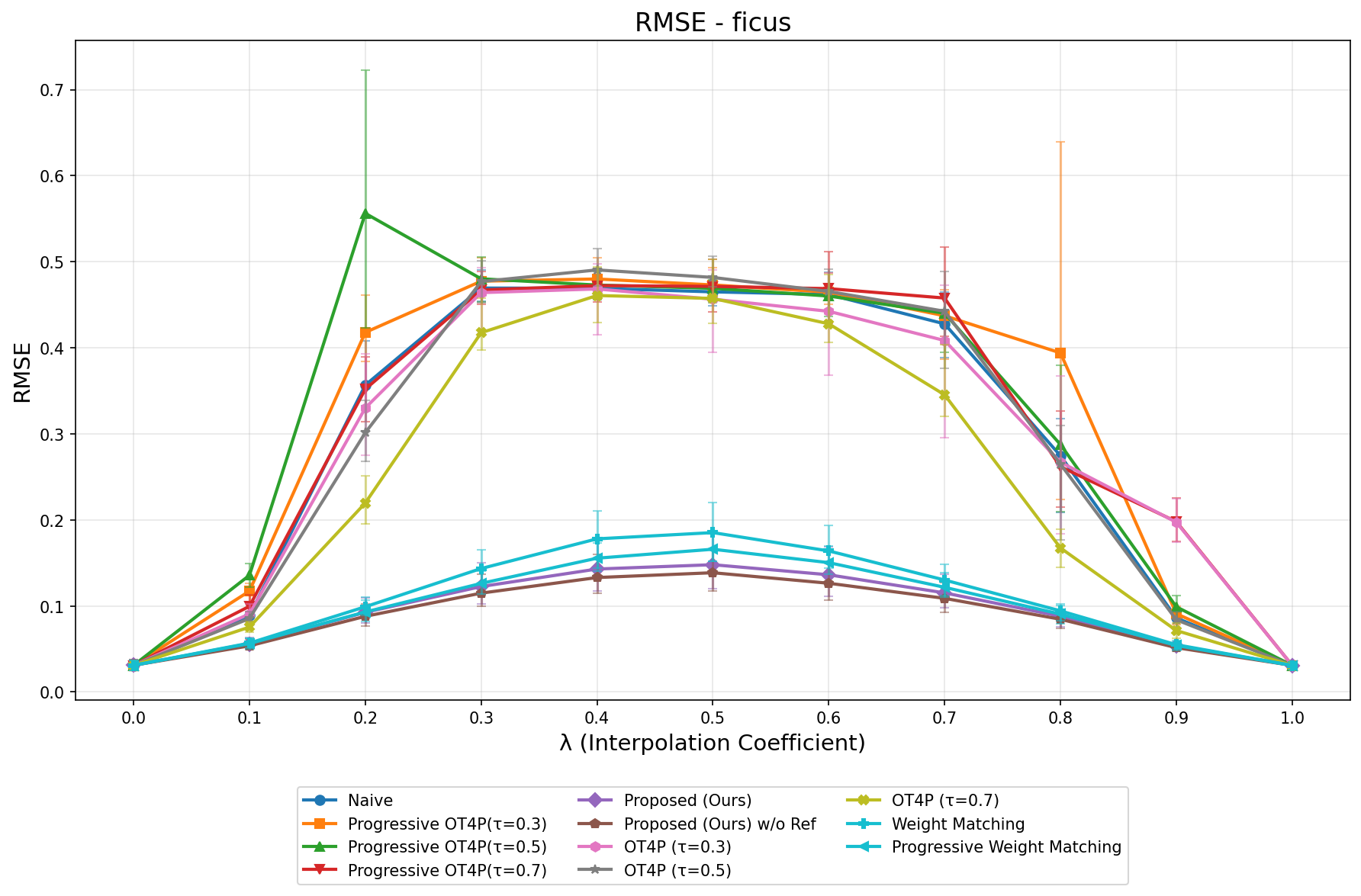}
        \caption{Ficus}
    \end{subfigure}
    \vspace{2mm}
    \begin{subfigure}{0.24\linewidth}
        \includegraphics[width=\linewidth]{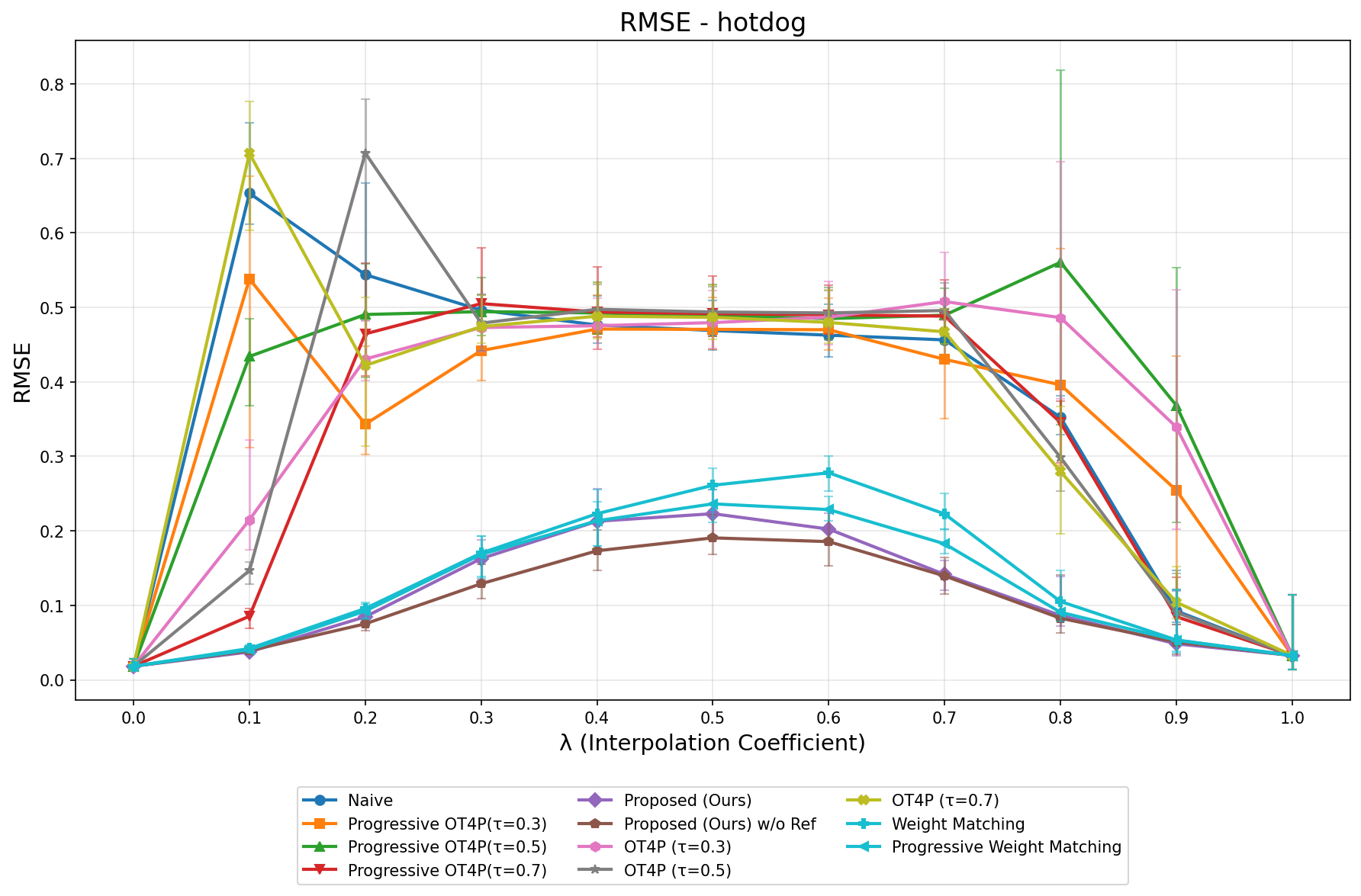}
        \caption{Hotdog}
    \end{subfigure}
    \hfill
    \begin{subfigure}{0.24\linewidth}
        \includegraphics[width=\linewidth]{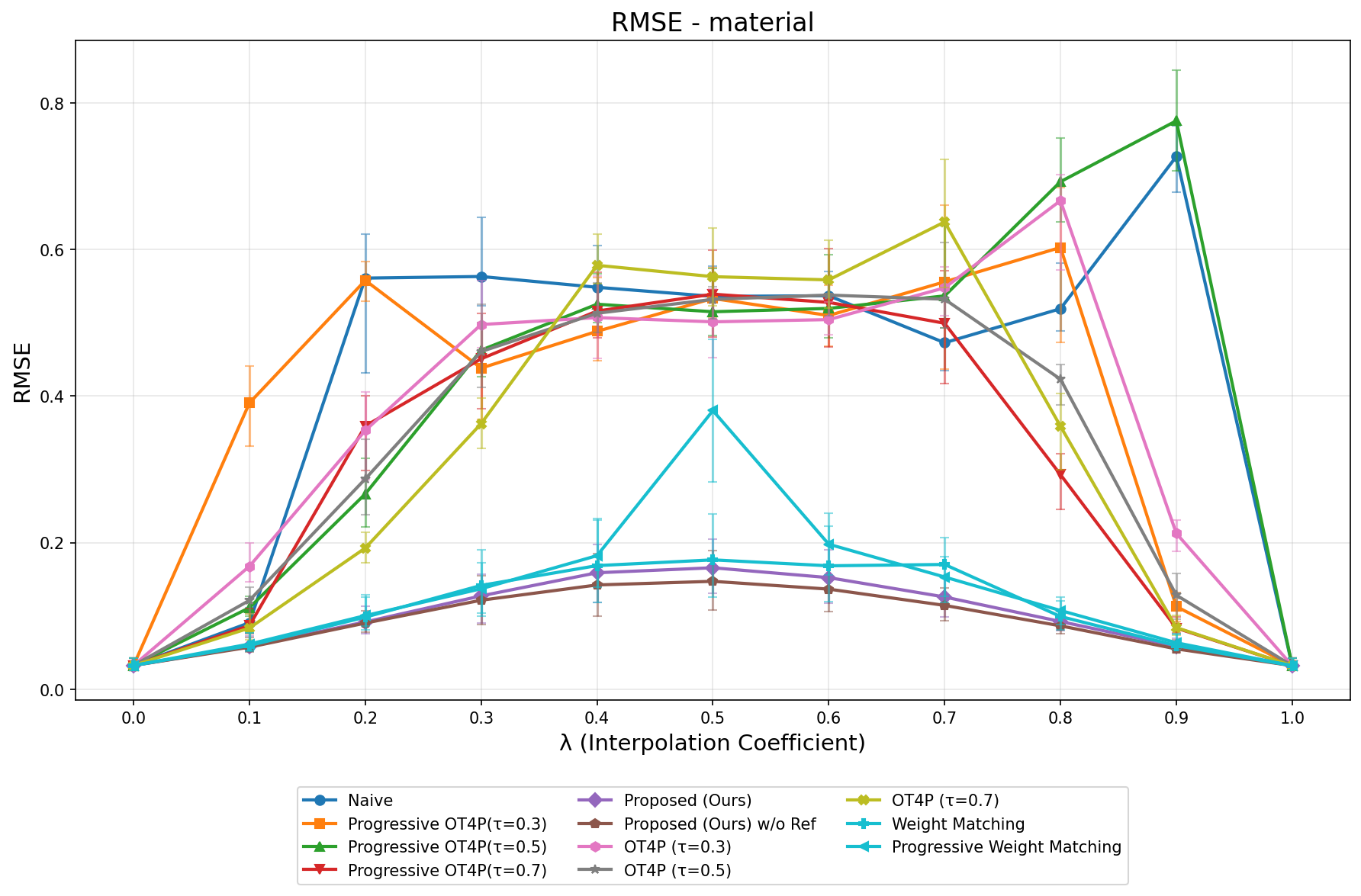}
        \caption{Materials}
    \end{subfigure}
    \hfill
    \begin{subfigure}{0.24\linewidth}
        \includegraphics[width=\linewidth]{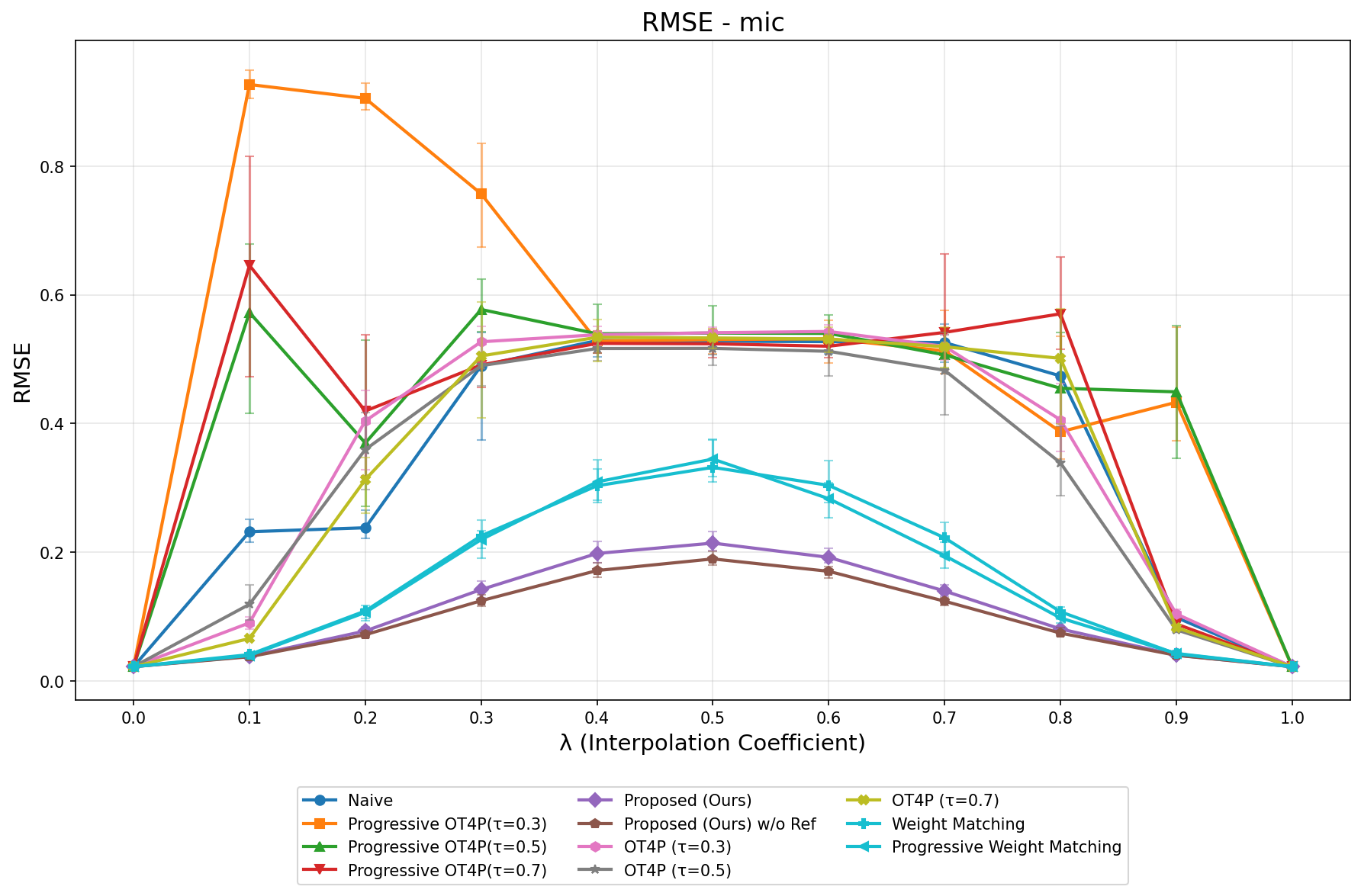}
        \caption{Mic}
    \end{subfigure}
    \hfill
    \begin{subfigure}{0.24\linewidth}
        \includegraphics[width=\linewidth]{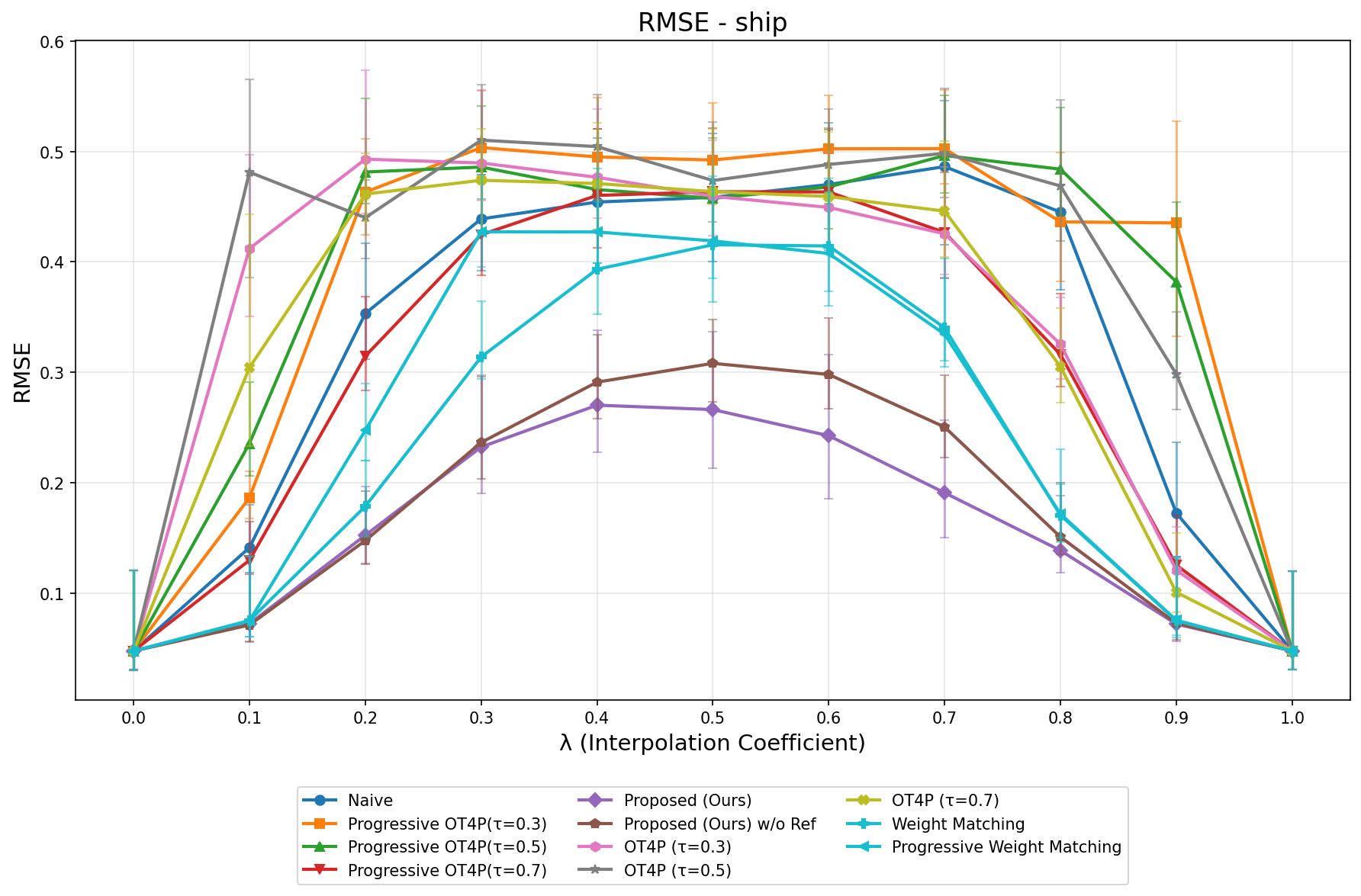}
        \caption{Ship}
    \end{subfigure}
    \caption{RMSE along the linear interpolation path for all eight scenes of the NeRF Synthetic dataset. Lower is better.}
    \label{fig:nerf_lmc_rmse_all_datasets}
\end{figure}

Figure~\ref{fig:nerf_lmc_rmse_all_datasets} shows the RMSE curves.
The proposed method maintains a lower reconstruction error along the interpolation path.

\begin{figure}[t]
    \centering
    \begin{subfigure}{0.24\linewidth}
        \includegraphics[width=\linewidth]{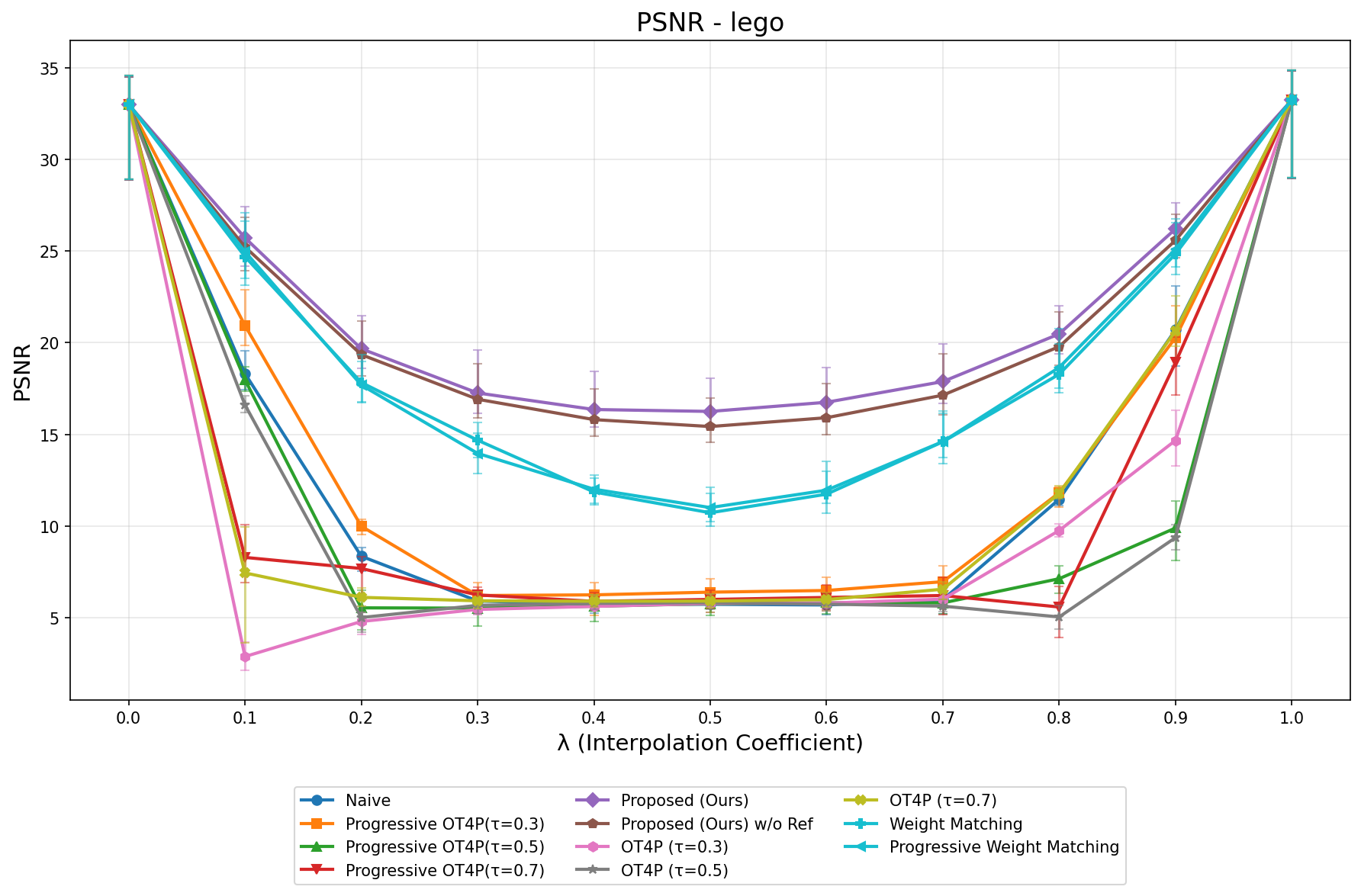}
        \caption{Lego}
    \end{subfigure}
    \hfill
    \begin{subfigure}{0.24\linewidth}
        \includegraphics[width=\linewidth]{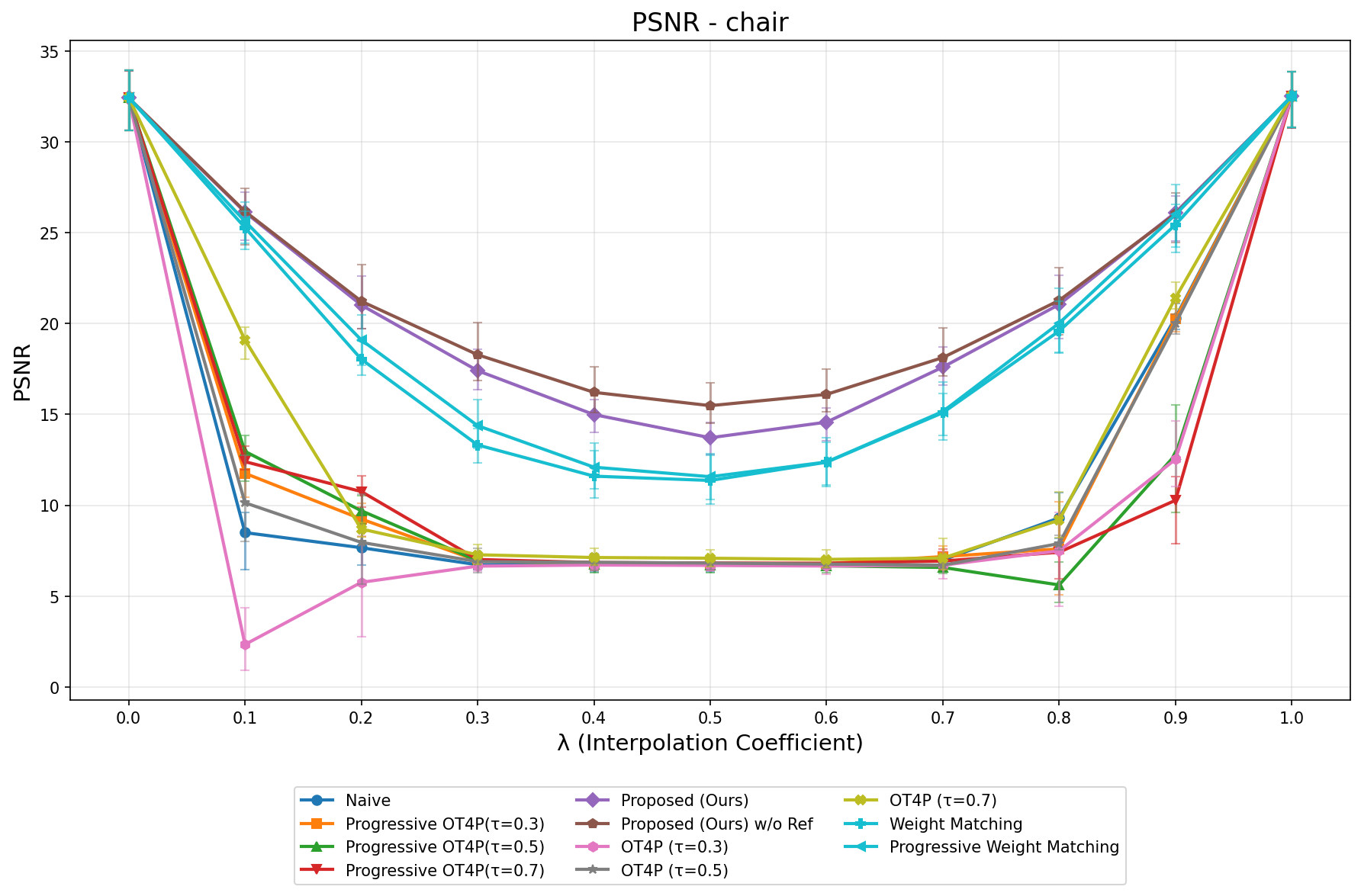}
        \caption{Chair}
    \end{subfigure}
    \hfill
    \begin{subfigure}{0.24\linewidth}
        \includegraphics[width=\linewidth]{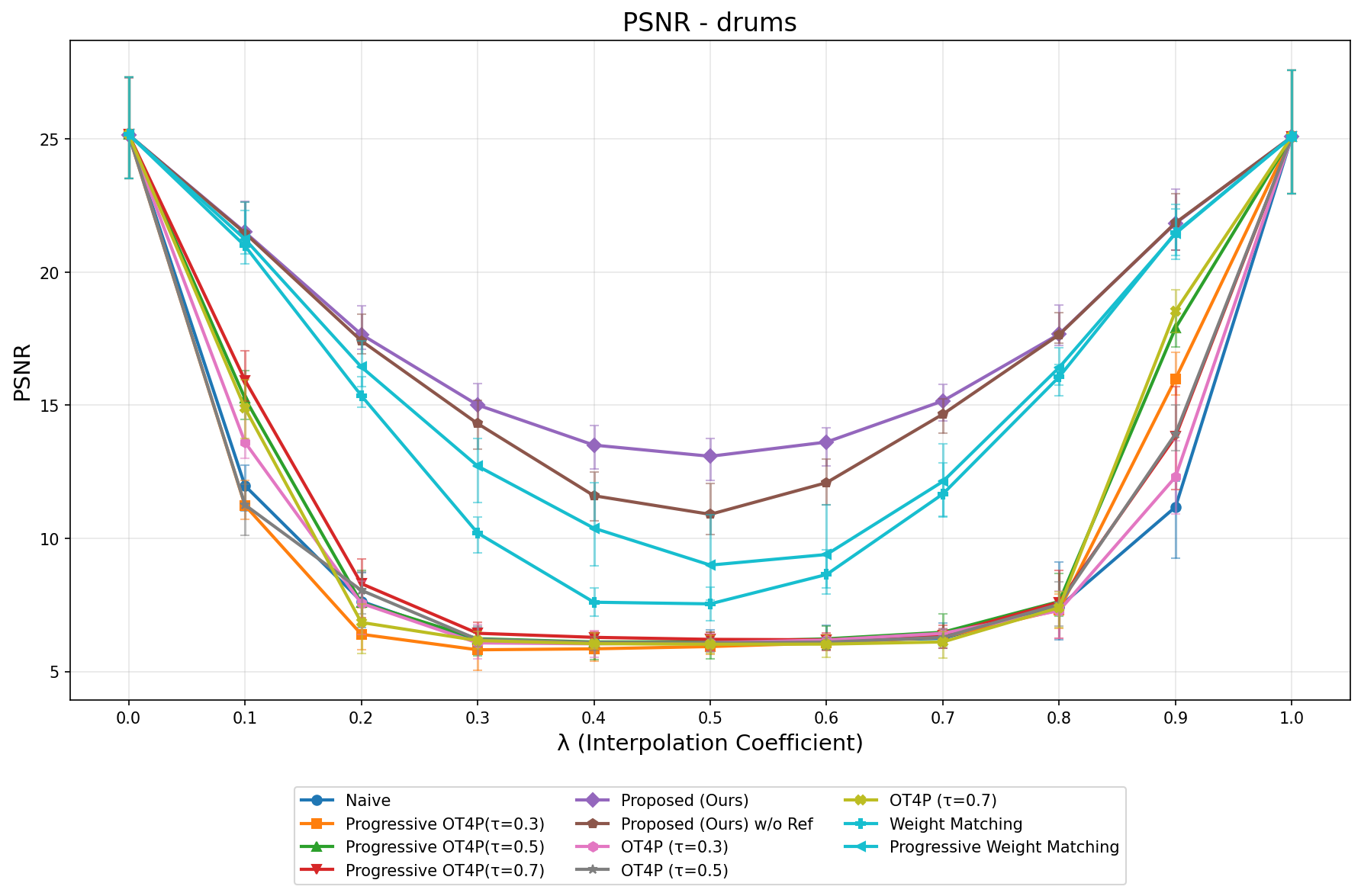}
        \caption{Drums}
    \end{subfigure}
    \hfill
    \begin{subfigure}{0.24\linewidth}
        \includegraphics[width=\linewidth]{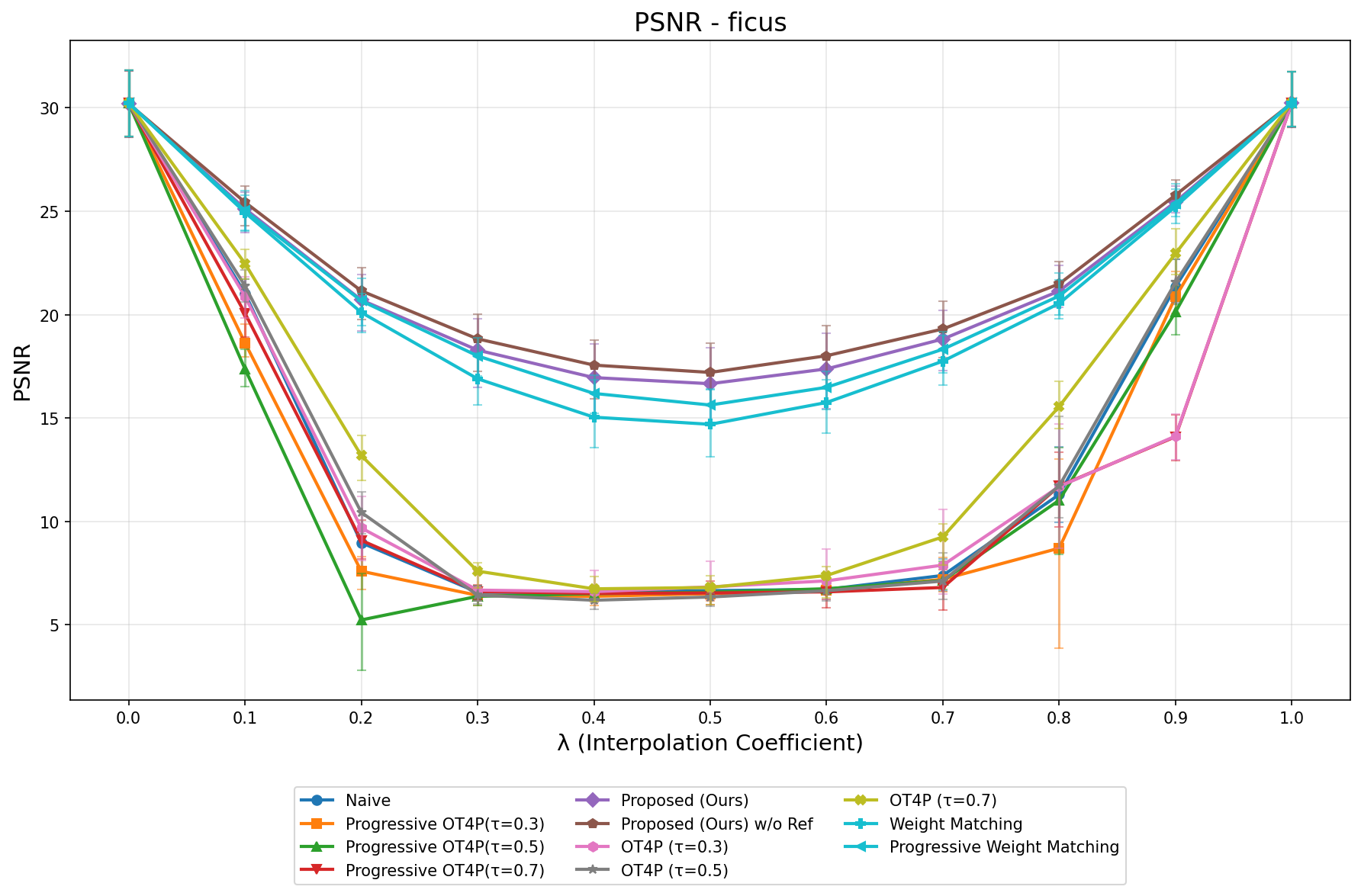}
        \caption{Ficus}
    \end{subfigure}
    \vspace{2mm}
    \begin{subfigure}{0.24\linewidth}
        \includegraphics[width=\linewidth]{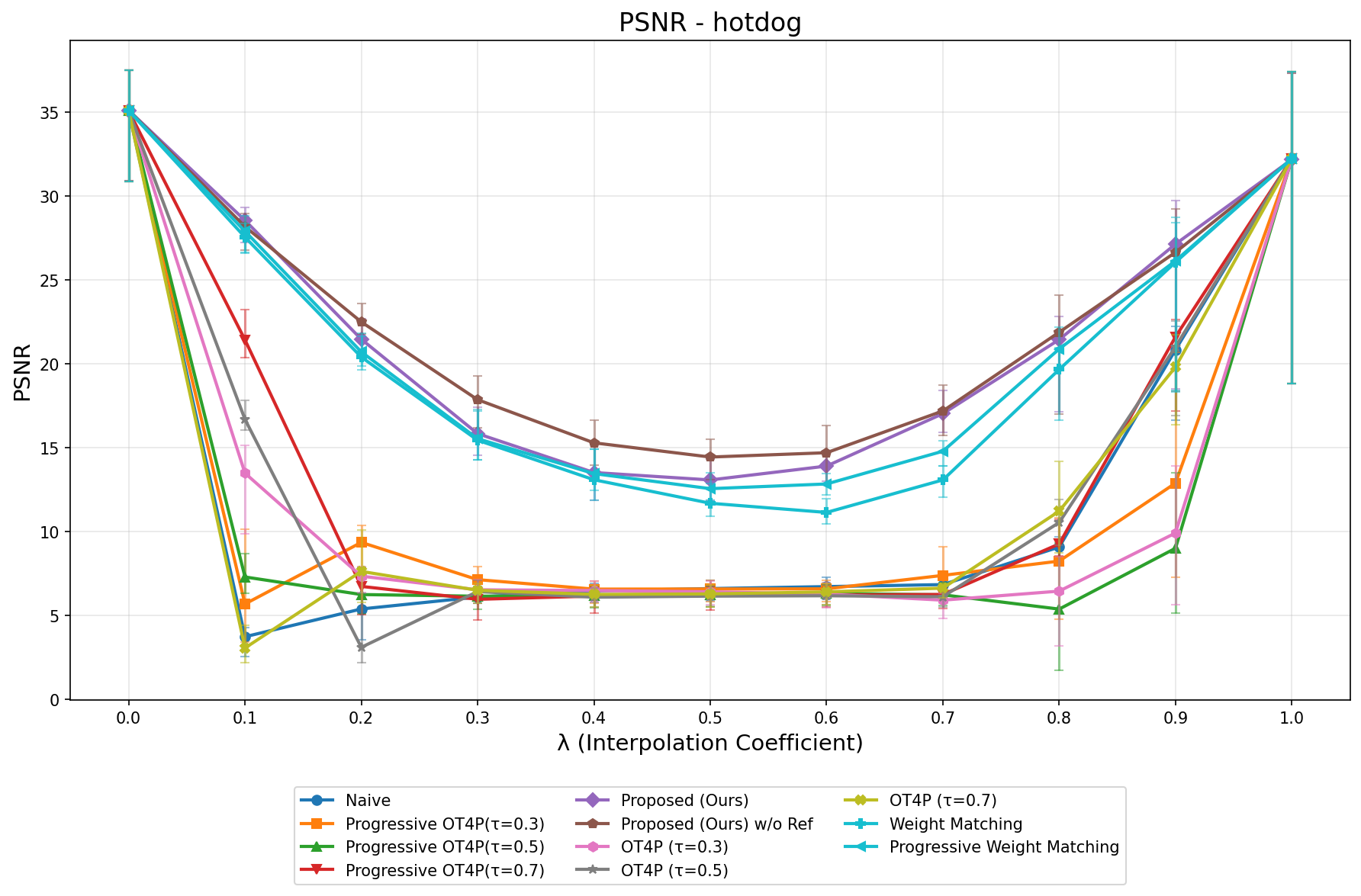}
        \caption{Hotdog}
    \end{subfigure}
    \hfill
    \begin{subfigure}{0.24\linewidth}
        \includegraphics[width=\linewidth]{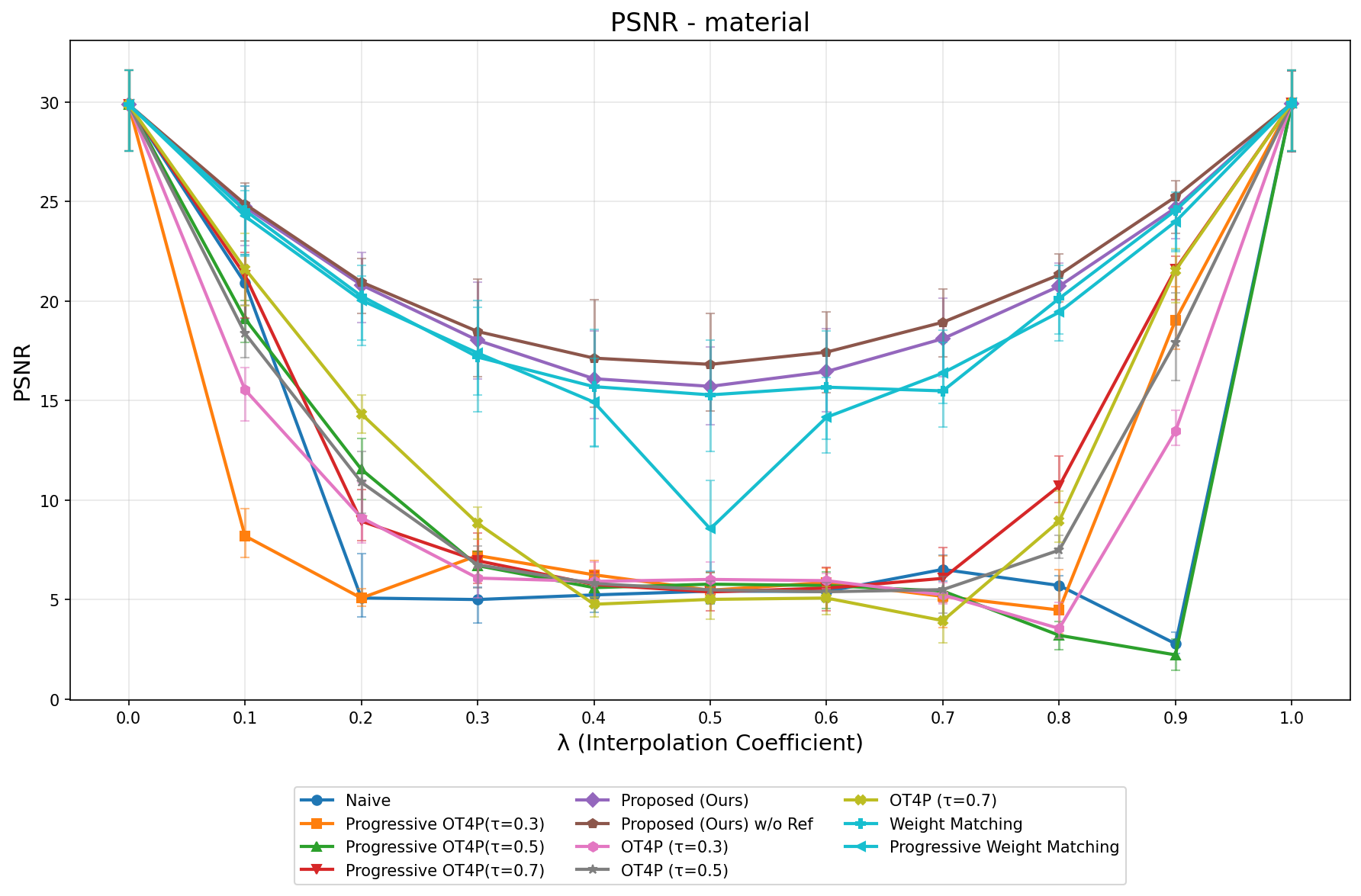}
        \caption{Materials}
    \end{subfigure}
    \hfill
    \begin{subfigure}{0.24\linewidth}
        \includegraphics[width=\linewidth]{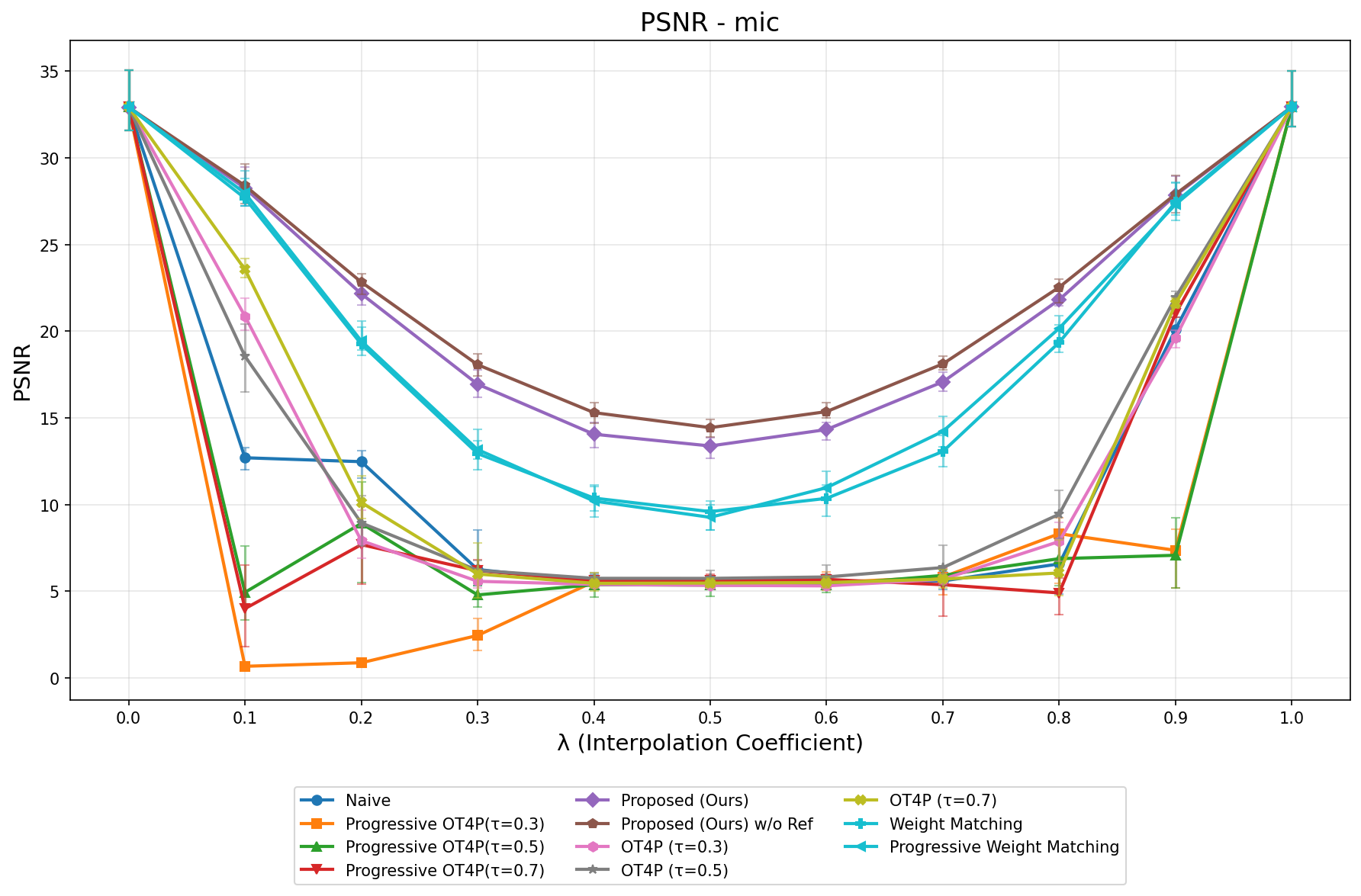}
        \caption{Mic}
    \end{subfigure}
    \hfill
    \begin{subfigure}{0.24\linewidth}
        \includegraphics[width=\linewidth]{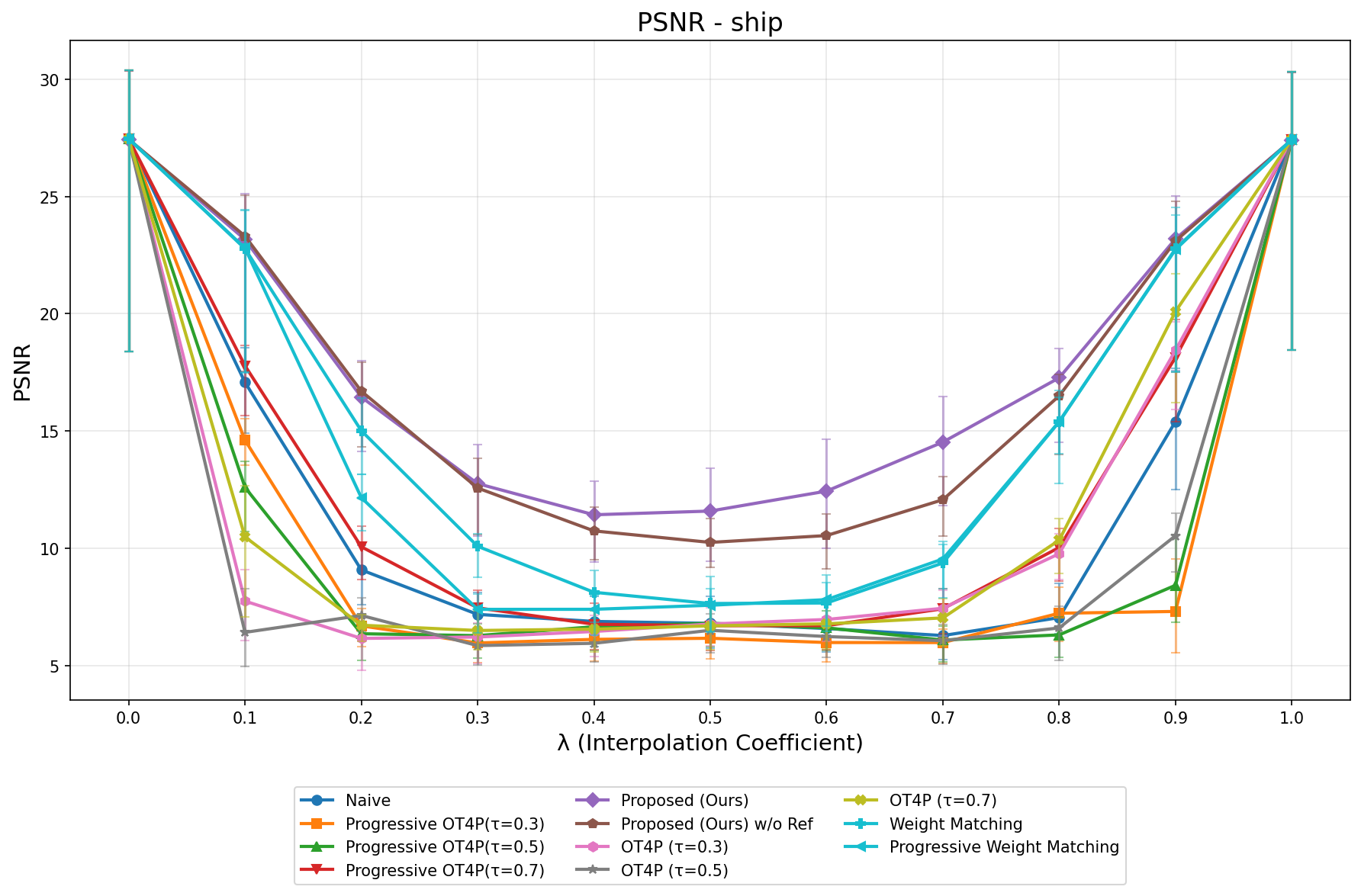}
        \caption{Ship}
    \end{subfigure}
    \caption{PSNR along the linear interpolation path for all eight scenes of the NeRF Synthetic dataset. Higher is better.}
    \label{fig:nerf_lmc_psnr_all_datasets}
\end{figure}

Figure~\ref{fig:nerf_lmc_psnr_all_datasets} shows the PSNR curves.
The proposed method achieves a higher PSNR over a broad range of interpolation coefficients, with the largest differences appearing near the middle of the interpolation path.

We next present the qualitative interpolation results for each scene.
The figures show rendered images at $\lambda \in \{0.0, 0.25, 0.50, 0.75, 1.0\}$ for different alignment methods.
When the alignment is poor, the intermediate models often produce blurry or distorted renderings, even though both endpoints are valid.

\begin{figure}[t]
\centering
\resizebox{0.99\columnwidth}{!}{%
\setlength{\tabcolsep}{2pt}
\begin{tabular}{lccccc|c}
\toprule
& $\lambda=0.0$ & $\lambda=0.25$ & $\lambda=0.50$ & $\lambda=0.75$ & $\lambda=1.0$ & Ground Truth \\
\midrule
\rowlabel{(a)}
& \includegraphics[width=0.12\textwidth]{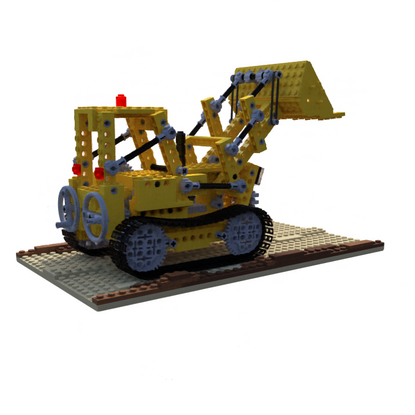}
& \includegraphics[width=0.12\textwidth]{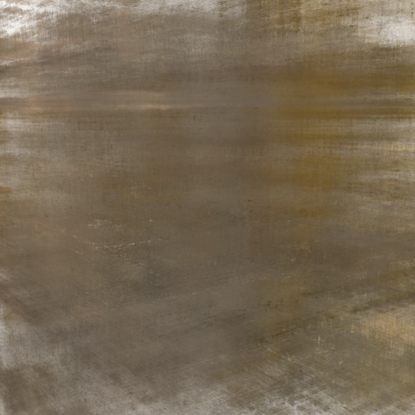}
& \includegraphics[width=0.12\textwidth]{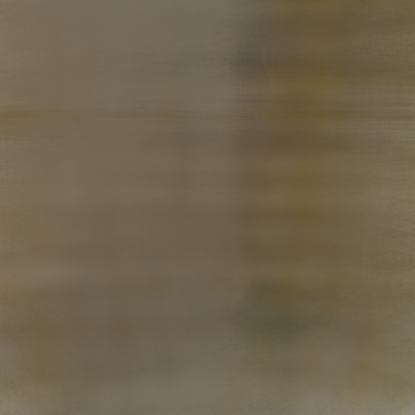}
& \includegraphics[width=0.12\textwidth]{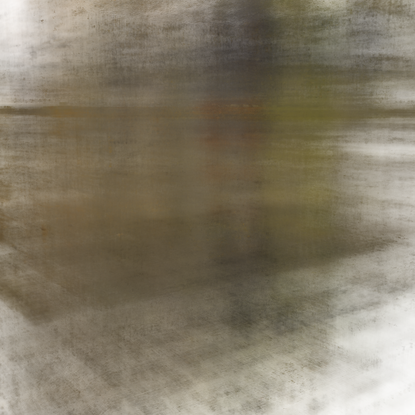}
& \includegraphics[width=0.12\textwidth]{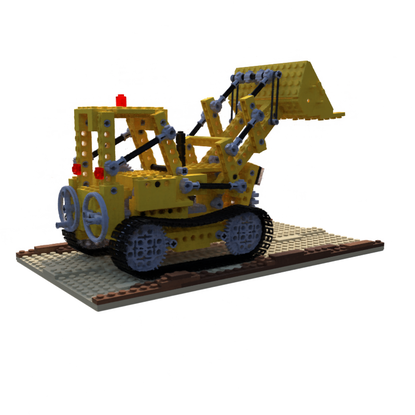}
& \includegraphics[width=0.12\textwidth]{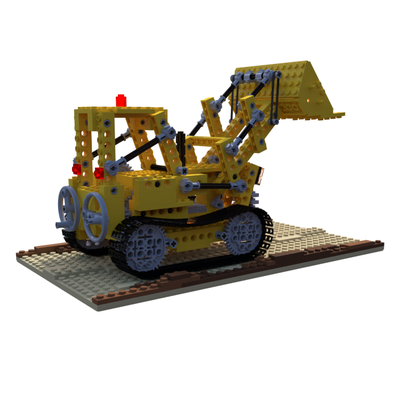} \\
\rowlabel{(b)}
& \includegraphics[width=0.12\textwidth]{figures/supp_figures/0.0.png}
& \includegraphics[width=0.12\textwidth]{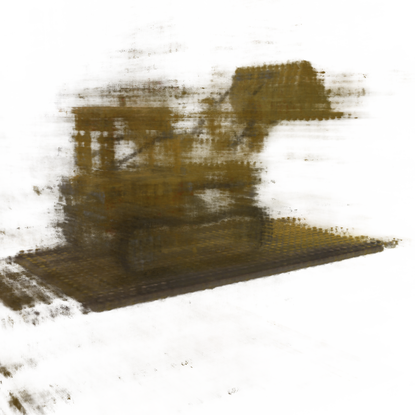}
& \includegraphics[width=0.12\textwidth]{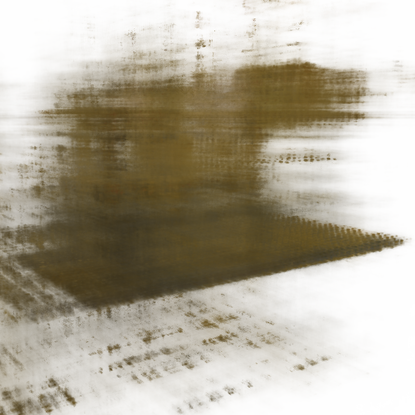}
& \includegraphics[width=0.12\textwidth]{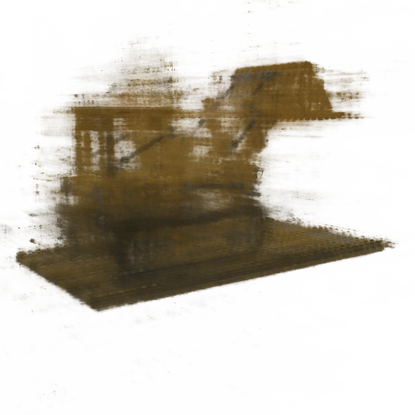}
& \includegraphics[width=0.12\textwidth]{figures/supp_figures/1.0.png}
& \includegraphics[width=0.12\textwidth]{figures/supp_figures/GT.png} \\
\rowlabel{(c)}
& \includegraphics[width=0.12\textwidth]{figures/supp_figures/0.0.png}
& \includegraphics[width=0.12\textwidth]{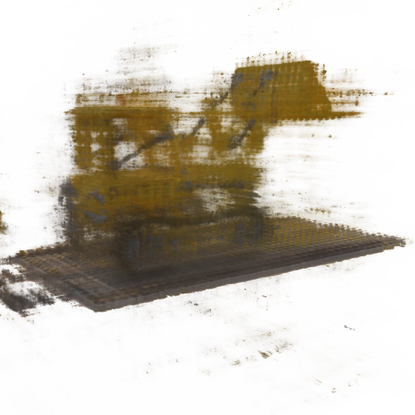}
& \includegraphics[width=0.12\textwidth]{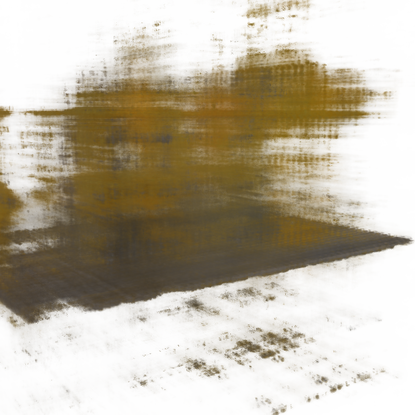}
& \includegraphics[width=0.12\textwidth]{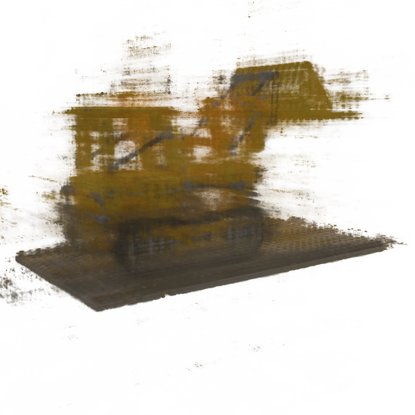}
& \includegraphics[width=0.12\textwidth]{figures/supp_figures/1.0.png}
& \includegraphics[width=0.12\textwidth]{figures/supp_figures/GT.png} \\
\rowlabel{(d)}
& \includegraphics[width=0.12\textwidth]{figures/supp_figures/0.0.png}
& \includegraphics[width=0.12\textwidth]{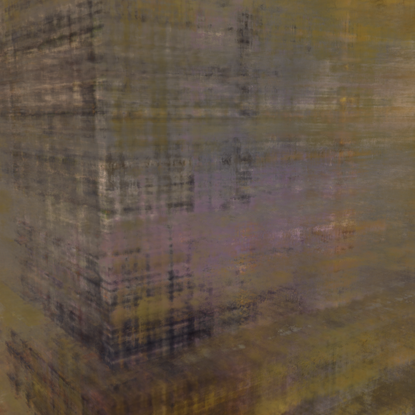}
& \includegraphics[width=0.12\textwidth]{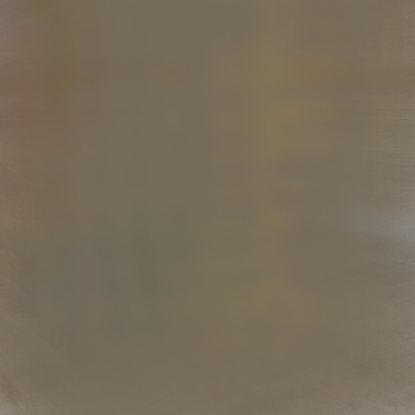}
& \includegraphics[width=0.12\textwidth]{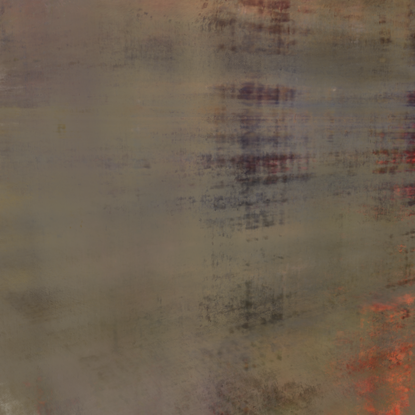}
& \includegraphics[width=0.12\textwidth]{figures/supp_figures/1.0.png}
& \includegraphics[width=0.12\textwidth]{figures/supp_figures/GT.png} \\
\rowlabel{(e)}
& \includegraphics[width=0.12\textwidth]{figures/supp_figures/0.0.png}
& \includegraphics[width=0.12\textwidth]{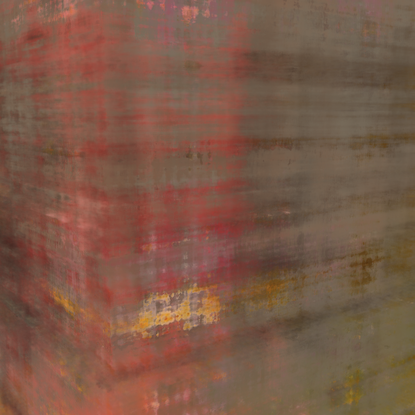}
& \includegraphics[width=0.12\textwidth]{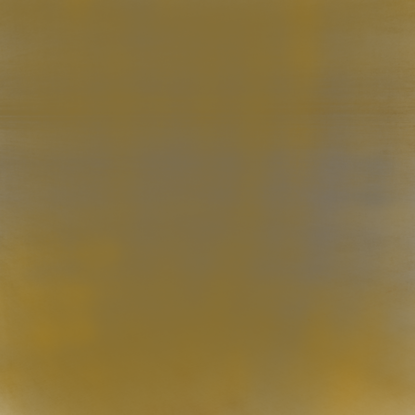}
& \includegraphics[width=0.12\textwidth]{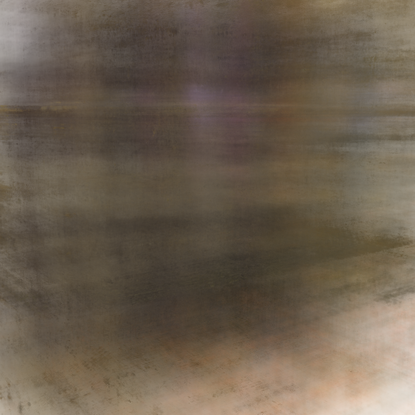}
& \includegraphics[width=0.12\textwidth]{figures/supp_figures/1.0.png}
& \includegraphics[width=0.12\textwidth]{figures/supp_figures/GT.png} \\
\rowlabel{(f)}
& \includegraphics[width=0.12\textwidth]{figures/supp_figures/0.0.png}
& \includegraphics[width=0.12\textwidth]{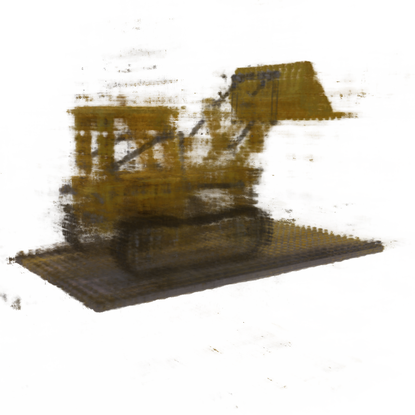}
& \includegraphics[width=0.12\textwidth]{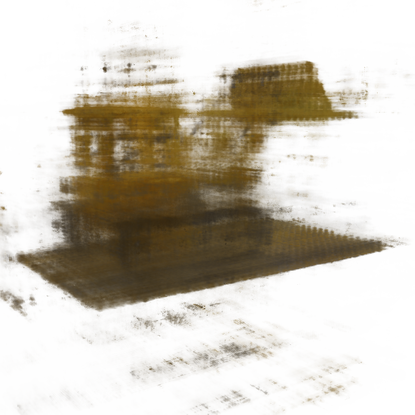}
& \includegraphics[width=0.12\textwidth]{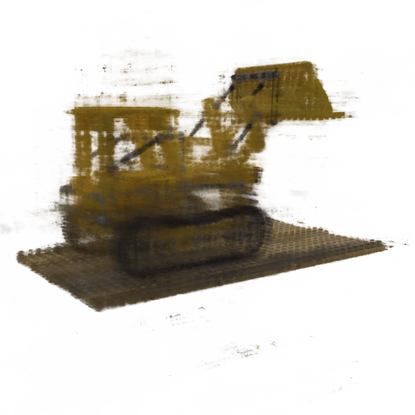}
& \includegraphics[width=0.12\textwidth]{figures/supp_figures/1.0.png}
& \includegraphics[width=0.12\textwidth]{figures/supp_figures/GT.png} \\
\rowlabel{(g)}
& \includegraphics[width=0.12\textwidth]{figures/supp_figures/0.0.png}
& \includegraphics[width=0.12\textwidth]{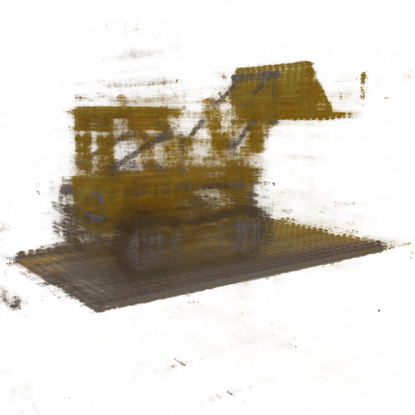}
& \includegraphics[width=0.12\textwidth]{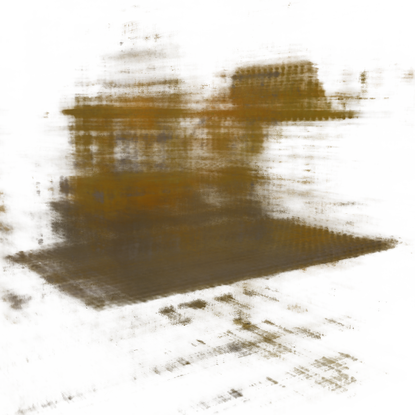}
& \includegraphics[width=0.12\textwidth]{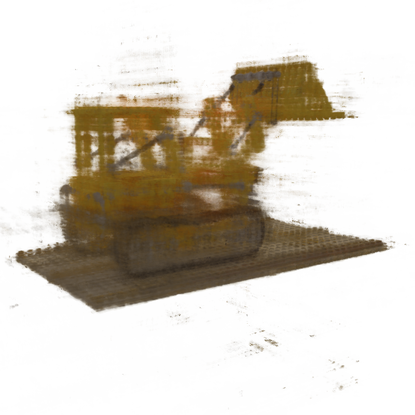}
& \includegraphics[width=0.12\textwidth]{figures/supp_figures/1.0.png}
& \includegraphics[width=0.12\textwidth]{figures/supp_figures/GT.png} \\
\bottomrule
\end{tabular}
}
\caption{Interpolation results at different values of $\lambda$ for a novel test view. 
(a) Naive, (b) Weight Matching, (c) Progressive Weight Matching, (d) OT4P, (e) Progressive OT4P, (f) Proposed w/o Ref, and (g) Proposed.}
\label{fig:nerf_lmc_different_view}
\end{figure}

Figure~\ref{fig:nerf_lmc_different_view} shows qualitative interpolation results for the Lego scene from another test view.
Naive interpolation produces severe artifacts in the intermediate models, and existing matching methods alleviate them only partially.
In contrast, the proposed method preserves the object structure and appearance more consistently along the interpolation path.
The variant with sequential refinement further stabilizes the transition.

\begin{figure}[t]
\centering
\resizebox{0.99\columnwidth}{!}{%
\setlength{\tabcolsep}{2pt}
\begin{tabular}{lccccc|c}
\toprule
& $\lambda=0.0$ & $\lambda=0.25$ & $\lambda=0.50$ & $\lambda=0.75$ & $\lambda=1.0$ & Ground Truth \\
\midrule
\rowlabel{(a)}
& \includegraphics[width=0.12\textwidth]{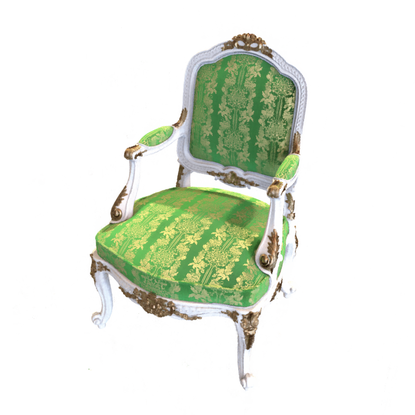}
& \includegraphics[width=0.12\textwidth]{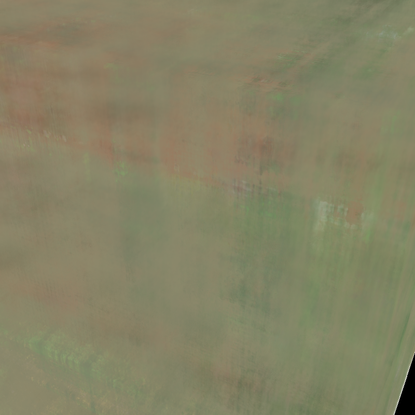}
& \includegraphics[width=0.12\textwidth]{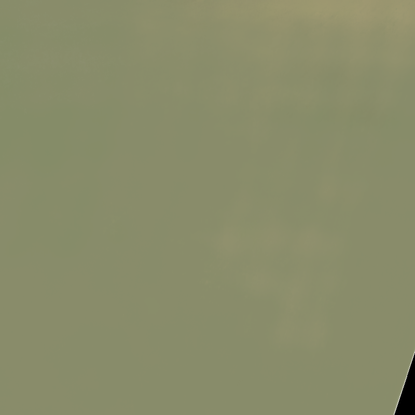}
& \includegraphics[width=0.12\textwidth]{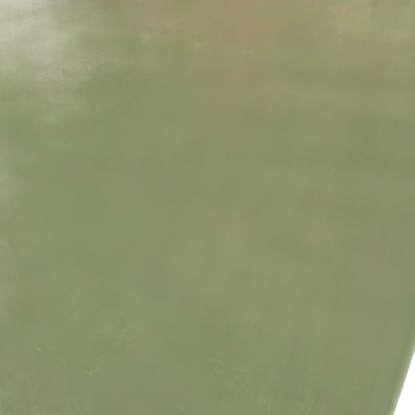}
& \includegraphics[width=0.12\textwidth]{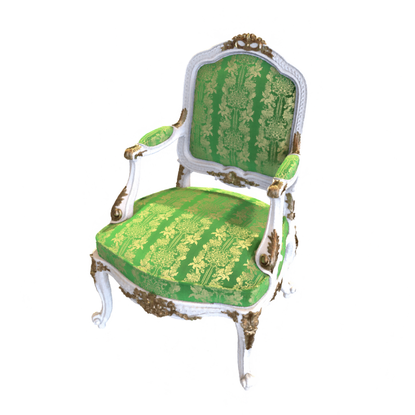}
& \includegraphics[width=0.12\textwidth]{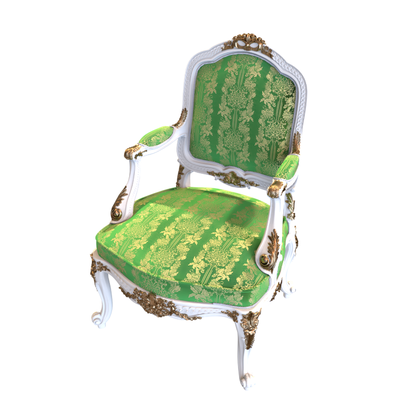} \\
\rowlabel{(b)}
& \includegraphics[width=0.12\textwidth]{figures/supp_figures/chair/0.0.png}
& \includegraphics[width=0.12\textwidth]{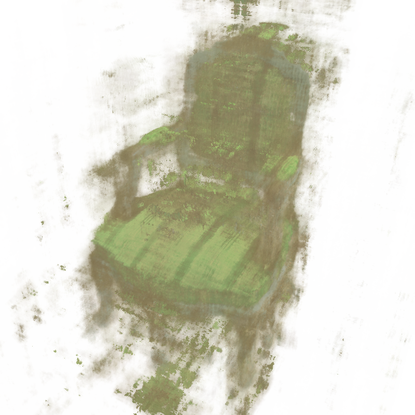}
& \includegraphics[width=0.12\textwidth]{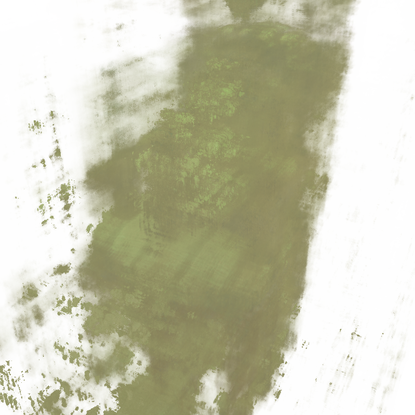}
& \includegraphics[width=0.12\textwidth]{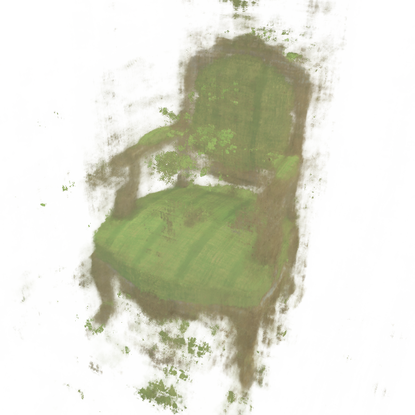}
& \includegraphics[width=0.12\textwidth]{figures/supp_figures/chair/1.0.png}
& \includegraphics[width=0.12\textwidth]{figures/supp_figures/chair/GT.png} \\
\rowlabel{(c)}
& \includegraphics[width=0.12\textwidth]{figures/supp_figures/chair/0.0.png}
& \includegraphics[width=0.12\textwidth]{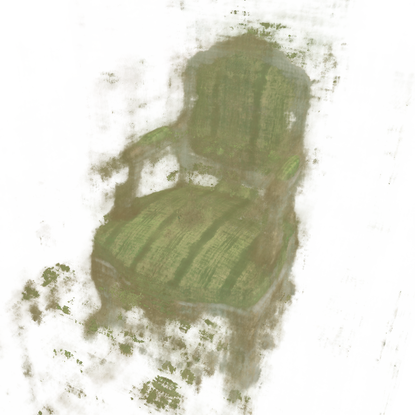}
& \includegraphics[width=0.12\textwidth]{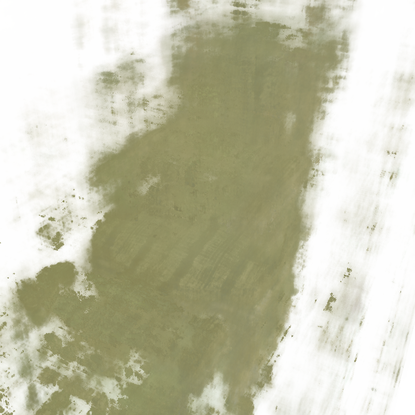}
& \includegraphics[width=0.12\textwidth]{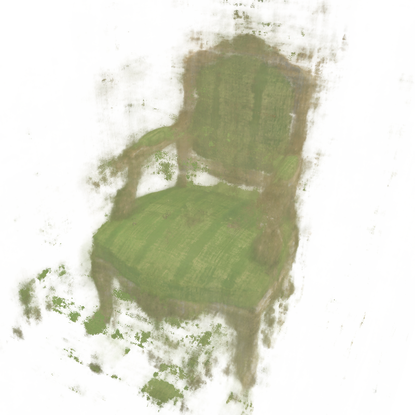}
& \includegraphics[width=0.12\textwidth]{figures/supp_figures/chair/1.0.png}
& \includegraphics[width=0.12\textwidth]{figures/supp_figures/chair/GT.png} \\
\rowlabel{(d)}
& \includegraphics[width=0.12\textwidth]{figures/supp_figures/chair/0.0.png}
& \includegraphics[width=0.12\textwidth]{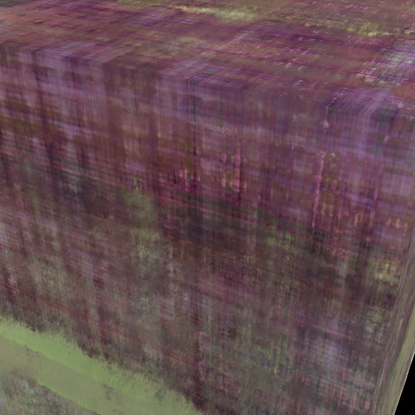}
& \includegraphics[width=0.12\textwidth]{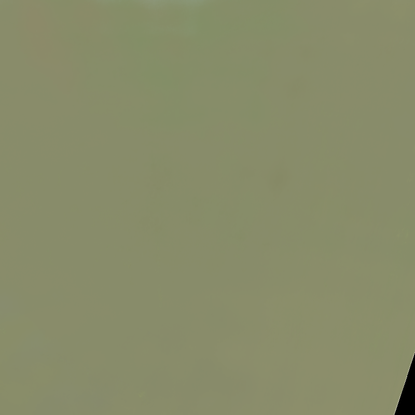}
& \includegraphics[width=0.12\textwidth]{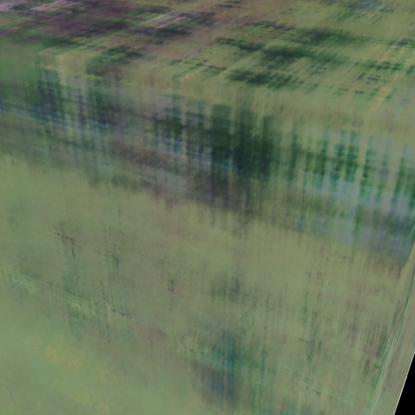}
& \includegraphics[width=0.12\textwidth]{figures/supp_figures/chair/1.0.png}
& \includegraphics[width=0.12\textwidth]{figures/supp_figures/chair/GT.png} \\
\rowlabel{(e)}
& \includegraphics[width=0.12\textwidth]{figures/supp_figures/chair/0.0.png}
& \includegraphics[width=0.12\textwidth]{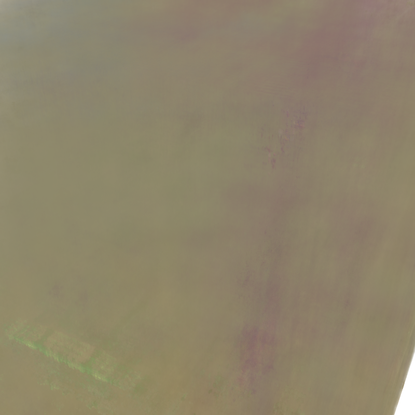}
& \includegraphics[width=0.12\textwidth]{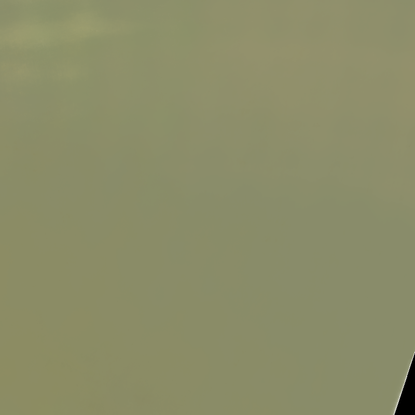}
& \includegraphics[width=0.12\textwidth]{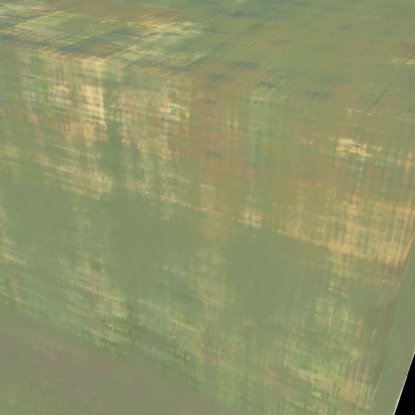}
& \includegraphics[width=0.12\textwidth]{figures/supp_figures/chair/1.0.png}
& \includegraphics[width=0.12\textwidth]{figures/supp_figures/chair/GT.png} \\
\rowlabel{(f)}
& \includegraphics[width=0.12\textwidth]{figures/supp_figures/chair/0.0.png}
& \includegraphics[width=0.12\textwidth]{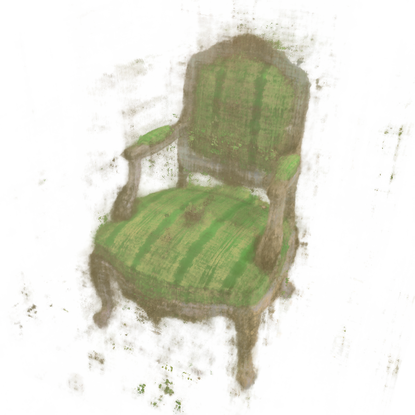}
& \includegraphics[width=0.12\textwidth]{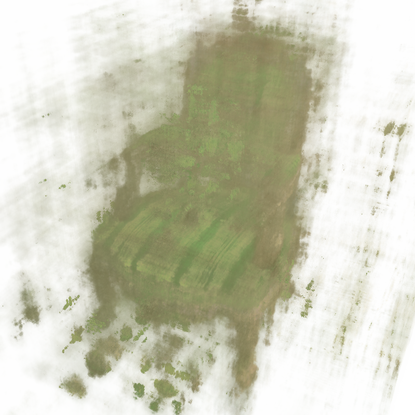}
& \includegraphics[width=0.12\textwidth]{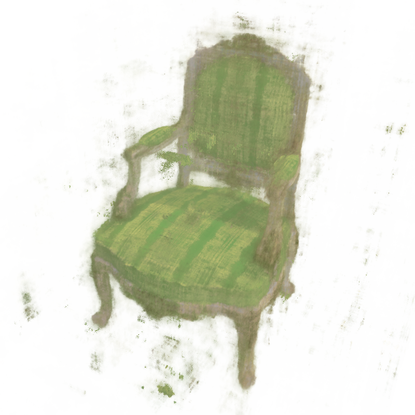}
& \includegraphics[width=0.12\textwidth]{figures/supp_figures/chair/1.0.png}
& \includegraphics[width=0.12\textwidth]{figures/supp_figures/chair/GT.png} \\
\rowlabel{(g)}
& \includegraphics[width=0.12\textwidth]{figures/supp_figures/chair/0.0.png}
& \includegraphics[width=0.12\textwidth]{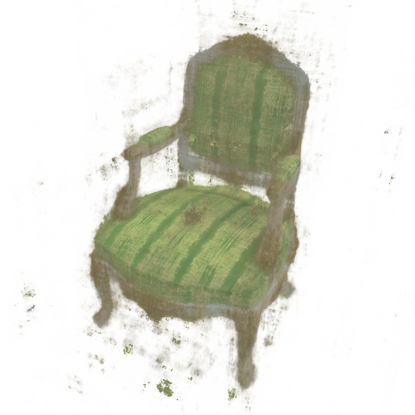}
& \includegraphics[width=0.12\textwidth]{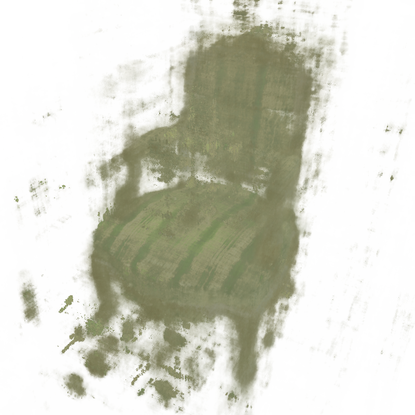}
& \includegraphics[width=0.12\textwidth]{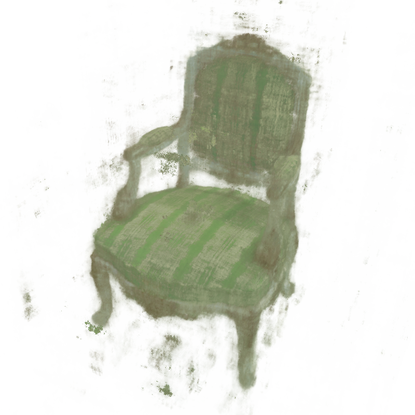}
& \includegraphics[width=0.12\textwidth]{figures/supp_figures/chair/1.0.png}
& \includegraphics[width=0.12\textwidth]{figures/supp_figures/chair/GT.png} \\
\bottomrule
\end{tabular}
}
\caption{Example of interpolation results for different $\lambda$ values for the Chair scene. (a) Naive, (b) Weight Matching
(c) Progressive Weight Matching, (d) OT4P, (e) Progressive OT4P, (f) Proposed w/o Ref, (g) Proposed.}
\label{fig:supp_chair}
\end{figure}

\begin{figure}[t]
\centering
\resizebox{0.99\columnwidth}{!}{%
\setlength{\tabcolsep}{2pt}
\begin{tabular}{lccccc|c}
\toprule
& $\lambda=0.0$ & $\lambda=0.25$ & $\lambda=0.50$ & $\lambda=0.75$ & $\lambda=1.0$ & Ground Truth \\
\midrule
\rowlabel{(a)}
& \includegraphics[width=0.12\textwidth]{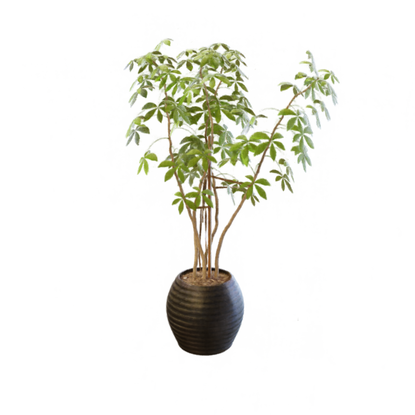}
& \includegraphics[width=0.12\textwidth]{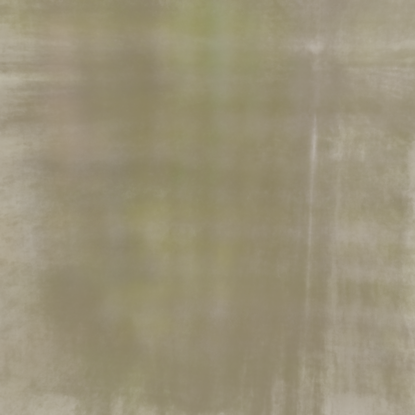}
& \includegraphics[width=0.12\textwidth]{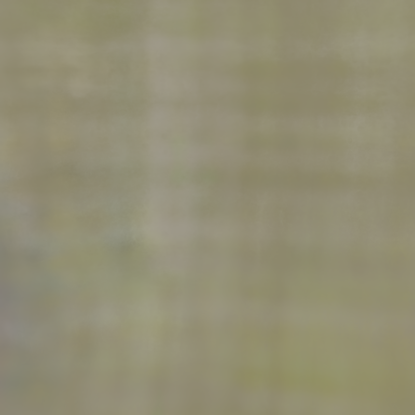}
& \includegraphics[width=0.12\textwidth]{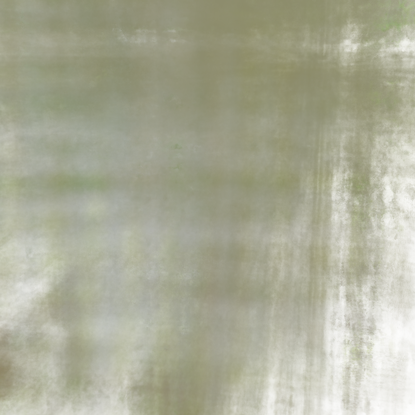}
& \includegraphics[width=0.12\textwidth]{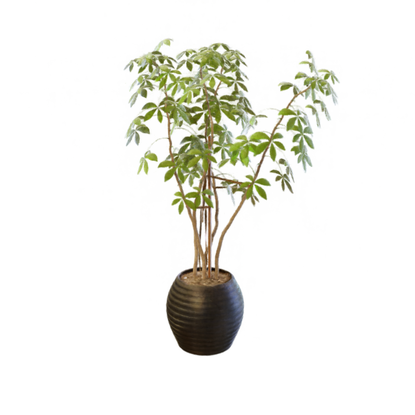}
& \includegraphics[width=0.12\textwidth]{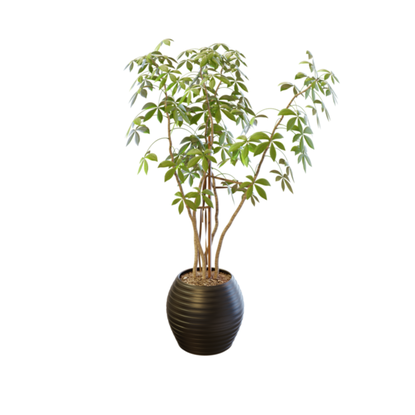} \\
\rowlabel{(b)}
& \includegraphics[width=0.12\textwidth]{figures/supp_figures/ficus/0.0.png}
& \includegraphics[width=0.12\textwidth]{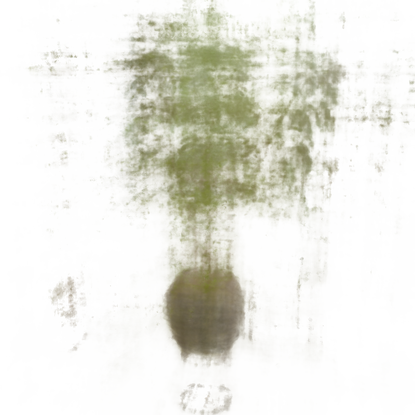}
& \includegraphics[width=0.12\textwidth]{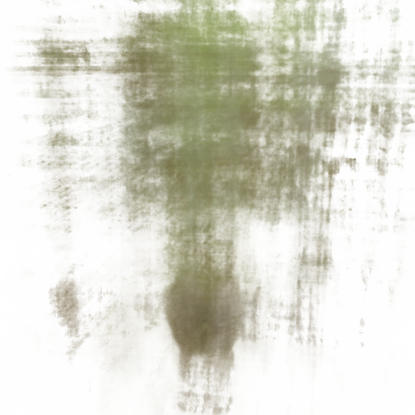}
& \includegraphics[width=0.12\textwidth]{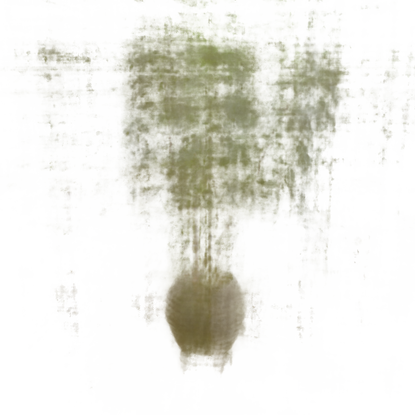}
& \includegraphics[width=0.12\textwidth]{figures/thesis_images/Ficus/1.0.png}
& \includegraphics[width=0.12\textwidth]{figures/supp_figures/ficus/GT.png} \\
\rowlabel{(c)}
& \includegraphics[width=0.12\textwidth]{figures/supp_figures/ficus/0.0.png}
& \includegraphics[width=0.12\textwidth]{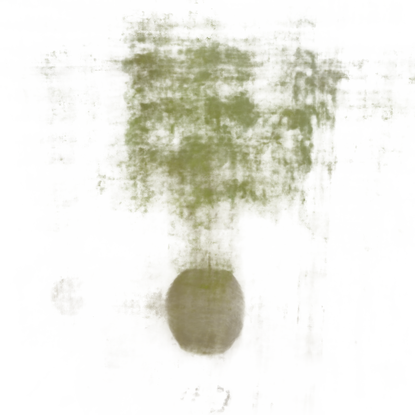}
& \includegraphics[width=0.12\textwidth]{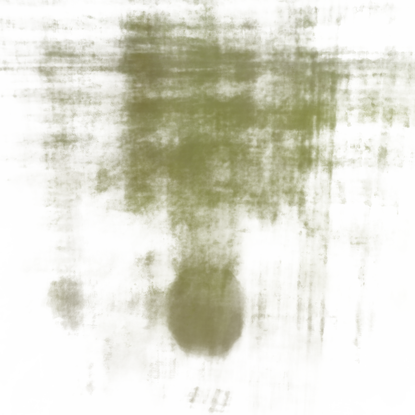}
& \includegraphics[width=0.12\textwidth]{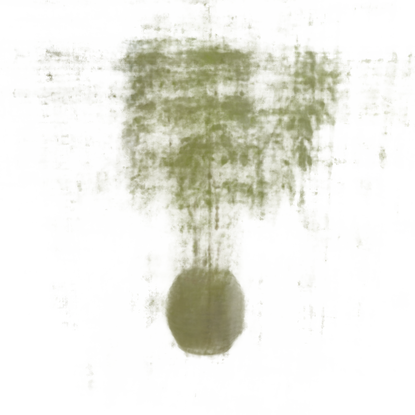}
& \includegraphics[width=0.12\textwidth]{figures/thesis_images/Ficus/1.0.png}
& \includegraphics[width=0.12\textwidth]{figures/supp_figures/ficus/GT.png} \\
\rowlabel{(d)}
& \includegraphics[width=0.12\textwidth]{figures/supp_figures/ficus/0.0.png}
& \includegraphics[width=0.12\textwidth]{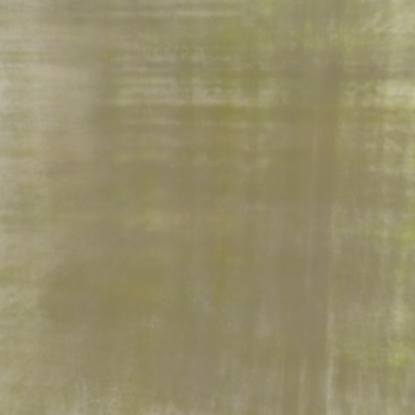}
& \includegraphics[width=0.12\textwidth]{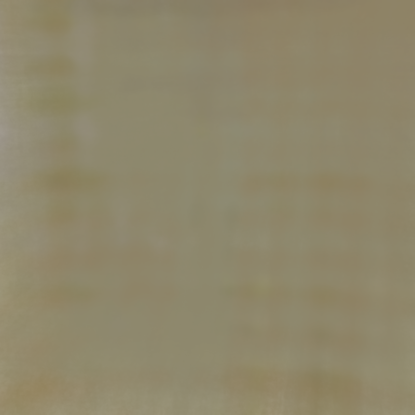}
& \includegraphics[width=0.12\textwidth]{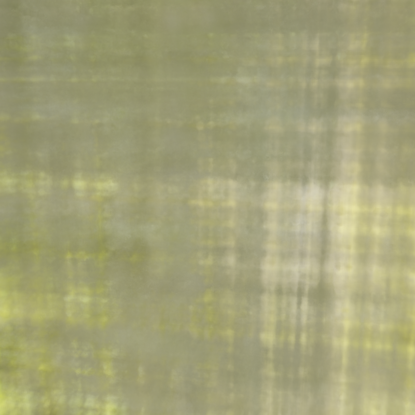}
& \includegraphics[width=0.12\textwidth]{figures/thesis_images/Ficus/1.0.png}
& \includegraphics[width=0.12\textwidth]{figures/supp_figures/ficus/GT.png} \\
\rowlabel{(e)}
& \includegraphics[width=0.12\textwidth]{figures/supp_figures/ficus/0.0.png}
& \includegraphics[width=0.12\textwidth]{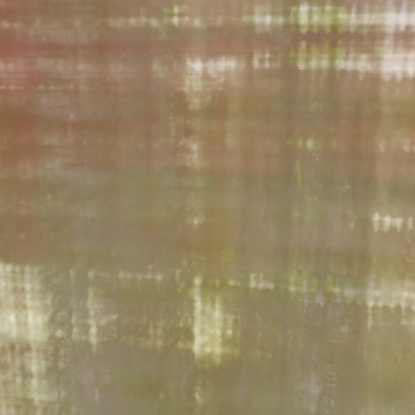}
& \includegraphics[width=0.12\textwidth]{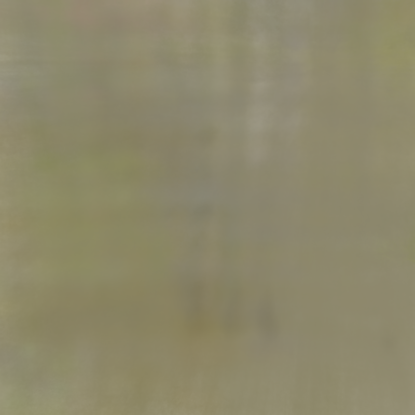}
& \includegraphics[width=0.12\textwidth]{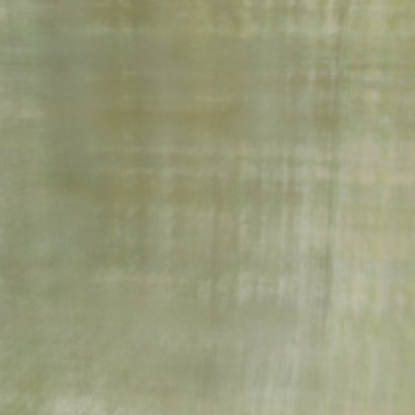}
& \includegraphics[width=0.12\textwidth]{figures/thesis_images/Ficus/1.0.png}
& \includegraphics[width=0.12\textwidth]{figures/supp_figures/ficus/GT.png} \\
\rowlabel{(f)}
& \includegraphics[width=0.12\textwidth]{figures/supp_figures/ficus/0.0.png}
& \includegraphics[width=0.12\textwidth]{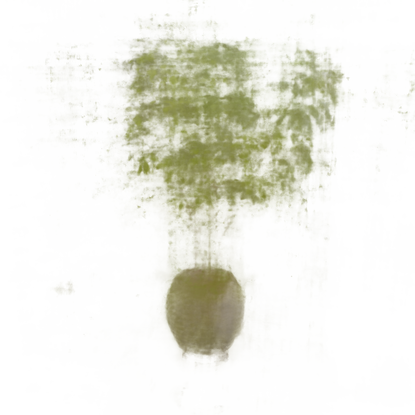}
& \includegraphics[width=0.12\textwidth]{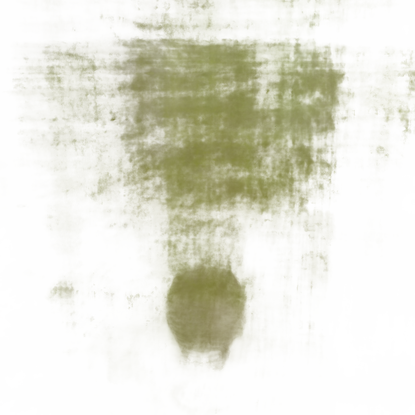}
& \includegraphics[width=0.12\textwidth]{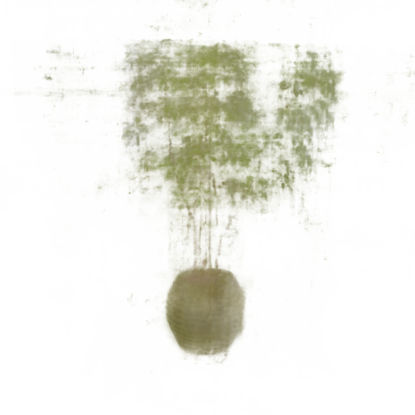}
& \includegraphics[width=0.12\textwidth]{figures/thesis_images/Ficus/1.0.png}
& \includegraphics[width=0.12\textwidth]{figures/supp_figures/ficus/GT.png} \\
\rowlabel{(g)}
& \includegraphics[width=0.12\textwidth]{figures/supp_figures/ficus/0.0.png}
& \includegraphics[width=0.12\textwidth]{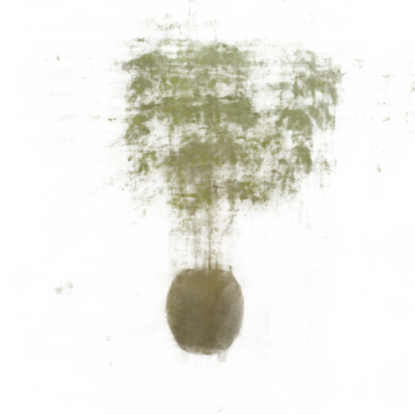}
& \includegraphics[width=0.12\textwidth]{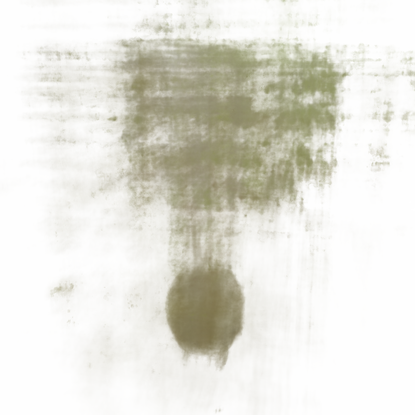}
& \includegraphics[width=0.12\textwidth]{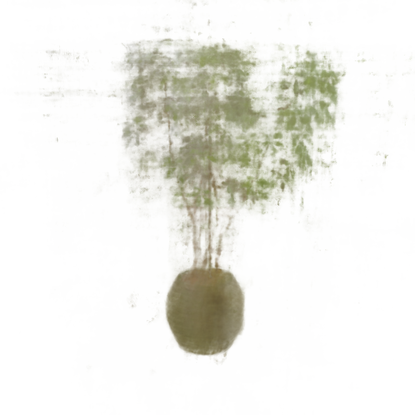}
& \includegraphics[width=0.12\textwidth]{figures/thesis_images/Ficus/1.0.png}
& \includegraphics[width=0.12\textwidth]{figures/supp_figures/ficus/GT.png} \\
\bottomrule
\end{tabular}
}
\caption{Example of interpolation results for different $\lambda$ values for the Ficus scene. (a) Naive, (b) Weight Matching
(c) Progressive Weight Matching, (d) OT4P, (e) Progressive OT4P, (f) Proposed w/o Ref, (g) Proposed.}
\label{fig:supp_ficus}
\end{figure}

\begin{figure}[t]
\centering
\resizebox{0.99\columnwidth}{!}{%
\setlength{\tabcolsep}{2pt}
\begin{tabular}{lccccc|c}
\toprule
& $\lambda=0.0$ & $\lambda=0.25$ & $\lambda=0.50$ & $\lambda=0.75$ & $\lambda=1.0$ & Ground Truth \\
\midrule
\rowlabel{(a)}
& \includegraphics[width=0.12\textwidth]{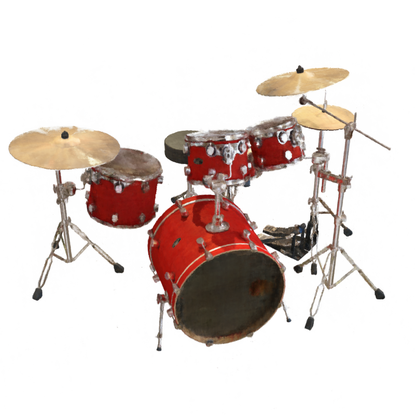}
& \includegraphics[width=0.12\textwidth]{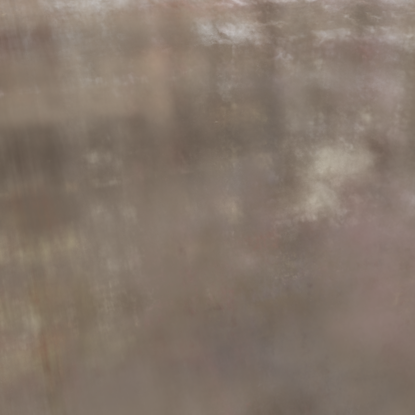}
& \includegraphics[width=0.12\textwidth]{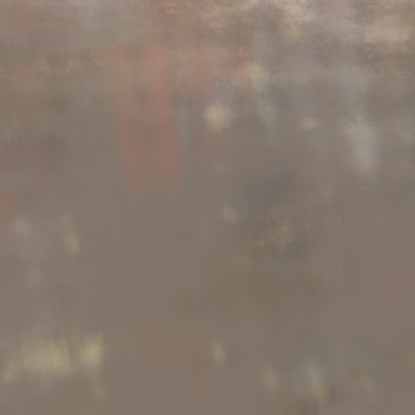}
& \includegraphics[width=0.12\textwidth]{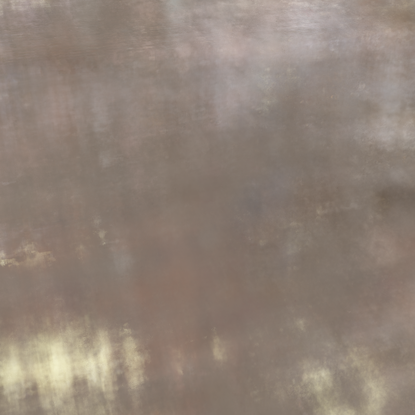}
& \includegraphics[width=0.12\textwidth]{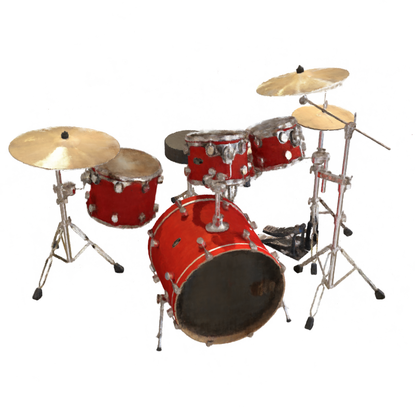}
& \includegraphics[width=0.12\textwidth]{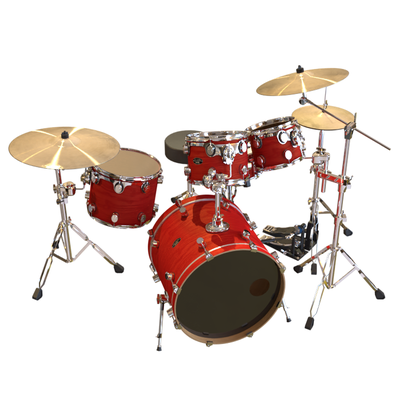} \\
\rowlabel{(b)}
& \includegraphics[width=0.12\textwidth]{figures/supp_figures/drums/0.0.png}
& \includegraphics[width=0.12\textwidth]{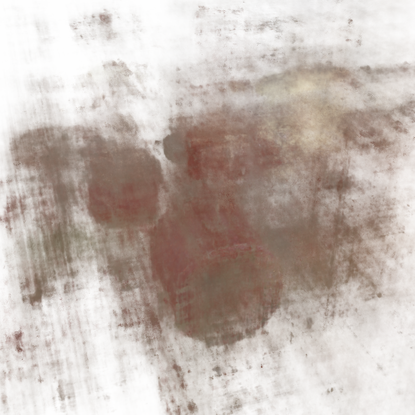}
& \includegraphics[width=0.12\textwidth]{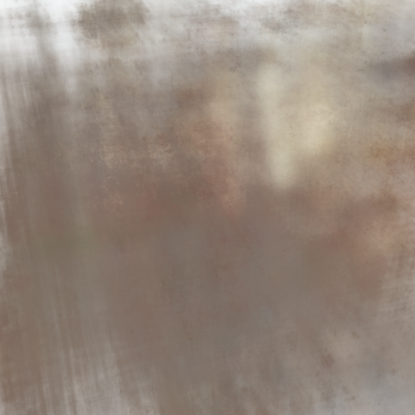}
& \includegraphics[width=0.12\textwidth]{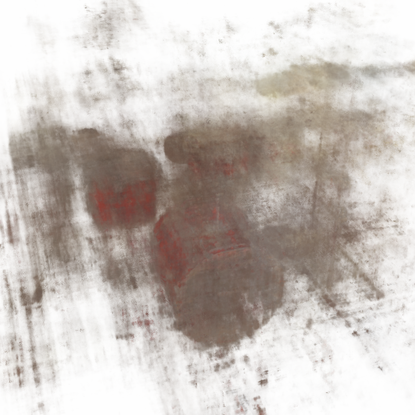}
& \includegraphics[width=0.12\textwidth]{figures/supp_figures/drums/1.0.png}
& \includegraphics[width=0.12\textwidth]{figures/supp_figures/drums/GT.png} \\
\rowlabel{(c)}
& \includegraphics[width=0.12\textwidth]{figures/supp_figures/drums/0.0.png}
& \includegraphics[width=0.12\textwidth]{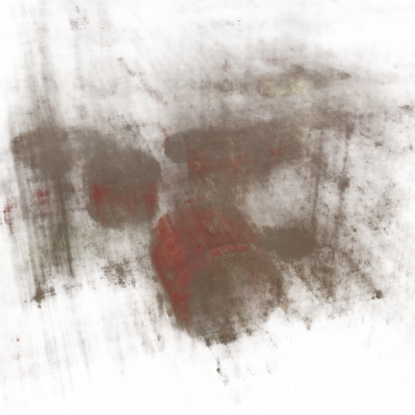}
& \includegraphics[width=0.12\textwidth]{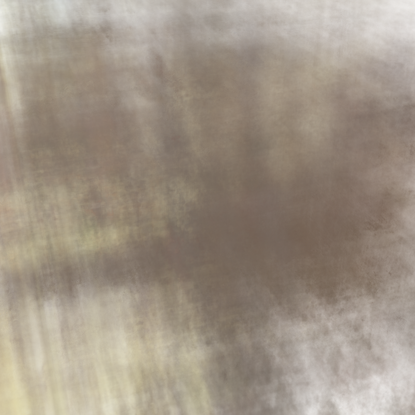}
& \includegraphics[width=0.12\textwidth]{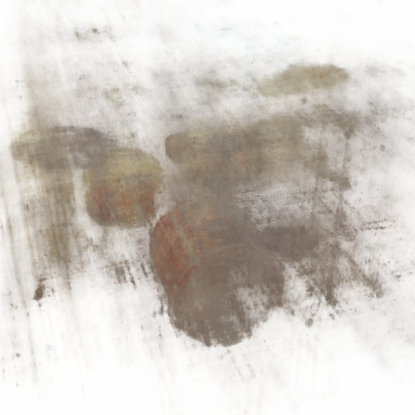}
& \includegraphics[width=0.12\textwidth]{figures/supp_figures/drums/1.0.png}
& \includegraphics[width=0.12\textwidth]{figures/supp_figures/drums/GT.png} \\
\rowlabel{(d)}
& \includegraphics[width=0.12\textwidth]{figures/supp_figures/drums/0.0.png}
& \includegraphics[width=0.12\textwidth]{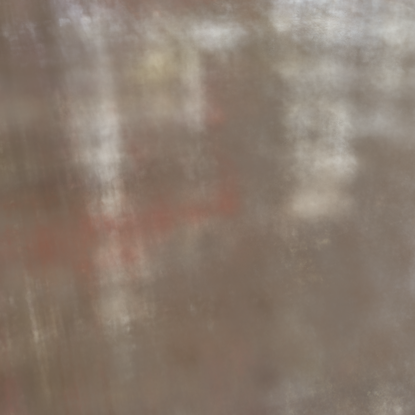}
& \includegraphics[width=0.12\textwidth]{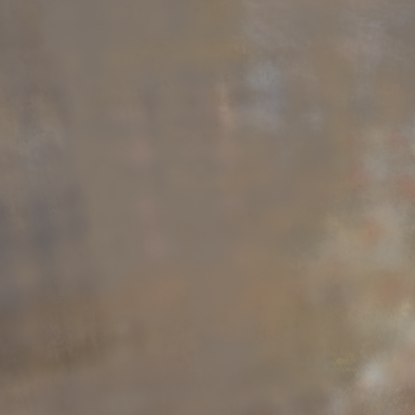}
& \includegraphics[width=0.12\textwidth]{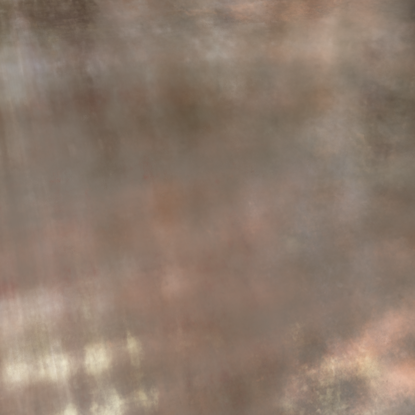}
& \includegraphics[width=0.12\textwidth]{figures/supp_figures/drums/1.0.png}
& \includegraphics[width=0.12\textwidth]{figures/supp_figures/drums/GT.png} \\
\rowlabel{(e)}
& \includegraphics[width=0.12\textwidth]{figures/supp_figures/drums/0.0.png}
& \includegraphics[width=0.12\textwidth]{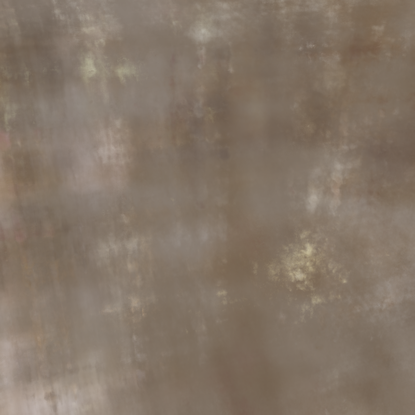}
& \includegraphics[width=0.12\textwidth]{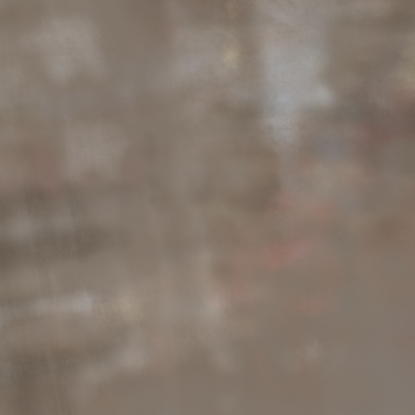}
& \includegraphics[width=0.12\textwidth]{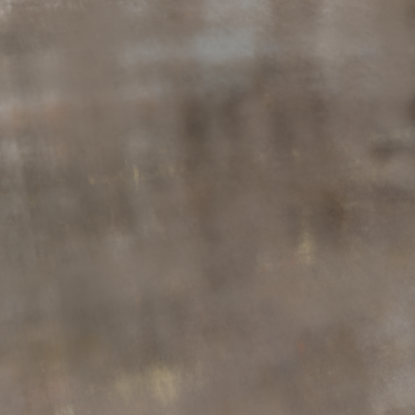}
& \includegraphics[width=0.12\textwidth]{figures/supp_figures/drums/1.0.png}
& \includegraphics[width=0.12\textwidth]{figures/supp_figures/drums/GT.png} \\
\rowlabel{(f)}
& \includegraphics[width=0.12\textwidth]{figures/supp_figures/drums/0.0.png}
& \includegraphics[width=0.12\textwidth]{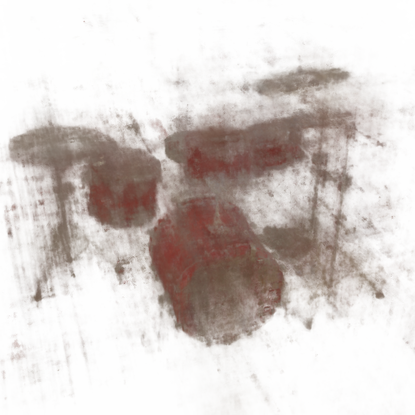}
& \includegraphics[width=0.12\textwidth]{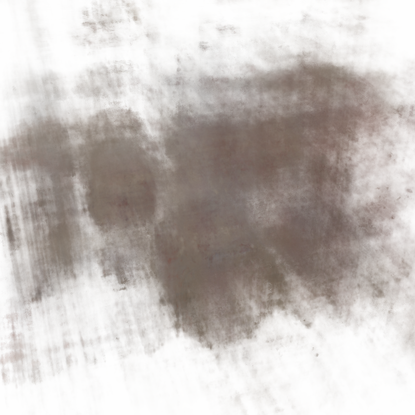}
& \includegraphics[width=0.12\textwidth]{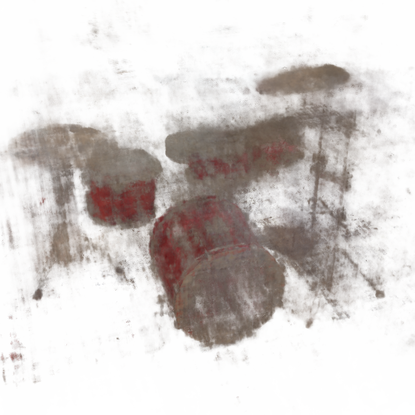}
& \includegraphics[width=0.12\textwidth]{figures/supp_figures/drums/1.0.png}
& \includegraphics[width=0.12\textwidth]{figures/supp_figures/drums/GT.png} \\
\rowlabel{(g)}
& \includegraphics[width=0.12\textwidth]{figures/supp_figures/drums/0.0.png}
& \includegraphics[width=0.12\textwidth]{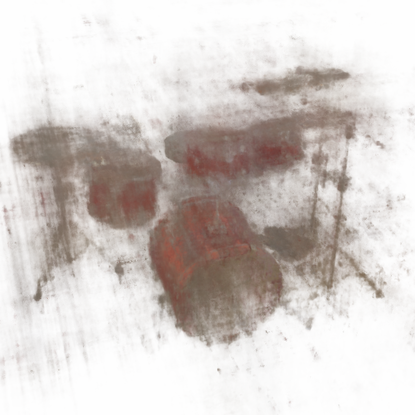}
& \includegraphics[width=0.12\textwidth]{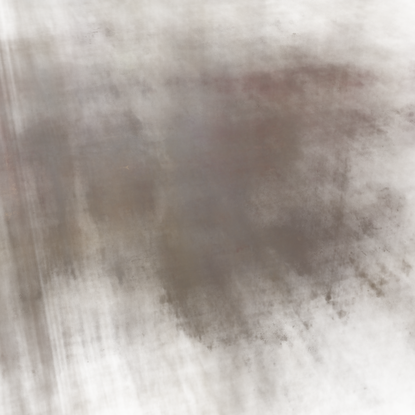}
& \includegraphics[width=0.12\textwidth]{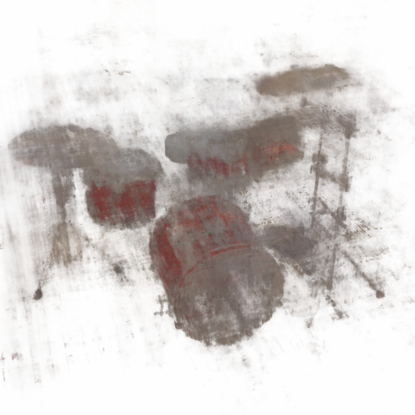}
& \includegraphics[width=0.12\textwidth]{figures/supp_figures/drums/1.0.png}
& \includegraphics[width=0.12\textwidth]{figures/supp_figures/drums/GT.png} \\
\bottomrule
\end{tabular}
}
\caption{Example of interpolation results for different $\lambda$ values for the Drums scene. (a) Naive, (b) Weight Matching
(c) Progressive Weight Matching, (d) OT4P, (e) Progressive OT4P, (f) Proposed w/o Ref, (g) Proposed.}
\label{fig:supp_drums}
\end{figure}

\begin{figure}[t]
\centering
\resizebox{0.99\columnwidth}{!}{%
\setlength{\tabcolsep}{2pt}
\begin{tabular}{lccccc|c}
\toprule
& $\lambda=0.0$ & $\lambda=0.25$ & $\lambda=0.50$ & $\lambda=0.75$ & $\lambda=1.0$ & Ground Truth \\
\midrule
\rowlabel{(a)}
& \includegraphics[width=0.12\textwidth]{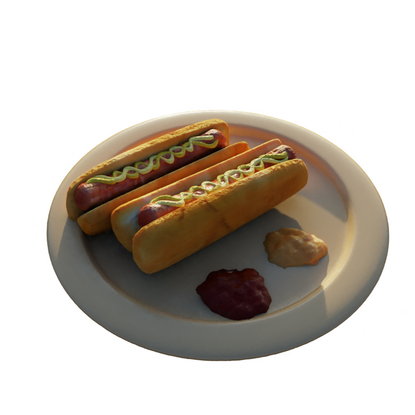}
& \includegraphics[width=0.12\textwidth]{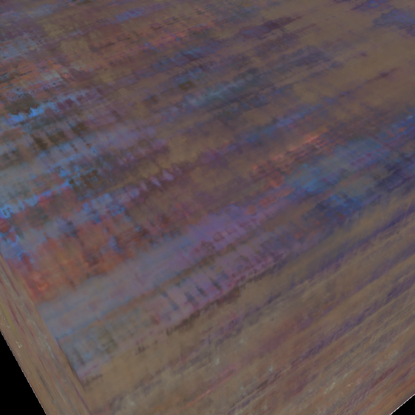}
& \includegraphics[width=0.12\textwidth]{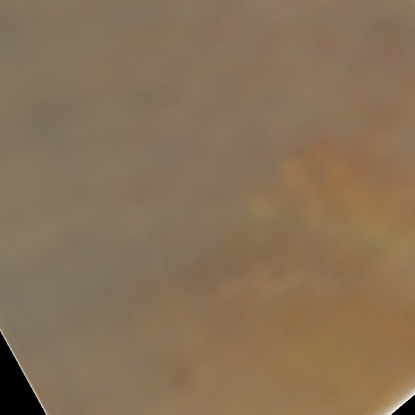}
& \includegraphics[width=0.12\textwidth]{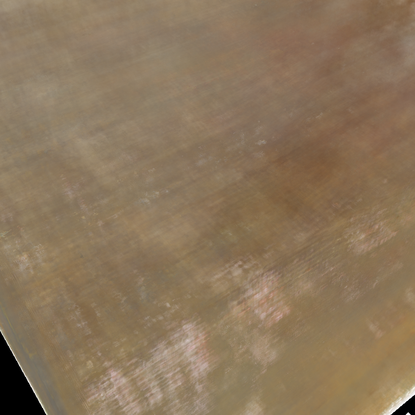}
& \includegraphics[width=0.12\textwidth]{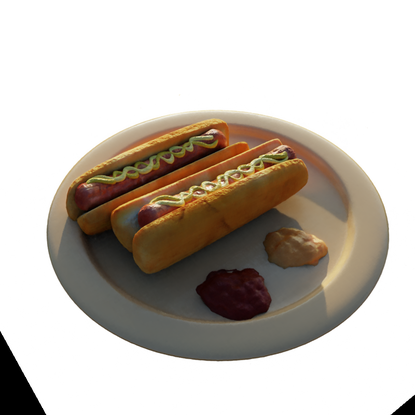}
& \includegraphics[width=0.12\textwidth]{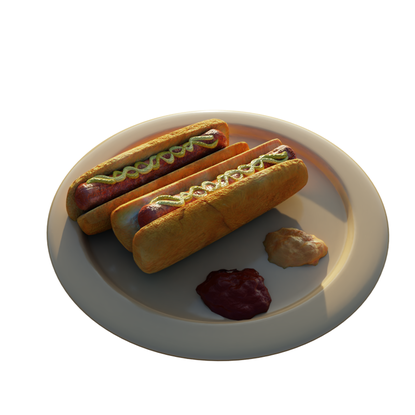} \\
\rowlabel{(b)}
& \includegraphics[width=0.12\textwidth]{figures/supp_figures/hotdog/0.0.png}
& \includegraphics[width=0.12\textwidth]{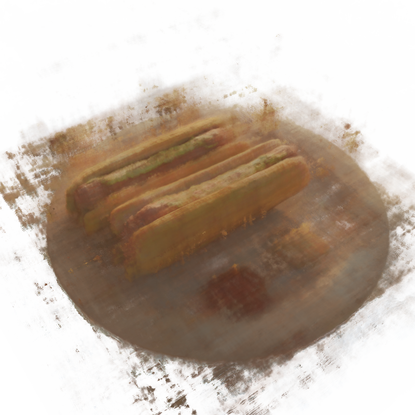}
& \includegraphics[width=0.12\textwidth]{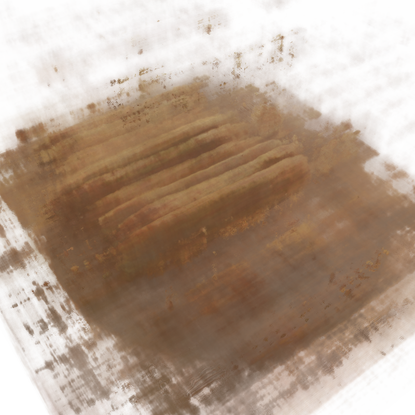}
& \includegraphics[width=0.12\textwidth]{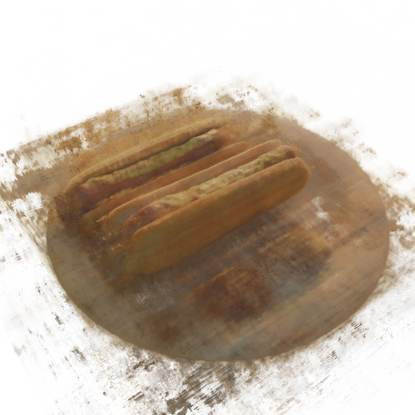}
& \includegraphics[width=0.12\textwidth]{figures/supp_figures/hotdog/1.0.png}
& \includegraphics[width=0.12\textwidth]{figures/supp_figures/hotdog/GT.png} \\
\rowlabel{(c)}
& \includegraphics[width=0.12\textwidth]{figures/supp_figures/hotdog/0.0.png}
& \includegraphics[width=0.12\textwidth]{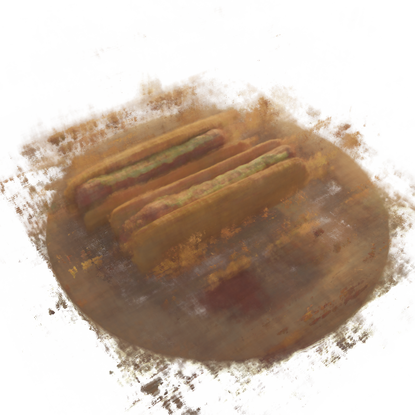}
& \includegraphics[width=0.12\textwidth]{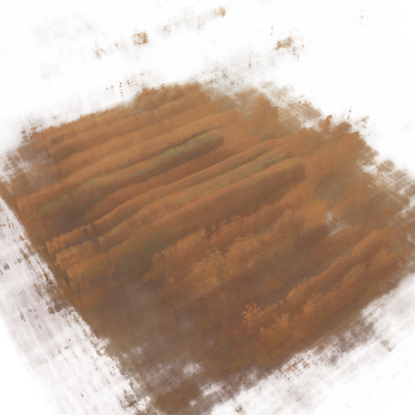}
& \includegraphics[width=0.12\textwidth]{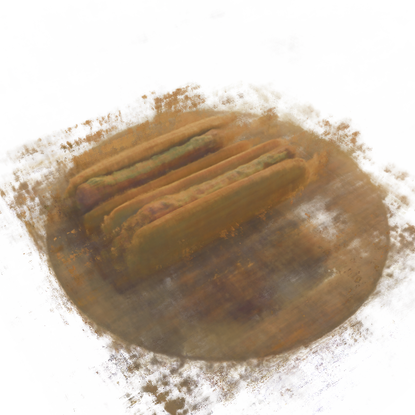}
& \includegraphics[width=0.12\textwidth]{figures/supp_figures/hotdog/1.0.png}
& \includegraphics[width=0.12\textwidth]{figures/supp_figures/hotdog/GT.png} \\
\rowlabel{(d)}
& \includegraphics[width=0.12\textwidth]{figures/supp_figures/hotdog/0.0.png}
& \includegraphics[width=0.12\textwidth]{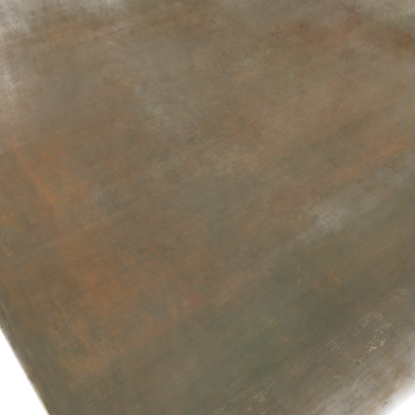}
& \includegraphics[width=0.12\textwidth]{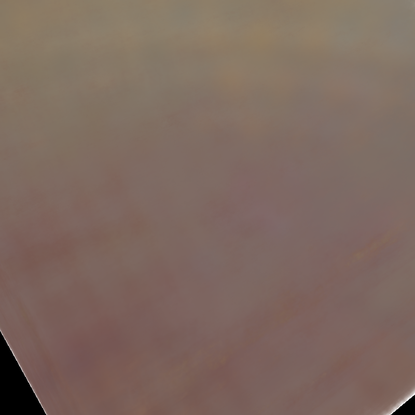}
& \includegraphics[width=0.12\textwidth]{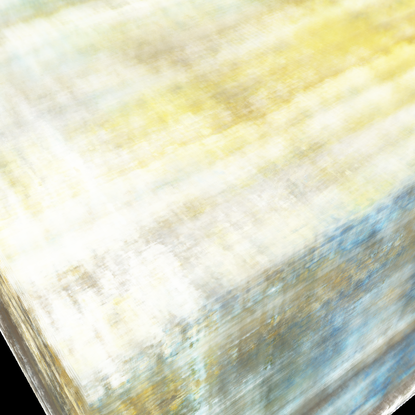}
& \includegraphics[width=0.12\textwidth]{figures/supp_figures/hotdog/1.0.png}
& \includegraphics[width=0.12\textwidth]{figures/supp_figures/hotdog/GT.png} \\
\rowlabel{(e)}
& \includegraphics[width=0.12\textwidth]{figures/supp_figures/hotdog/0.0.png}
& \includegraphics[width=0.12\textwidth]{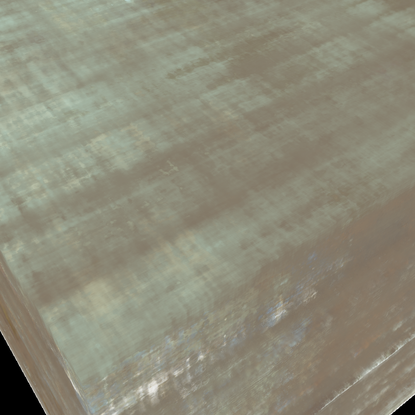}
& \includegraphics[width=0.12\textwidth]{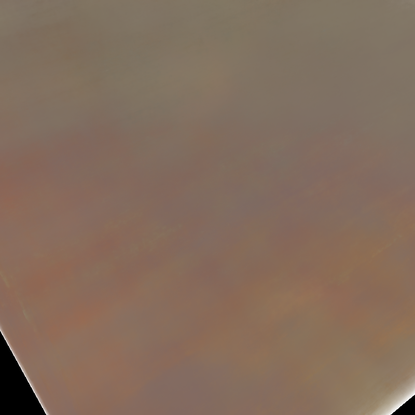}
& \includegraphics[width=0.12\textwidth]{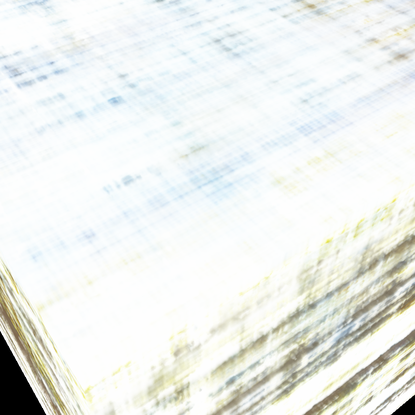}
& \includegraphics[width=0.12\textwidth]{figures/supp_figures/hotdog/1.0.png}
& \includegraphics[width=0.12\textwidth]{figures/supp_figures/hotdog/GT.png} \\
\rowlabel{(f)}
& \includegraphics[width=0.12\textwidth]{figures/supp_figures/hotdog/0.0.png}
& \includegraphics[width=0.12\textwidth]{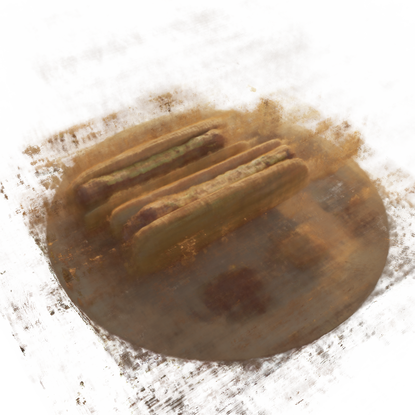}
& \includegraphics[width=0.12\textwidth]{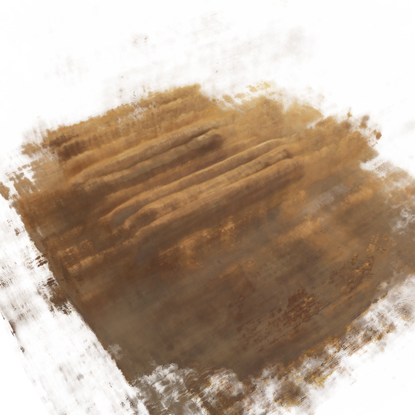}
& \includegraphics[width=0.12\textwidth]{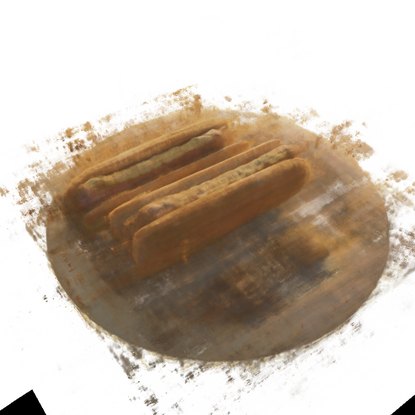}
& \includegraphics[width=0.12\textwidth]{figures/supp_figures/hotdog/1.0.png}
& \includegraphics[width=0.12\textwidth]{figures/supp_figures/hotdog/GT.png} \\
\rowlabel{(g)}
& \includegraphics[width=0.12\textwidth]{figures/supp_figures/hotdog/0.0.png}
& \includegraphics[width=0.12\textwidth]{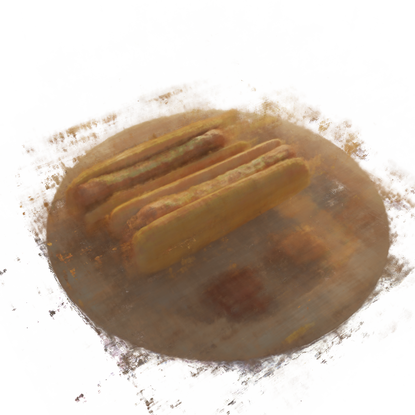}
& \includegraphics[width=0.12\textwidth]{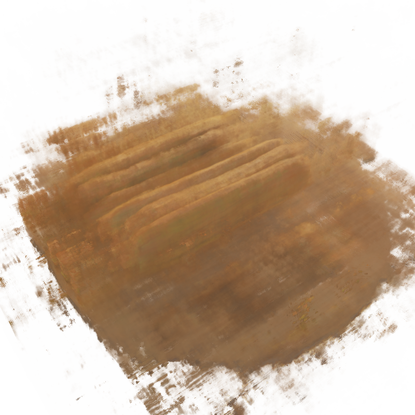}
& \includegraphics[width=0.12\textwidth]{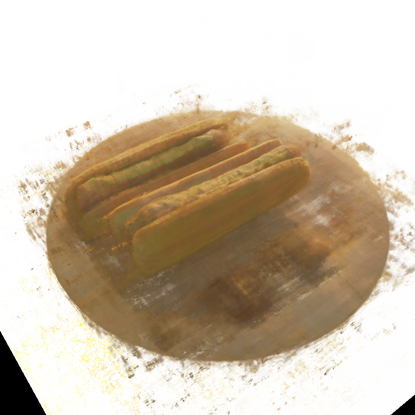}
& \includegraphics[width=0.12\textwidth]{figures/supp_figures/hotdog/1.0.png}
& \includegraphics[width=0.12\textwidth]{figures/supp_figures/hotdog/GT.png} \\
\bottomrule
\end{tabular}
}
\caption{Example of interpolation results for different $\lambda$ values for the Hotdog scene. (a) Naive, (b) Weight Matching
(c) Progressive Weight Matching, (d) OT4P, (e) Progressive OT4P, (f) Proposed w/o Ref, (g) Proposed.}
\label{fig:supp_hotdog}
\end{figure}

\begin{figure}[t]
\centering
\resizebox{0.99\columnwidth}{!}{%
\setlength{\tabcolsep}{2pt}
\begin{tabular}{lccccc|c}
\toprule
& $\lambda=0.0$ & $\lambda=0.25$ & $\lambda=0.50$ & $\lambda=0.75$ & $\lambda=1.0$ & Ground Truth \\
\midrule
\rowlabel{(a)}
& \includegraphics[width=0.12\textwidth]{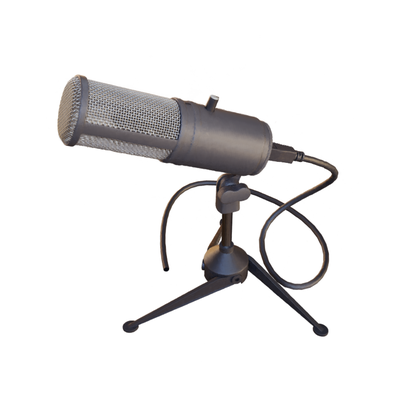}
& \includegraphics[width=0.12\textwidth]{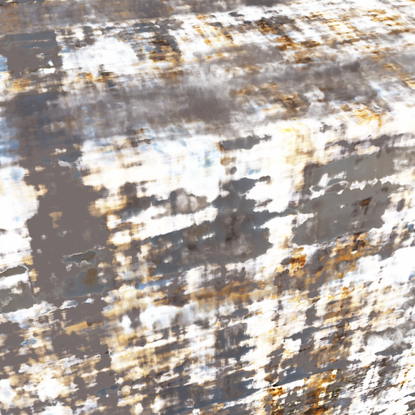}
& \includegraphics[width=0.12\textwidth]{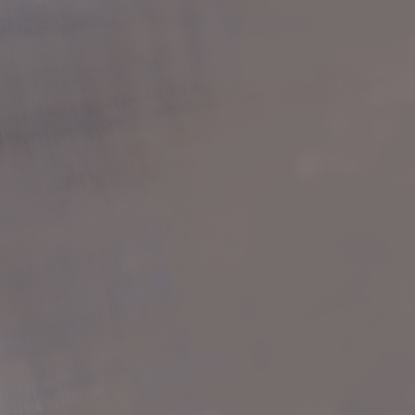}
& \includegraphics[width=0.12\textwidth]{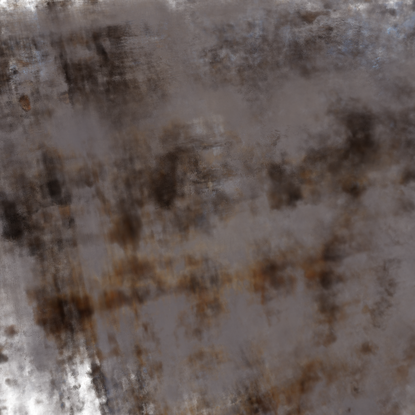}
& \includegraphics[width=0.12\textwidth]{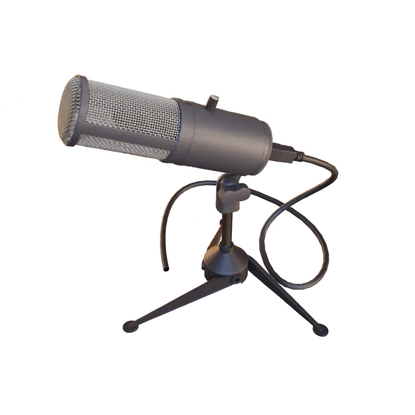}
& \includegraphics[width=0.12\textwidth]{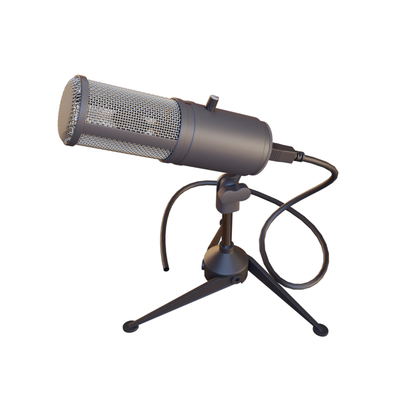} \\
\rowlabel{(b)}
& \includegraphics[width=0.12\textwidth]{figures/supp_figures/mic/0.0.png}
& \includegraphics[width=0.12\textwidth]{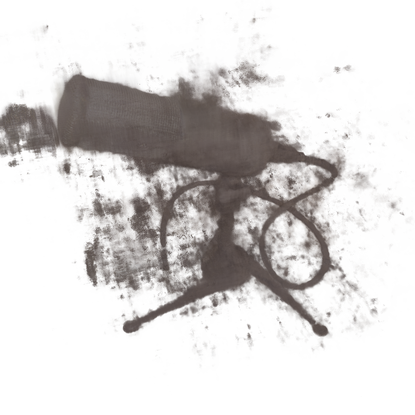}
& \includegraphics[width=0.12\textwidth]{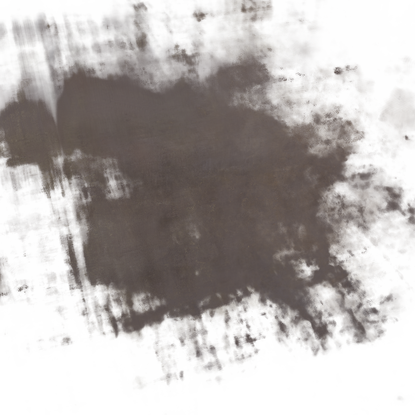}
& \includegraphics[width=0.12\textwidth]{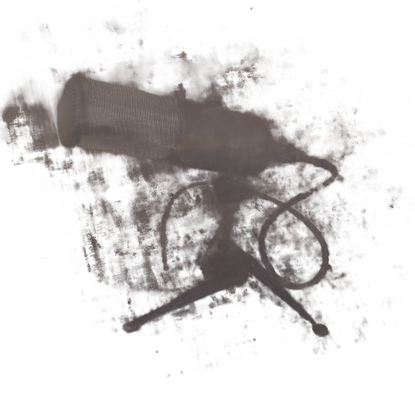}
& \includegraphics[width=0.12\textwidth]{figures/supp_figures/mic/1.0.png}
& \includegraphics[width=0.12\textwidth]{figures/supp_figures/mic/GT.png} \\
\rowlabel{(c)}
& \includegraphics[width=0.12\textwidth]{figures/supp_figures/mic/0.0.png}
& \includegraphics[width=0.12\textwidth]{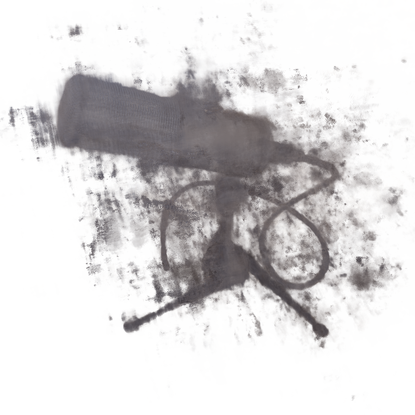}
& \includegraphics[width=0.12\textwidth]{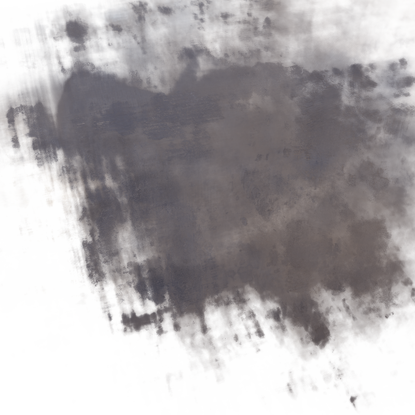}
& \includegraphics[width=0.12\textwidth]{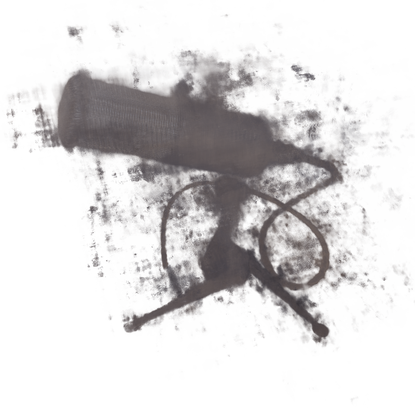}
& \includegraphics[width=0.12\textwidth]{figures/supp_figures/mic/1.0.png}
& \includegraphics[width=0.12\textwidth]{figures/supp_figures/mic/GT.png} \\
\rowlabel{(d)}
& \includegraphics[width=0.12\textwidth]{figures/supp_figures/mic/0.0.png}
& \includegraphics[width=0.12\textwidth]{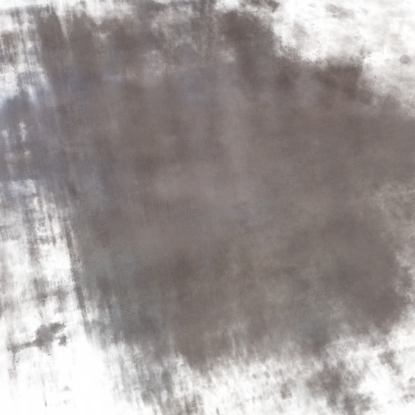}
& \includegraphics[width=0.12\textwidth]{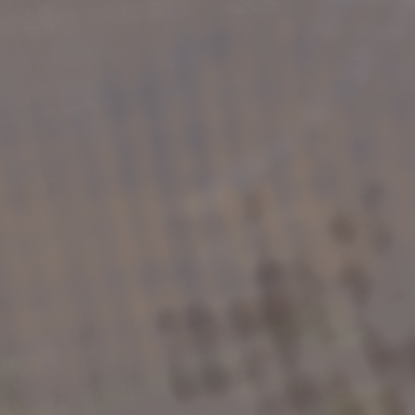}
& \includegraphics[width=0.12\textwidth]{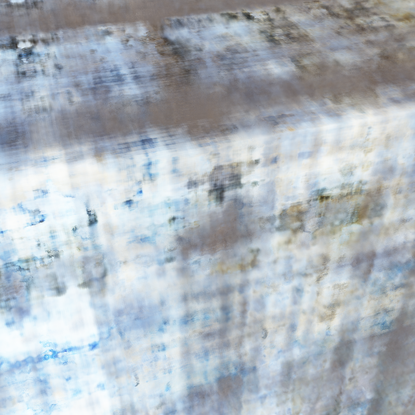}
& \includegraphics[width=0.12\textwidth]{figures/supp_figures/mic/1.0.png}
& \includegraphics[width=0.12\textwidth]{figures/supp_figures/mic/GT.png} \\
\rowlabel{(e)}
& \includegraphics[width=0.12\textwidth]{figures/supp_figures/mic/0.0.png}
& \includegraphics[width=0.12\textwidth]{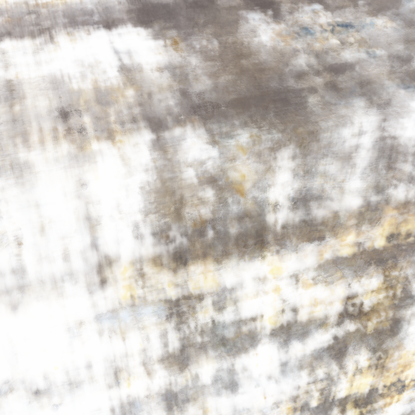}
& \includegraphics[width=0.12\textwidth]{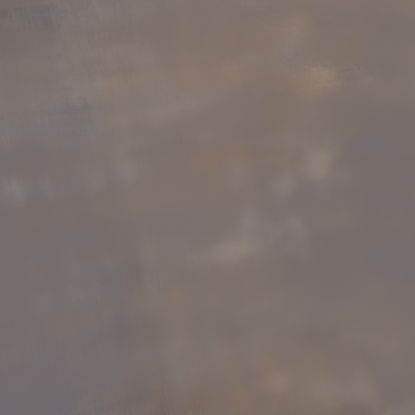}
& \includegraphics[width=0.12\textwidth]{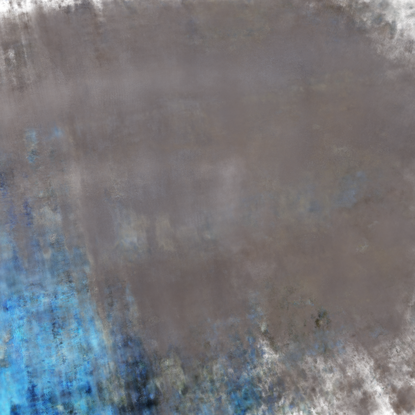}
& \includegraphics[width=0.12\textwidth]{figures/supp_figures/mic/1.0.png}
& \includegraphics[width=0.12\textwidth]{figures/supp_figures/mic/GT.png} \\
\rowlabel{(f)}
& \includegraphics[width=0.12\textwidth]{figures/supp_figures/mic/0.0.png}
& \includegraphics[width=0.12\textwidth]{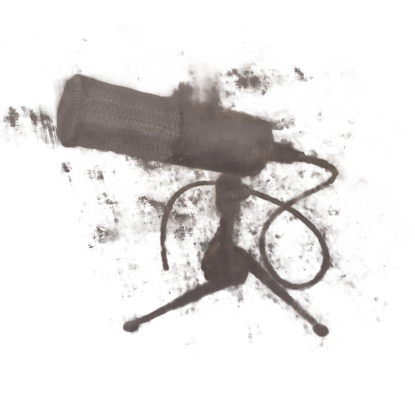}
& \includegraphics[width=0.12\textwidth]{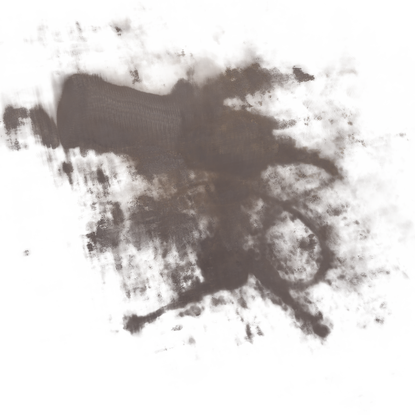}
& \includegraphics[width=0.12\textwidth]{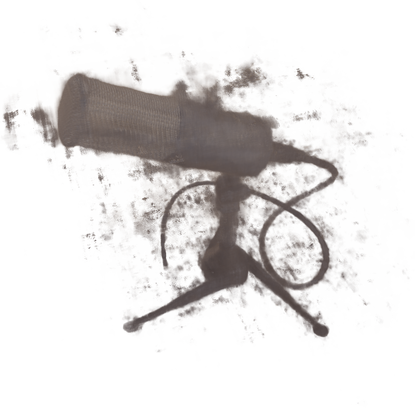}
& \includegraphics[width=0.12\textwidth]{figures/supp_figures/mic/1.0.png}
& \includegraphics[width=0.12\textwidth]{figures/supp_figures/mic/GT.png} \\
\rowlabel{(g)}
& \includegraphics[width=0.12\textwidth]{figures/supp_figures/mic/0.0.png}
& \includegraphics[width=0.12\textwidth]{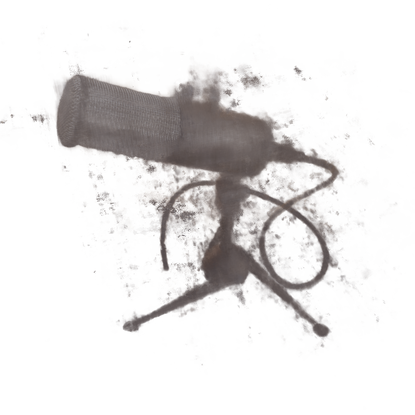}
& \includegraphics[width=0.12\textwidth]{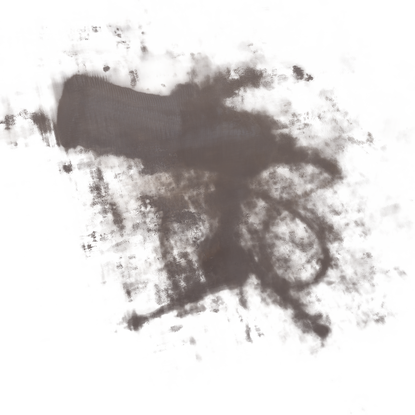}
& \includegraphics[width=0.12\textwidth]{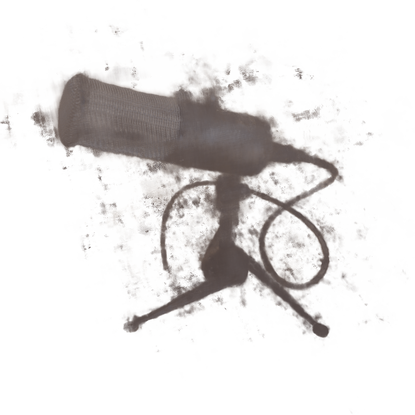}
& \includegraphics[width=0.12\textwidth]{figures/supp_figures/mic/1.0.png}
& \includegraphics[width=0.12\textwidth]{figures/supp_figures/mic/GT.png} \\
\bottomrule
\end{tabular}
}
\caption{Example of interpolation results for different $\lambda$ values for the Mic scene. (a) Naive, (b) Weight Matching
(c) Progressive Weight Matching, (d) OT4P, (e) Progressive OT4P, (f) Proposed  w/o Ref, (g) Proposed.}
\label{fig:supp_mic}
\end{figure}

\begin{figure}[t]
\centering
\resizebox{0.99\columnwidth}{!}{%
\setlength{\tabcolsep}{2pt}
\begin{tabular}{lccccc|c}
\toprule
& $\lambda=0.0$ & $\lambda=0.25$ & $\lambda=0.50$ & $\lambda=0.75$ & $\lambda=1.0$ & Ground Truth \\
\midrule
\rowlabel{(a)}
& \includegraphics[width=0.12\textwidth]{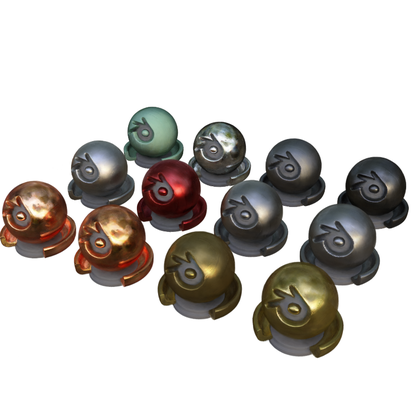}
& \includegraphics[width=0.12\textwidth]{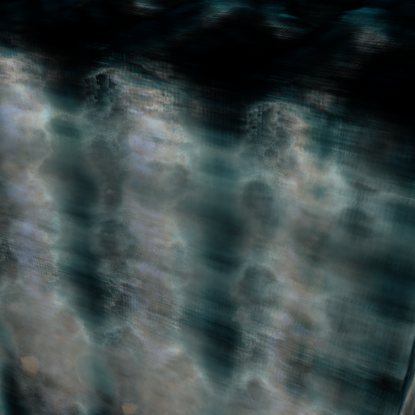}
& \includegraphics[width=0.12\textwidth]{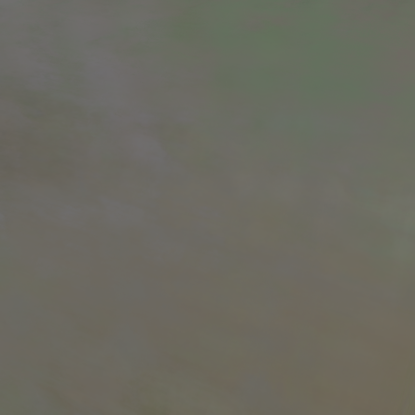}
& \includegraphics[width=0.12\textwidth]{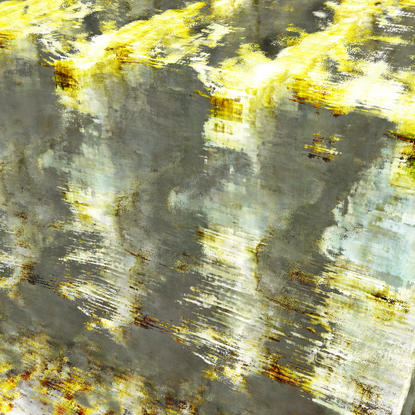}
& \includegraphics[width=0.12\textwidth]{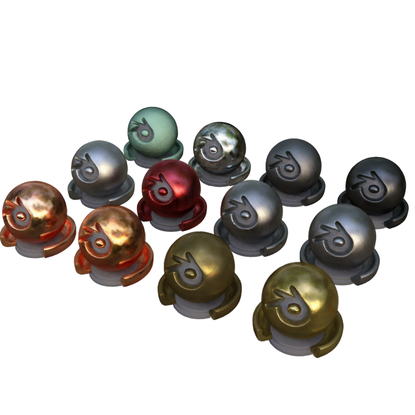}
& \includegraphics[width=0.12\textwidth]{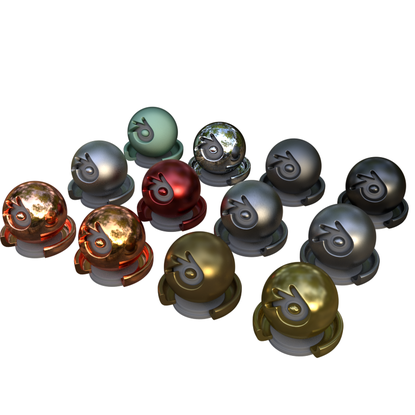} \\
\rowlabel{(b)}
& \includegraphics[width=0.12\textwidth]{figures/supp_figures/material/0.0.png}
& \includegraphics[width=0.12\textwidth]{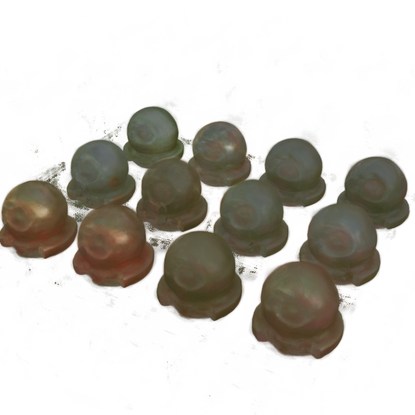}
& \includegraphics[width=0.12\textwidth]{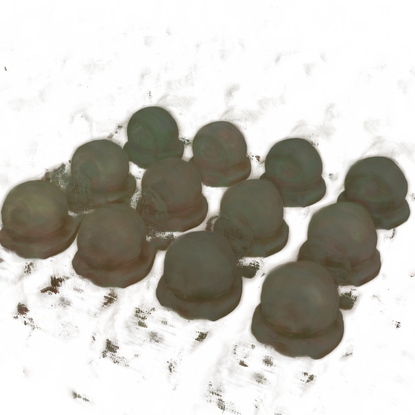}
& \includegraphics[width=0.12\textwidth]{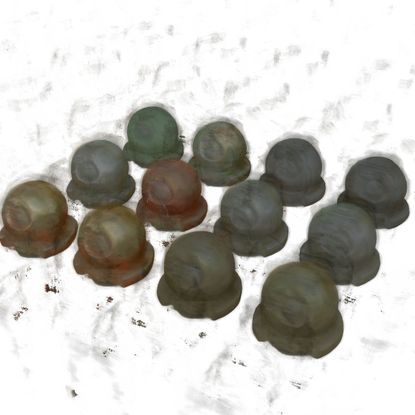}
& \includegraphics[width=0.12\textwidth]{figures/supp_figures/material/1.0.png}
& \includegraphics[width=0.12\textwidth]{figures/supp_figures/material/GT.png} \\
\rowlabel{(c)}
& \includegraphics[width=0.12\textwidth]{figures/supp_figures/material/0.0.png}
& \includegraphics[width=0.12\textwidth]{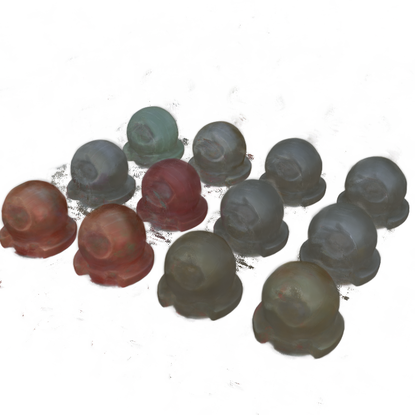}
& \includegraphics[width=0.12\textwidth]{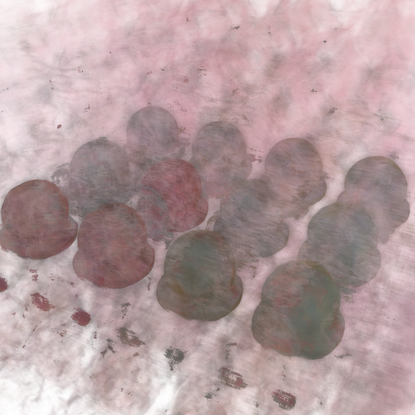}
& \includegraphics[width=0.12\textwidth]{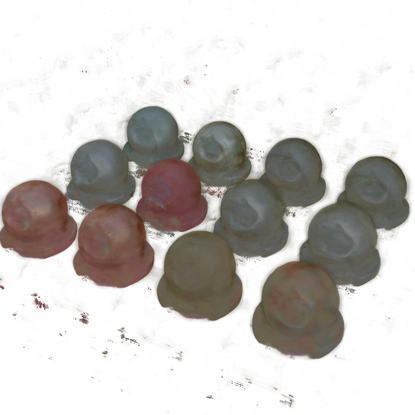}
& \includegraphics[width=0.12\textwidth]{figures/supp_figures/material/1.0.png}
& \includegraphics[width=0.12\textwidth]{figures/supp_figures/material/GT.png} \\
\rowlabel{(d)}
& \includegraphics[width=0.12\textwidth]{figures/supp_figures/material/0.0.png}
& \includegraphics[width=0.12\textwidth]{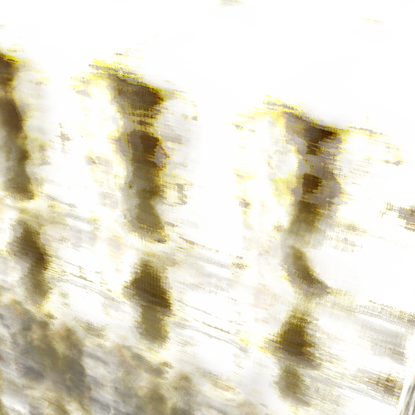}
& \includegraphics[width=0.12\textwidth]{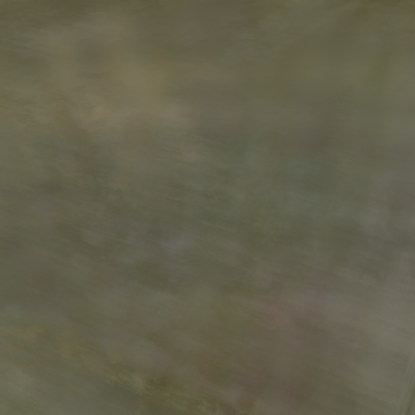}
& \includegraphics[width=0.12\textwidth]{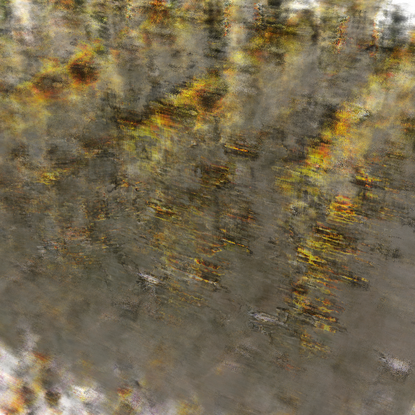}
& \includegraphics[width=0.12\textwidth]{figures/supp_figures/material/1.0.png}
& \includegraphics[width=0.12\textwidth]{figures/supp_figures/material/GT.png} \\
\rowlabel{(e)}
& \includegraphics[width=0.12\textwidth]{figures/supp_figures/material/0.0.png}
& \includegraphics[width=0.12\textwidth]{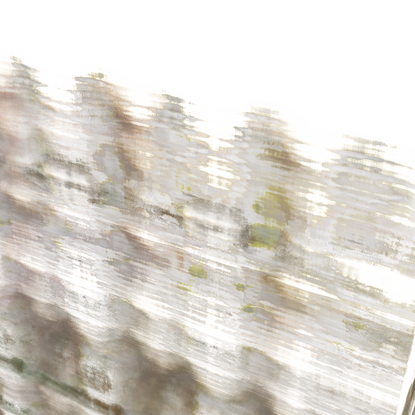}
& \includegraphics[width=0.12\textwidth]{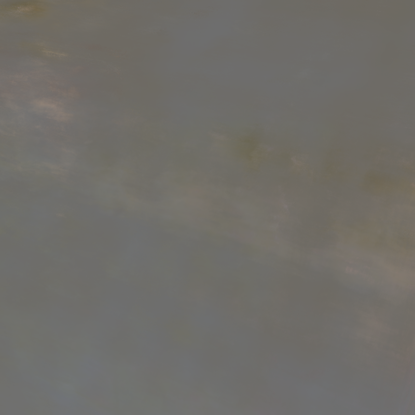}
& \includegraphics[width=0.12\textwidth]{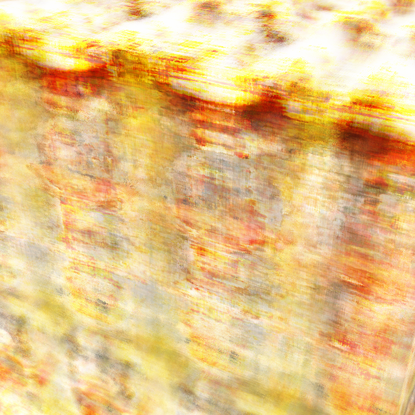}
& \includegraphics[width=0.12\textwidth]{figures/supp_figures/material/1.0.png}
& \includegraphics[width=0.12\textwidth]{figures/supp_figures/material/GT.png} \\
\rowlabel{(f)}
& \includegraphics[width=0.12\textwidth]{figures/supp_figures/material/0.0.png}
& \includegraphics[width=0.12\textwidth]{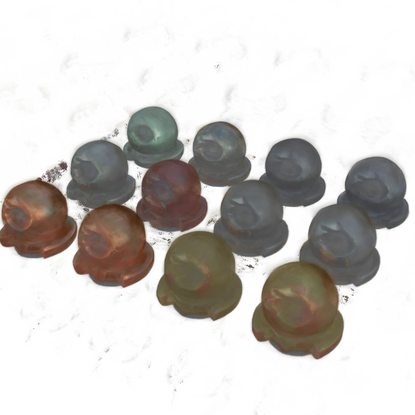}
& \includegraphics[width=0.12\textwidth]{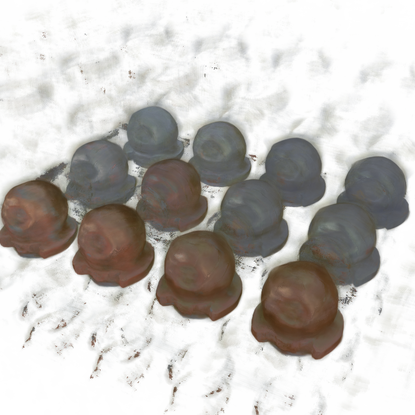}
& \includegraphics[width=0.12\textwidth]{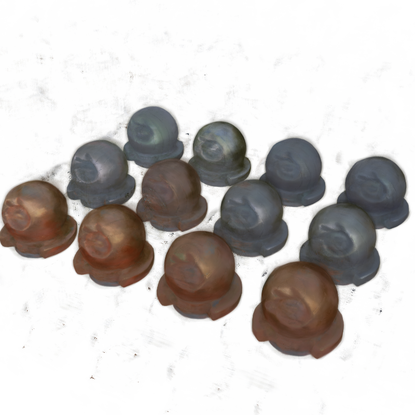}
& \includegraphics[width=0.12\textwidth]{figures/supp_figures/material/1.0.png}
& \includegraphics[width=0.12\textwidth]{figures/supp_figures/material/GT.png} \\
\rowlabel{(g)}
& \includegraphics[width=0.12\textwidth]{figures/supp_figures/material/0.0.png}
& \includegraphics[width=0.12\textwidth]{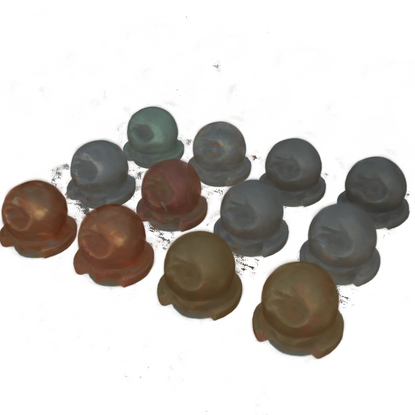}
& \includegraphics[width=0.12\textwidth]{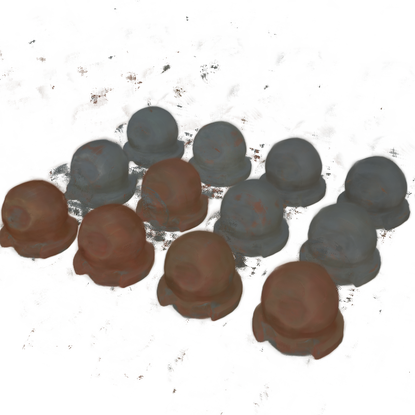}
& \includegraphics[width=0.12\textwidth]{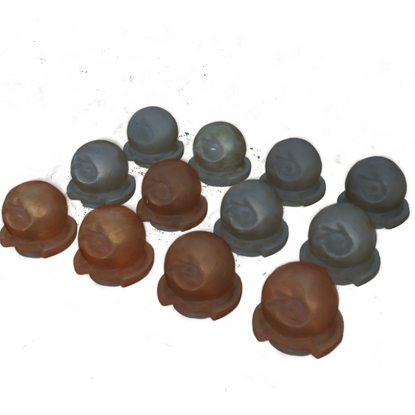}
& \includegraphics[width=0.12\textwidth]{figures/supp_figures/material/1.0.png}
& \includegraphics[width=0.12\textwidth]{figures/supp_figures/material/GT.png} \\
\bottomrule
\end{tabular}
}
\caption{Example of interpolation results for different $\lambda$ values for the Materials scene. (a) Naive, (b) Weight Matching
(c) Progressive Weight Matching, (d) OT4P, (e) Progressive OT4P, (f) Proposed w/o Ref, (g) Proposed.}
\label{fig:supp_materials}
\end{figure}

\begin{figure}[t]
\centering
\resizebox{0.99\columnwidth}{!}{%
\setlength{\tabcolsep}{2pt}
\begin{tabular}{lccccc|c}
\toprule
& $\lambda=0.0$ & $\lambda=0.25$ & $\lambda=0.50$ & $\lambda=0.75$ & $\lambda=1.0$ & Ground Truth \\
\midrule
\rowlabel{(a)}
& \includegraphics[width=0.12\textwidth]{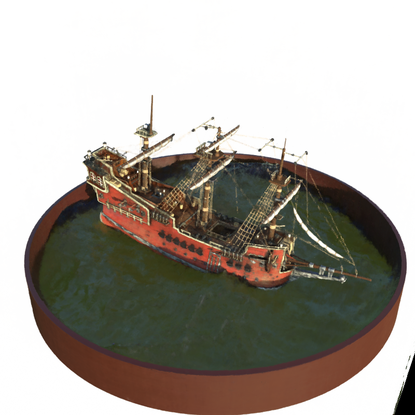}
& \includegraphics[width=0.12\textwidth]{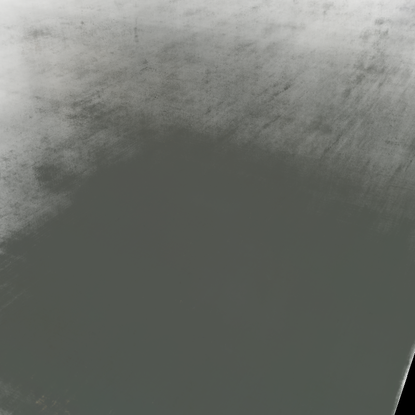}
& \includegraphics[width=0.12\textwidth]{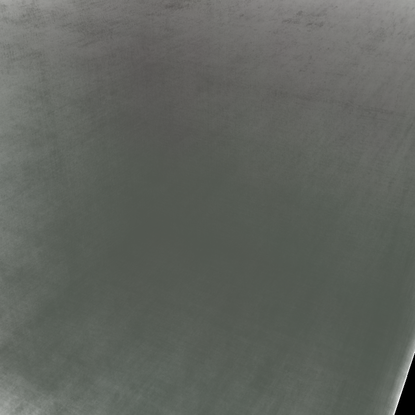}
& \includegraphics[width=0.12\textwidth]{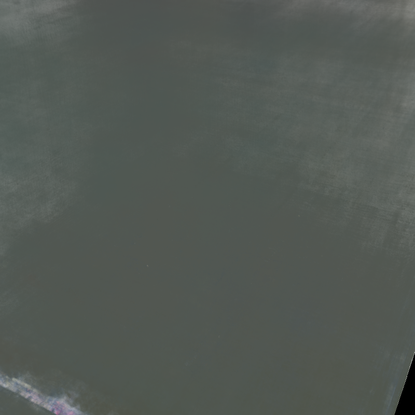}
& \includegraphics[width=0.12\textwidth]{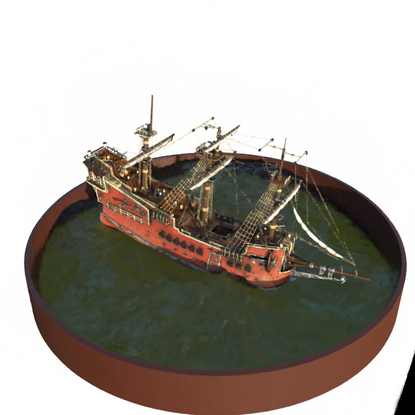}
& \includegraphics[width=0.12\textwidth]{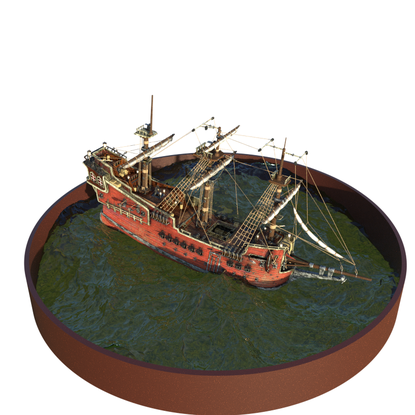} \\
\rowlabel{(b)}
& \includegraphics[width=0.12\textwidth]{figures/supp_figures/ship/0.0.png}
& \includegraphics[width=0.12\textwidth]{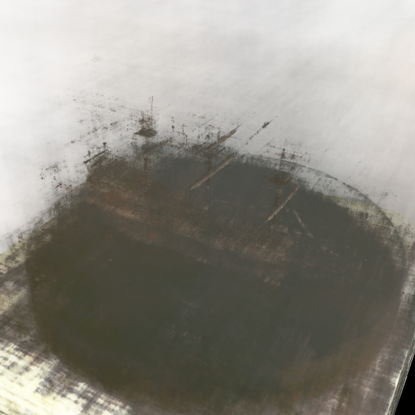}
& \includegraphics[width=0.12\textwidth]{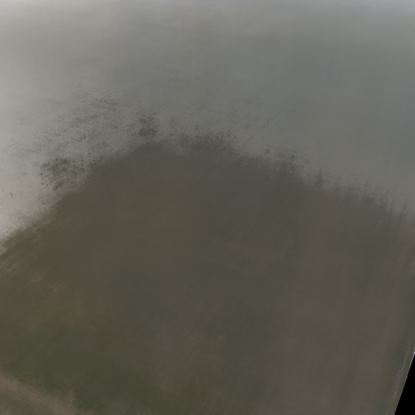}
& \includegraphics[width=0.12\textwidth]{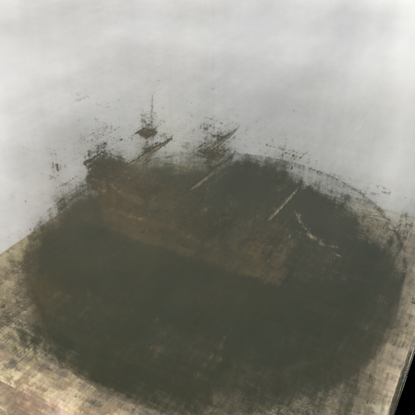}
& \includegraphics[width=0.12\textwidth]{figures/supp_figures/ship/1.0.png}
& \includegraphics[width=0.12\textwidth]{figures/supp_figures/ship/GT.png} \\
\rowlabel{(c)}
& \includegraphics[width=0.12\textwidth]{figures/supp_figures/ship/0.0.png}
& \includegraphics[width=0.12\textwidth]{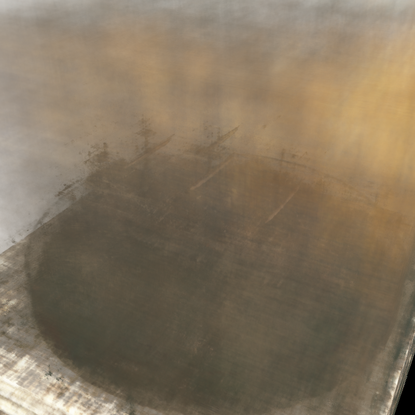}
& \includegraphics[width=0.12\textwidth]{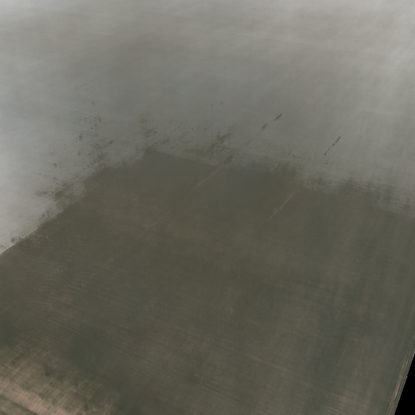}
& \includegraphics[width=0.12\textwidth]{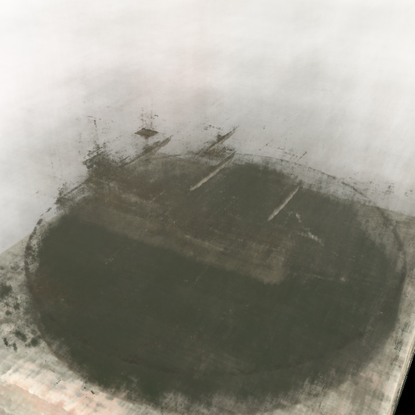}
& \includegraphics[width=0.12\textwidth]{figures/supp_figures/ship/1.0.png}
& \includegraphics[width=0.12\textwidth]{figures/supp_figures/ship/GT.png} \\
\rowlabel{(d)}
& \includegraphics[width=0.12\textwidth]{figures/supp_figures/ship/0.0.png}
& \includegraphics[width=0.12\textwidth]{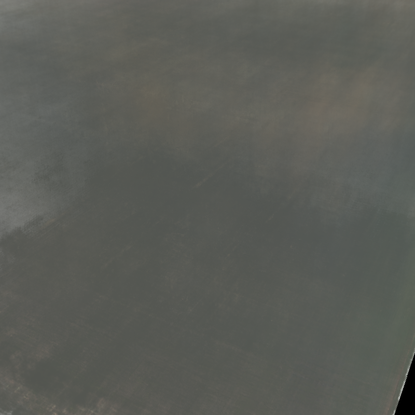}
& \includegraphics[width=0.12\textwidth]{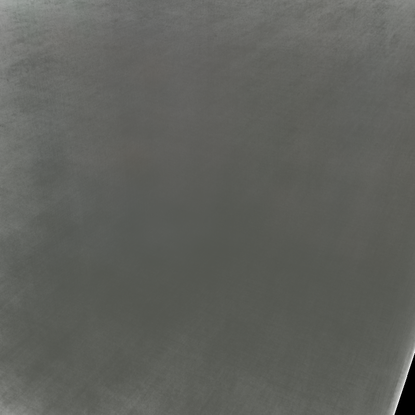}
& \includegraphics[width=0.12\textwidth]{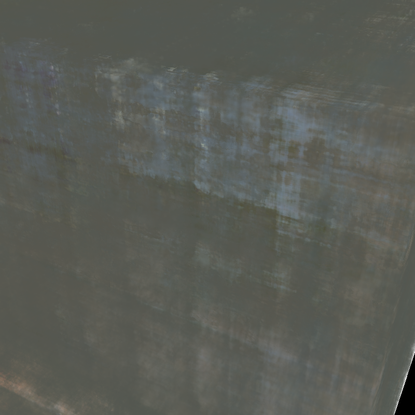}
& \includegraphics[width=0.12\textwidth]{figures/supp_figures/ship/1.0.png}
& \includegraphics[width=0.12\textwidth]{figures/supp_figures/ship/GT.png} \\
\rowlabel{(e)}
& \includegraphics[width=0.12\textwidth]{figures/supp_figures/ship/0.0.png}
& \includegraphics[width=0.12\textwidth]{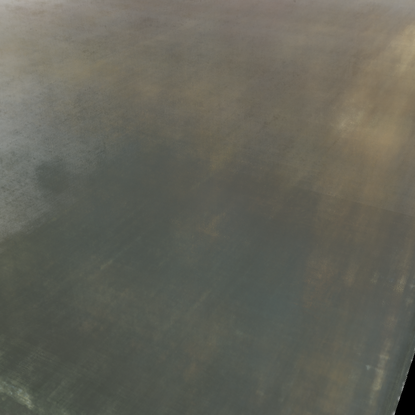}
& \includegraphics[width=0.12\textwidth]{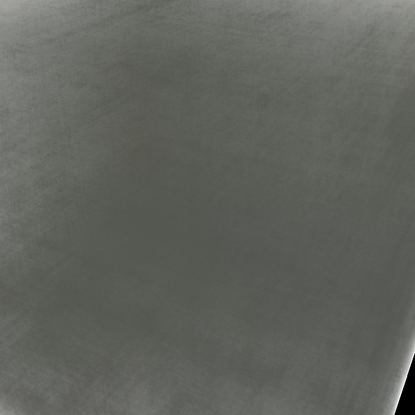}
& \includegraphics[width=0.12\textwidth]{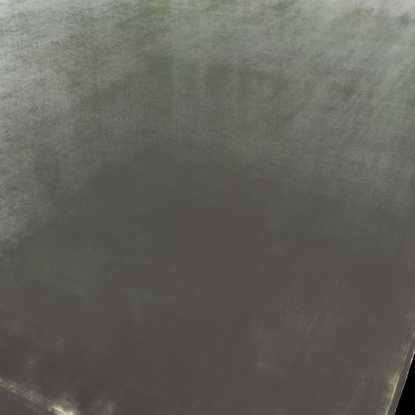}
& \includegraphics[width=0.12\textwidth]{figures/supp_figures/ship/1.0.png}
& \includegraphics[width=0.12\textwidth]{figures/supp_figures/ship/GT.png} \\
\rowlabel{(f)}
& \includegraphics[width=0.12\textwidth]{figures/supp_figures/ship/0.0.png}
& \includegraphics[width=0.12\textwidth]{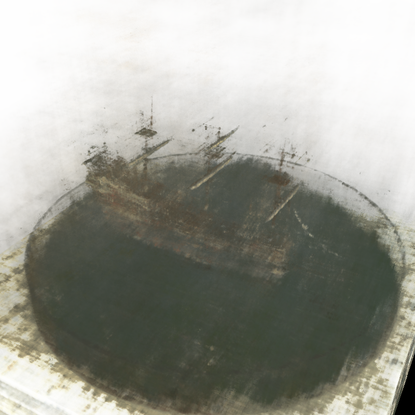}
& \includegraphics[width=0.12\textwidth]{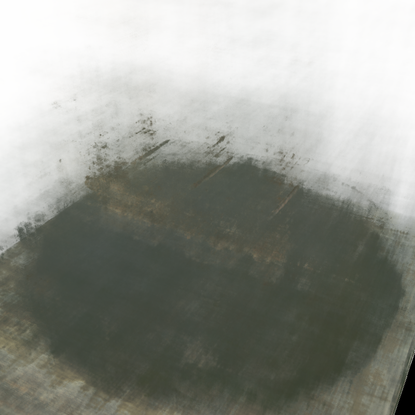}
& \includegraphics[width=0.12\textwidth]{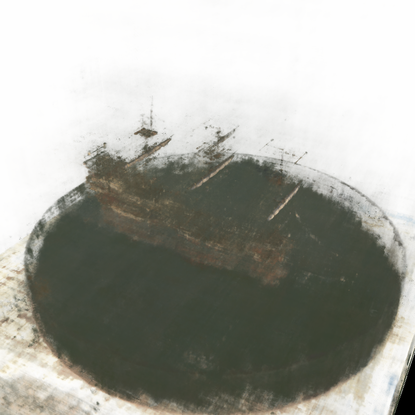}
& \includegraphics[width=0.12\textwidth]{figures/supp_figures/ship/1.0.png}
& \includegraphics[width=0.12\textwidth]{figures/supp_figures/ship/GT.png} \\
\rowlabel{(g)}
& \includegraphics[width=0.12\textwidth]{figures/supp_figures/ship/0.0.png}
& \includegraphics[width=0.12\textwidth]{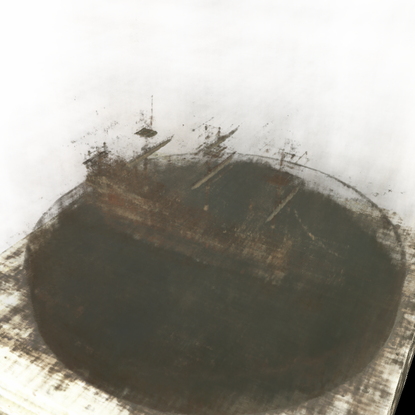}
& \includegraphics[width=0.12\textwidth]{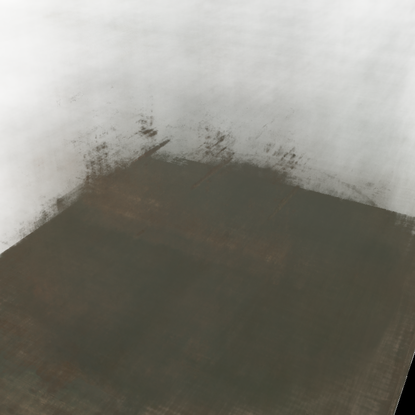}
& \includegraphics[width=0.12\textwidth]{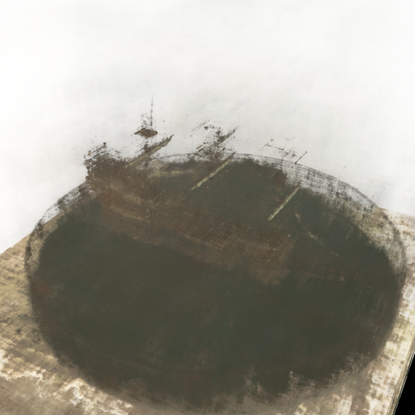}
& \includegraphics[width=0.12\textwidth]{figures/supp_figures/ship/1.0.png}
& \includegraphics[width=0.12\textwidth]{figures/supp_figures/ship/GT.png} \\
\bottomrule
\end{tabular}
}
\caption{Example of interpolation results for different $\lambda$ values for the Ship scene. (a) Naive, (b) Weight Matching
(c) Progressive Weight Matching, (d) OT4P, (e) Progressive OT4P, (f) Proposed w/o Ref, (g) Proposed.}
\label{fig:supp_ship}
\end{figure}

A similar trend is observed in the Chair (Figure~\ref{fig:supp_chair}) and Ficus (Figure~\ref{fig:supp_ficus}) scenes, where geometric consistency is particularly important.
Baseline methods often produce collapsed or view-inconsistent structures around $\lambda=0.5$, suggesting imperfect alignment in the weight space.
The proposed method reduces such failures and yields smoother visual transitions.
For Drums (Figure~\ref{fig:supp_drums}), Hotdog (Figure~\ref{fig:supp_hotdog}), Mic (Figure~\ref{fig:supp_mic}), Materials (Figure~\ref{fig:supp_materials}), and Ship (Figure~\ref{fig:supp_ship}), the proposed method again exhibits the most stable interpolation behavior among the compared methods.

Overall, the qualitative results were consistent with the quantitative trends.
Better alignment leads to smoother interpolation in the NeRF weight space, particularly near the middle of the interpolation path, where naive interpolation often fails.
These results further support the effectiveness of progressive alignment with temperature-controlled spectral sharpening in independently trained neural fields.

\clearpage

\section{Derivation of the Trace-Maximization Form}

Consider the following standard weight-alignment objective:
\begin{equation}
\min_{\{\mat{P}_l\}} \sum_{l=1}^{L}
\left\|\mat{W}_l^{(A)}-\mat{P}_l \mat{W}_l^{(B)} \mat{P}_{l-1}^\top \right\|_F^2,
\qquad
\mat{P}_l \text{ is a permutation matrix.}
\end{equation}
When $\mat{P}_{l-1}$ and $\mat{P}_{l+1}$ are fixed, the terms that depend on $\mat{P}_l$ are
\begin{align}
\left\|\mat{W}_l^{(A)}-\mat{P}_l \mat{W}_l^{(B)} \mat{P}_{l-1}^\top \right\|_F^2
+
\left\|\mat{W}_{l+1}^{(A)}-\mat{P}_{l+1}\mat{W}_{l+1}^{(B)} \mat{P}_{l}^\top \right\|_F^2.
\end{align}
Expanding the Frobenius norms and using $\|\mat{P}\mat{X}\mat{Q}\|_F=\|\mat{X}\|_F$ for orthogonal $\mat{P}$ and $\mat{Q}$, the $\mat{P}_l$-dependent part reduces to
\begin{align}
-2\Tr\!\Big((\mat{W}_l^{(A)})^\top \mat{P}_l \mat{W}_l^{(B)} \mat{P}_{l-1}^\top\Big)
-2\Tr\!\Big((\mat{W}_{l+1}^{(A)})^\top \mat{P}_{l+1}\mat{W}_{l+1}^{(B)} \mat{P}_{l}^\top\Big).
\end{align}
Using the cyclic property of the trace, this is equivalent, up to constants, to
\begin{equation}
\max_{\mat{P}_l}\ \Tr(\mat{P}_l^\top \mat{C}_l),
\end{equation}
where
\begin{equation}
\mat{C}_l=
\mat{W}_{l}^{(A)} (\mat{P}_{l-1} \mat{W}_{l}^{(B)})^\top
+
(\mat{W}_{l+1}^{(A)})^\top \mat{P}_{l+1} \mat{W}_{l+1}^{(B)}.
\end{equation}
\hfill$\blacksquare$

\section{An example where Progressive Weight Matching fails}
\label{sec:wm-fail}
Consider a 3-layer linear network with three feature dimensions:
\begin{equation}
f(x) = W_3 W_2 W_1 x, \qquad W_l \in \mathbb{R}^{3\times3}.
\end{equation}

As a numerical example, the weights of model A are
\begin{equation}
W_{1}^{(A)}=
\begin{bmatrix}
5&5&-3\\
3&-1&-2\\
-4&1&-3
\end{bmatrix},\quad
W_{2}^{(A)}=
\begin{bmatrix}
2&4&5\\
-5&1&-3\\
0&-2&-5
\end{bmatrix},
W_{3}^{(A)}=
\begin{bmatrix}
-5&-4&5\\
-2&2&-1\\
-3&2&-5
\end{bmatrix},
\end{equation}
and the weights of model B are
\begin{equation}
W_{1}^{(B)}=
\begin{bmatrix}
3&4&5\\
2&5&3\\
1&2&-4
\end{bmatrix},\quad
W_{2}^{(B)}=
\begin{bmatrix}
-2&-1&4\\
-3&2&2\\
2&2&4
\end{bmatrix},
W_{3}^{(B)}=
\begin{bmatrix}
0&-5&-3\\
5&-4&2\\
1&4&3
\end{bmatrix} .
\end{equation}
The ground-truth correspondences are the identity permutations, written here in index notation as
$P_1^\ast = \begin{bmatrix}0&1&2\end{bmatrix},\qquad P_2^\ast = \begin{bmatrix}0&1&2\end{bmatrix}$.
Progressive Weight Matching (WM) alternately solves the optimization problem at each layer:
\begin{equation}
\max_{P_l\in\mathcal P} \mathrm{Tr}(P_l^\top C_l)
\end{equation}
where the correlation matrices $C_1$ and $C_2$ depend on the permutation matrices of the other layers, as follows:
\begin{equation}
C_1(P_2)=W_{1}^{(A)}(W_{1}^{(B)})^\top + (W_{2}^{(A)})^\top P_2 W_{2}^{(B)},
\end{equation}
\begin{equation}
C_2(P_1)=W_{2}^{(A)}(P_1 W_{2}^{(B)})^\top + (W_{3}^{(A)})^\top W_{3}^{(B)},
\end{equation}
Here, $P_1$ and $P_2$ are permutation matrices chosen to maximize the trace terms iteratively, reflecting the alignment between the corresponding weight matrices of models A and B at each layer.

Given $P_1=P_2=\begin{bmatrix}0&1&2\end{bmatrix}$, the correlation matrices are computed as:
\begin{equation}
C_1 =
\begin{bmatrix}
31&14&25\\
-20&-11&19\\
-34&-33&4
\end{bmatrix},
\qquad
\mathrm{Tr}(C_1)=24
\end{equation}
\begin{equation}
C_2 =
\begin{bmatrix}
-1&33&34\\
9&31&2\\
-28&-55&-56
\end{bmatrix},
\qquad
\mathrm{Tr}(C_2)=-26 .
\end{equation}
Given $P_1=\begin{bmatrix}0&1&2\end{bmatrix}$ fixed, optimizing $P_2$ via Weight Matching on the correlation matrix $C_2$ yields the following:
\begin{itemize}
\item When $P_1=\begin{bmatrix}0&1&2\end{bmatrix}$ (the correct identity permutation), the trace of $C_2$ is $-26$.
\item However, the maximum trace is achieved at $P_2=\begin{bmatrix}2&1&0\end{bmatrix}$, with a trace value of $37$.
\end{itemize}
Therefore, Weight Matching updates $P_2$ to $P_2=\begin{bmatrix}2&1&0\end{bmatrix}$.
Under this incorrect $P_2$, optimizing $P_1$ yields  
\begin{itemize}
\item When $P_1=\begin{bmatrix}0&1&2\end{bmatrix}$ (the correct permutation), the trace value is $6$.
\item While the maximum trace is achieved at $P_1=\begin{bmatrix}1&0&2\end{bmatrix}$ with a value of 18.
\end{itemize}
Repeating this alternating optimization leads to a local fixed point at these incorrect permutations, which differs from the true correspondence  $P_1=P_2=\begin{bmatrix}0&1&2\end{bmatrix}$.
This demonstrates that the iterative Weight Matching procedure can converge to a spurious local optimum rather than the correct alignment.

On the other hand, applying the proposed method of this paper from the same initial values $P_1=P_2=\begin{bmatrix}0&1&2\end{bmatrix}$ exhibits different behavior.

In each iteration, the proposed method first obtains a provisional anchor $\hat{P}_l$ from the correlation matrix using the Hungarian algorithm. Subsequently, the matrix $\hat{P}_l^\top C_l$ is spectrally decomposed and sharpened by applying the exponent $\tau$ to its spectrum, yielding a matrix that is projected onto the orthogonal group.
This procedure gradually sharpens the continuous correlation into a permutation-like orthogonal matrix.
The temperature $\tau$ is gradually decreased from $1.0$ to $0.0$.

The sequence of iterates in this toy example is summarized below:
\begin{center}
\begin{tabular}{c|c}
iter & $(P_1, P_2)$ (after discretization) \\
\hline
1 & $(0,1,2),(0,1,2)$ \\
2 & $(0,2,1),(1,2,0)$ \\
3 & $(1,0,2),(0,1,2)$ \\
4 & $(1,2,0),(1,2,0)$ \\
5 & $(0,1,2),(0,1,2)$ \\
6--8 & $(0,1,2),(0,1,2)$
\end{tabular}
\end{center}
As $\tau$ decreases, the iterates gradually converge toward permutation-like structures, and the final solution recovers the correct correspondence,
$P_1 = P_2 = \begin{bmatrix}0 & 1 & 2\end{bmatrix}$.

\clearpage

\end{document}